\catcode`\@=11\relax
\newwrite\@unused
\def\typeout#1{{\let\protect\string\immediate\write\@unused{#1}}}
\def\@nnil{\@nil}
\def\@empty{}
\def\@psdonoop#1\@@#2#3{}
\def\@psdo#1:=#2\do#3{\edef\@psdotmp{#2}\ifx\@psdotmp\@empty \else
    \expandafter\@psdoloop#2,\@nil,\@nil\@@#1{#3}\fi}
\def\@psdoloop#1,#2,#3\@@#4#5{\def#4{#1}\ifx #4\@nnil \else
       #5\def#4{#2}\ifx #4\@nnil \else#5\@ipsdoloop #3\@@#4{#5}\fi\fi}
\def\@ipsdoloop#1,#2\@@#3#4{\def#3{#1}\ifx #3\@nnil 
       \let\@nextwhile=\@psdonoop \else
      #4\relax\let\@nextwhile=\@ipsdoloop\fi\@nextwhile#2\@@#3{#4}}
\def\@tpsdo#1:=#2\do#3{\xdef\@psdotmp{#2}\ifx\@psdotmp\@empty \else
    \@tpsdoloop#2\@nil\@nil\@@#1{#3}\fi}
\def\@tpsdoloop#1#2\@@#3#4{\def#3{#1}\ifx #3\@nnil 
       \let\@nextwhile=\@psdonoop \else
      #4\relax\let\@nextwhile=\@tpsdoloop\fi\@nextwhile#2\@@#3{#4}}
\def\psdraft{
	\def\@psdraft{0}
}
\def\psfull{
	\def\@psdraft{100}
}
\psfull
\newif\if@prologfile
\newif\if@postlogfile
\newif\if@bbllx
\newif\if@bblly
\newif\if@bburx
\newif\if@bbury
\newif\if@height
\newif\if@width
\newif\if@rheight
\newif\if@rwidth
\newif\if@clip
\def\@p@@sclip#1{\@cliptrue}
\def\@p@@sfile#1{
		   \def\@p@sfile{#1}
}
\def\@p@@sfigure#1{\def\@p@sfile{#1}}
\def\@p@@sbbllx#1{
		\@bbllxtrue
		\dimen100=#1
		\edef\@p@sbbllx{\number\dimen100}
}
\def\@p@@sbblly#1{
		\@bbllytrue
		\dimen100=#1
		\edef\@p@sbblly{\number\dimen100}
}
\def\@p@@sbburx#1{
		\@bburxtrue
		\dimen100=#1
		\edef\@p@sbburx{\number\dimen100}
}
\def\@p@@sbbury#1{
		\@bburytrue
		\dimen100=#1
		\edef\@p@sbbury{\number\dimen100}
}
\def\@p@@sheight#1{
		\@heighttrue
		\dimen100=#1
   		\edef\@p@sheight{\number\dimen100}
}
\def\@p@@swidth#1{
		\@widthtrue
		\dimen100=#1
		\edef\@p@swidth{\number\dimen100}
}
\def\@p@@srheight#1{
		\@rheighttrue
		\dimen100=#1
		\edef\@p@srheight{\number\dimen100}
}
\def\@p@@srwidth#1{
		\@rwidthtrue
		\dimen100=#1
		\edef\@p@srwidth{\number\dimen100}
}
\def\@p@@sprolog#1{\@prologfiletrue\def\@prologfileval{#1}}
\def\@p@@spostlog#1{\@postlogfiletrue\def\@postlogfileval{#1}}
\def\@cs@name#1{\csname #1\endcsname}
\def\@setparms#1=#2,{\@cs@name{@p@@s#1}{#2}}
\def\ps@init@parms{
		\@bbllxfalse \@bbllyfalse
		\@bburxfalse \@bburyfalse
		\@heightfalse \@widthfalse
		\@rheightfalse \@rwidthfalse
		\def\@p@sbbllx{}\def\@p@sbblly{}
		\def\@p@sbburx{}\def\@p@sbbury{}
		\def\@p@sheight{}\def\@p@swidth{}
		\def\@p@srheight{}\def\@p@srwidth{}
		\def\@p@sfile{}
		\def\@p@scost{10}
		\def\@sc{}
		\@prologfilefalse
		\@postlogfilefalse
		\@clipfalse
}
\def\parse@ps@parms#1{
	 	\@psdo\@psfiga:=#1\do
		   {\expandafter\@setparms\@psfiga,}}
\newif\ifno@bb
\newif\ifnot@eof
\newread\ps@stream
\def\bb@missing{
	\typeout{psfig: searching \@p@sfile \space  for bounding box}
	\openin\ps@stream=\@p@sfile
	\no@bbtrue
	\not@eoftrue
	\catcode`\%=12
	\loop
		\read\ps@stream to \line@in
		\global\toks200=\expandafter{\line@in}
		\ifeof\ps@stream \not@eoffalse \fi
		\@bbtest{\toks200}
		\if@bbmatch\not@eoffalse\expandafter\bb@cull\the\toks200\fi
	\ifnot@eof \repeat
	\catcode`\%=14
}	
\catcode`\%=12
\newif\if@bbmatch
\def\@bbtest#1{\expandafter\@a@\the#1
\long\def\@a@#1
\long\def\bb@cull#1 #2 #3 #4 #5 {
	\dimen100=#2 bp\edef\@p@sbbllx{\number\dimen100}
	\dimen100=#3 bp\edef\@p@sbblly{\number\dimen100}
	\dimen100=#4 bp\edef\@p@sbburx{\number\dimen100}
	\dimen100=#5 bp\edef\@p@sbbury{\number\dimen100}
	\no@bbfalse
}
\catcode`\%=14
\def\compute@bb{
		\no@bbfalse
		\if@bbllx \else \no@bbtrue \fi
		\if@bblly \else \no@bbtrue \fi
		\if@bburx \else \no@bbtrue \fi
		\if@bbury \else \no@bbtrue \fi
		\ifno@bb \bb@missing \fi
		\ifno@bb \typeout{FATAL ERROR: no bb supplied or found}
			\no-bb-error
		\fi
		\count203=\@p@sbburx
		\count204=\@p@sbbury
		\advance\count203 by -\@p@sbbllx
		\advance\count204 by -\@p@sbblly
		\edef\@bbw{\number\count203}
		\edef\@bbh{\number\count204}
}
\def\in@hundreds#1#2#3{\count240=#2 \count241=#3
		     \count100=\count240	
		     \divide\count100 by \count241
		     \count101=\count100
		     \multiply\count101 by \count241
		     \advance\count240 by -\count101
		     \multiply\count240 by 10
		     \count101=\count240	
		     \divide\count101 by \count241
		     \count102=\count101
		     \multiply\count102 by \count241
		     \advance\count240 by -\count102
		     \multiply\count240 by 10
		     \count102=\count240	
		     \divide\count102 by \count241
		     \count200=#1\count205=0
		     \count201=\count200
			\multiply\count201 by \count100
		 	\advance\count205 by \count201
		     \count201=\count200
			\divide\count201 by 10
			\multiply\count201 by \count101
			\advance\count205 by \count201
		     \count201=\count200
			\divide\count201 by 100
			\multiply\count201 by \count102
			\advance\count205 by \count201
		     \edef\@result{\number\count205}
}
\def\compute@wfromh{
		\in@hundreds{\@p@sheight}{\@bbw}{\@bbh}
		\edef\@p@swidth{\@result}
}
\def\compute@hfromw{
		\in@hundreds{\@p@swidth}{\@bbh}{\@bbw}
		\edef\@p@sheight{\@result}
}
\def\compute@handw{
		\if@height 
			\if@width
			\else
				\compute@wfromh
			\fi
		\else 
			\if@width
				\compute@hfromw
			\else
				\edef\@p@sheight{\@bbh}
				\edef\@p@swidth{\@bbw}
			\fi
		\fi
}
\def\compute@resv{
		\if@rheight \else \edef\@p@srheight{\@p@sheight} \fi
		\if@rwidth \else \edef\@p@srwidth{\@p@swidth} \fi
}
\def\compute@sizes{
	\compute@bb
	\compute@handw
	\compute@resv
}
\def\psfig#1{\vbox {
	\ps@init@parms
	\parse@ps@parms{#1}
	\compute@sizes
	\ifnum\@p@scost<\@psdraft{
		\typeout{psfig: including \@p@sfile \space }
		\special{ps::[begin] 	\@p@swidth \space \@p@sheight \space
				\@p@sbbllx \space \@p@sbblly \space
				\@p@sbburx \space \@p@sbbury \space
				startTexFig \space }
		\if@clip{
			\typeout{(clip)}
			\special{ps:: \@p@sbbllx \space \@p@sbblly \space
				\@p@sbburx \space \@p@sbbury \space
				doclip \space }
		}\fi
		\if@prologfile
		    \special{ps: plotfile \@prologfileval \space } \fi
		\special{ps: plotfile \@p@sfile \space }
		\if@postlogfile
		    \special{ps: plotfile \@postlogfileval \space } \fi
		\special{ps::[end] endTexFig \space }
		\vbox to \@p@srheight true sp{
			\hbox to \@p@srwidth true sp{
				\hfil
			}
		\vfil
		}
	}\else{
		\vbox to \@p@srheight true sp{
		\vss
			\hbox to \@p@srwidth true sp{
				\hss
				\@p@sfile
				\hss
			}
		\vss
		}
	}\fi
}}
\catcode`\@=12\relax

\newenvironment{code}{\samepage \vskip 0.1in \tt\nofill}{\endnofill \vskip 0.1in}

\catcode`\@=11\relax         

\def\nofill {%
  \begingroup
    \obeyspaces
    \parskip=\z@
    \parindent=\z@
    \let\p@r=\par
    \def\par{\p@r \ifspacenext{\noindent}{}}%
    \obeylines}

\def\endnofill{
  \endgroup}

{\obeyspaces\global\let\sp@ce= \relax}

\def\ifspacenext #1#2{%
  \def\truet@ks{#1}%
  \def\falset@ks{#2}%
  \futurelet\next\ifsp@cenext}
\def\ifsp@cenext {%
  \ifx\next\sp@ce \truet@ks \else \falset@ks \fi}

\catcode`\@=12\relax    

\newcommand{\mycap}[1]{\caption{#1}}
\def\bo1{\mbox{BO-}1}
\def\boo1{\mbox{BO-o}1}
\def\MLE1{\mbox{MLE-}1}
\def\MLEo1{\mbox{MLE-o}1}
\def\rand{\mbox{RAND}}
\def\conf{P_{{\sc{C}}}}
\def\confarg{\conf(\obj' | \obj)}
\def\pone{q}
\def\ptwo{r}
\def\pthree{s}
\def\alph{{\cal Y}}
\def\x{y}
\def\X{Y}
\def\typex{P_{\bar{\x}}}
\def\Both{\mbox{\em Both}}
\def\both{\mbox{\small {\em Both}}}
\def\Poneonly{\mbox{\em Just\pone}}
\def\poneonly{\mbox{\small {\em Just\pone}}}
\def\Ptwoonly{\mbox{\em Just\ptwo}}
\def\ptwoonly{\mbox{\small {\em Just\ptwo}}}
\def\Lone{L_1}
\def\vhref{Hatzivassiloglou+McKeown:93a}
\def\vh{\mbox{H\&M}}
\def\defas{=}
\def\contexts{\mbox{$\alph$}}
\def\objects{{\cal X}}
\def\obj{x}
\def\context{\x}
\def\counts{C}
\def\pmle{P_{{\sc MLE}}}
\def\backoff{back-off}
\def\pbo{P_{\sc BO}}
\def\clusters{{\cal C}}
\def\mp{\tilde{P}}
\def\pmem{\mp(c|\obj)}
\def\pcen{\mp(\context|c)}
\def\pmemp{\mp(c|\obj)}
\def\pcenp{\mp(\context'|c')}
\def\vocab{{\cal W}}
\def\word{w}
\def\psim{P_{{\sc SIM}}}
\def\chap{chapter }
\def\chaps{chapters }

\newcommand{\delmemp}[1]{\frac{\partial #1}{\partial\pmem}}
\newcommand{\delmempslash}[1]{\partial #1/\partial\pmem}
\newcommand{\delpcen}[1]{\frac{\partial~~~}{\partial\pcen} #1}
\newcommand{\delpcenshort}[1]{\frac{\partial #1}{\partial\pcen}}
\newcommand{\delpcenp}[1]{\frac{\partial~~}{\partial\pcenp} #1}
\newcommand{\smrm}[1]{\mbox{\scriptsize #1}}
\newcommand{\eqpunc}[1]{{\makebox[0pt][l]{\qquad\rm{#1}}}}
\newcommand{\constrained}[1]{{#1}^+}
\newcommand{\class}[1]{c(#1)}

\newcommand{\epsfscaledbox}[2]{\psfig{figure=#1,width=#2}}
\newcommand{\set}[1]{\{#1\}}

\newcommand{\proof}[1]{
{\noindent {\it Proof.} {#1} \rule{2mm}{3mm}} }

\documentclass{report}
\usepackage{thesis}

\setstretch{1} 
\renewcommand{\arraystretch}{1}

\begin{document}

\bibliographystyle{fullname}
\pagenumbering{roman}

\thispagestyle{empty}

\mbox{}
\vfill

\vspace{-300pt}
{\normalsize \tt \hfill Harvard University Technical Report TR-11-97}
\\ \mbox{}\\

\vspace{1.5in}

{\large
\begin{center}
{\bf \LARGE Similarity-Based Approaches to Natural Language Processing} \\
\mbox{} \\
A thesis presented \\
by \\
\bigskip
{\bf \Large Lillian Jane Lee} \\
\bigskip
to \\
The Division of Engineering and Applied Sciences \\
in partial fulfillment of the requirements \\
for the degree of \\
Doctor of Philosophy \\
in the subject of \\
Computer Science \\
\mbox{} \\
Harvard University \\
Cambridge, Massachusetts \\
\mbox{} \\
May 1997
\end{center}
}

\vfill
\newpage

\mbox{}
\vfill

\begin{center}
\copyright 1997 by Lillian Jane Lee\\
All rights reserved.
\end{center}

\vfill

\chapter*{Abstract}

Statistical methods for automatically extracting information about
associations between words or documents from large collections of text
have the potential to have considerable impact in a number of areas,
such as information retrieval and natural-language-based user
interfaces.  However, even huge bodies of text yield highly unreliable
estimates of the probability of relatively common events, and, in
fact, perfectly reasonable events may not occur in the training data
at all.  This is known as the {\em sparse data problem}.  Traditional
approaches to the sparse data problem use crude approximations.  We
propose a different solution: if we are able to organize the data into
classes of similar events, then, if information about an event is
lacking, we can estimate its behavior from information about similar
events.  This thesis presents two such similarity-based approaches,
where, in general, we measure similarity by the Kullback-Leibler
divergence, an information-theoretic quantity.

Our first approach is to build soft, hierarchical clusters: soft,
because each event belongs to each cluster with some probability;
hierarchical, because cluster centroids are iteratively split to model
finer distinctions.  Our clustering method, which uses the technique
of deterministic annealing, represents (to our knowledge) the first
application of soft clustering to problems in natural language
processing.  We use this method to cluster words drawn from 44
million words of Associated Press Newswire and 10 million words from
Grolier's encyclopedia, and find that language models built from the
clusters have substantial predictive power.  Our algorithm also extends
with no modification to other domains, such as document clustering.

Our second approach is a nearest-neighbor approach: instead of
calculating a centroid for each class, we in essence build a cluster
around each word.  We compare several such nearest-neighbor approaches
on a word sense disambiguation task and find that as a whole, their
performance is far superior to that of standard methods.  In another
set of experiments, we show that using
estimation techniques based on the nearest-neighbor model enables us
to achieve perplexity reductions of more than 20 percent over standard
techniques in the prediction of low-frequency events, and
statistically significant speech recognition error-rate reduction.

\chapter*{Acknowledgements}

In my four years as a graduate student at Harvard, I have noticed a
rather ominous trend.  Aiken Computation Lab, where I spent so many
days and nights, is currently slated for demolition.  I worked at AT\&T Bell
Labs for several summers -- and now this institution no longer exists.
I finally began to catch on when, after a rash of repetitive strain
injury cases broke out at Harvard, one of my fellow graduate students
called the Center for Disease Control to suggest that I might be the
cause.  

All this aside, I feel that I have
been incredibly fortunate.  First of all, I have had the Dream Team of
NLP for my committee.  Stuart Shieber, my advisor, has been absolutely
terrific.  The best way to sum up my interactions with him is: he
never let me get away with anything.  The stuff I've produced has been
the clearer and the better for it.  Barbara Grosz has been wonderfully
supportive; she also throws a mean brunch.  And there is just no way I
can thank Fernando Pereira enough.  He is the one who started me off
on this research enterprise in the first place, and truly deserves his
title of mentor.

I'd also like to thank Harry Lewis for the whole CS 121 experience,
Les Valiant and Alan Yuille for being on my oral exam committee, and
Margo Seltzer for advice and encouragement.

There have been a number of people who made the grad school
process much, much easier to deal with: Mike (``at least you're not in
medieval history'') Bailey, Ellie Baker, 
Michael Bender, Alan Capil, Stan Chen (the one and only), Ric Crabbe, Adam
Deaton, Joshua Goodman (for the colon), Carol Harlow, Fatima
Holowinsky, Bree Horwitz,
Andy Kehler, Anne Kist,
Bobby Kleinberg, 
David Mazieres, Jeff Miller, Christine Nakatani, 
Wheeler Ruml, Kathy Ryall (my job buddy), Ben Scarlet, Rocco Servedio
(for the citroids),
Jen Smith, Nadia Shalaby, Carol
Sandstrom and Chris Small (and Emily and Sophie, the best
listeners in the entire universe), Peg
Schafer, 
 Keith Smith, Chris Thorpe, and Tony Yan.

AT\&T was a wonderful place to work, and I am quite grateful for
having had the opportunity to talk with all of the following people:
Hiyan Alshawi, Ido Dagan, Don Hindle, Julia Hirschberg, Yoram Singer,
Tali Tishby, and David Yarowsky.  I'd also like to mention the ``s
underscores'' -- the summer students at AT\&T who made it so difficult
to get anything done: David Ahn, Tom Chou, Tabora Constantennia,
Charles Isbell, Rajj Iyer, Kim Knowles,
Nadya Mason, Leah McKissic, Diana Meadows, Andrew Ng, Marianne Shaw, and
Ben Slusky.

Rebecca Hwa deserves special mention.  She was always there to say
``Now just calm down, Lillian'' whenever I got stressed out.  I thank
her for all those trips to Toscanini's, and forgive her for trying to
make me break my Aiken rule.  Maybe one day
I'll even get around to giving her her \LaTeX\ book back.

And finally, deepest thanks to my mom and dad, my sister Charlotte,
and Jon Kleinberg; they always believed, even when I didn't.

\medskip

The work described in this thesis was supported in part by the
National Science Foundation under Grant No. IRI-9350192.  I also
gratefully acknowledge support from an NSF Graduate Fellowship
and an grant from the AT\&T Labs Fellowship Program (formerly the
AT\&T Bell Labs Graduate Research Program for Women).

\section*{Bibliographic Notes}

Portions of this thesis are joint work and have appeared elsewhere.  

Chapter
\ref{ch:clust} is based on the paper ``Distributional Clustering of
English Words'' with Fernando Pereira and Naftali Tishby
\cite{Pereira+Tishby+Lee:93}, which appeared in the proceedings of the
31st meeting of the ACL.  We thank Don Hindle for making available the
1988 Associated Press verb-object data set, the Fidditch parser, and a
verb-object structure filter; Mats Rooth for selecting the data set
consisting of objects of ``fire'' and for many discussions; David
Yarowsky for help with his stemming and concordancing tools; and Ido
Dagan for suggesting ways to test cluster models.

Chapter \ref{ch:wsd} and portions of chapter \ref{ch:sim} are adapted
from ``Similarity-Based Methods for Word Sense Disambiguation''.  This
paper, co-written with Ido Dagan and Fernando Pereira, will appear in
the proceedings of the 35th meeting of the ACL
\cite{Dagan+Lee+Pereira:comp}.  We thank Hiyan Alshawi, Joshua Goodman, Rebecca Hwa,
Stuart Shieber, and Yoram Singer for many helpful comments and
discussions.

Chapter \ref{ch:pp} is based on work with Ido Dagan and Fernando
Pereira that is described in
 ``Similarity-Based Estimation of Word Cooccurrence
Probabilities'', which appeared in
the proceedings of the 32nd Annual meeting of the ACL \cite{Dagan+Pereira+Lee:94a}.
We thank Slava Katz for discussions on the topic of this paper, Doug
McIlroy for detailed comments, Doug Paul for help with his baseline
back-off model, and Andre Ljolje and Michael Riley for providing the
word lattices for our experiments.

\tableofcontents
\listoffigures
\listoftables

\pagenumbering{arabic}
\chapter{Introduction}
\label{ch:intro}
\begin{quote}
``You shall know a word by the company it keeps!''  \cite[pg. 11]{Firth:57a}  
\end{quote}

We begin by considering the problem of predicting string
probabilities.  Suppose we are presented with two strings,
\begin{enumerate}
\item ``Grill doctoral candidates'', and
\item ``Grill doctoral updates'',
\end{enumerate}
and are asked to determine which string is more likely.
Notice that this is {\em not} the same question as asking which of these
strings is grammatical.  In fact, both constitute legitimate English
sentences.  The first sentence is a command to ask a graduating
Ph.D. student many difficult questions; the second might be an order
to take lists of people who have just received doctorates and throw
them\footnote{(the lists)} on a Hibachi.  

Methods for assigning probabilities to strings are called {\em
language models}.  In this thesis, we will abuse the term somewhat and
refer to methods that assign probabilities to word associations as
language models, too.  That is, we will consider methods which
estimate the probability of word cooccurrence relations; these methods
need not be defined on sentences.  For example, in chapters
\ref{ch:clust} and \ref{ch:wsd} we will be concerned with the problem of
estimating the probability that a noun $\obj$ and a transitive verb
$\context$ appear in a sentence with $\obj$ being the head noun of the
direct object of $\context$.  

One important application of language modeling is error correction.
Current speech recognizers do not achieve perfect recognition rates,
and it is easy to imagine a situation in which a speech recognizer
cannot decide whether a speaker said ``Grill doctoral candidates'' or
``Grill doctoral updates''.  A language model can provide a speech
recognizer with the information that the former sentence is more
likely than the latter; this information would help the recognizer
make the right choice.  Similar situations arise in handwriting
recognition, spelling correction, optical character recognition, and
so on --- whenever the physical evidence itself may not be enough to
determine the corresponding string.

More formally, let $E$ be some physical evidence, and suppose we wish
to know whether the string $W$ is the message conveyed or encoded by
$E$.  Using Bayes' rule, we can combine the estimate $P(E|W)$ given by
an acoustic model with the probability $P_{{\sc LM}}(W)$ assigned by a
language model to find the posterior probability that $W$ is the true
string given the evidence at hand:
\begin{equation}
P(W|E) = \frac{P_{{\sc LM}}(W)P(E|W)}{P(E)}
\label{acoustic}
\end{equation}
(since the evidence $E$ is fixed, it is the same for every
hypothesized string $W$, so the $P(E)$ term is generally ignored in
practice).  Thus, in a situation where two hypothesized strings cannot
be distinguished on the basis of the physical evidence alone, a
language model can provide the information necessary for disambiguation.

Another application of language modeling is machine translation.
Suppose one needs to translate the phrase ``Grill doctoral
candidates'' to another language.  Two possible target sentences are
``Ask applicants many questions'' and ``Roast applicants on a spit''.
If we have a language model that furnishes us the information that the
first sentence is more likely than the second, then (in the absence of
context providing evidence to the contrary) we would pick the first
sentence as the correct translation.

This thesis is concerned with {\em statistical} approaches to problems
in natural language processing.  Typically, statistical approaches
take as input some large sample of text, which may or may not be
annotated in some fashion, and attempt to learn characteristics of the
language from the statistics in the sample.  They may also make use of
auxiliary information gained from such sources as on-line dictionaries
or WordNet \cite{Miller:95a}.  An important advantage of statistical
approaches over traditional linguistic models is that statistical
methods yield probabilities.  These probabilities can easily be
combined with estimates from other components, as in equation
(\ref{acoustic}) above.  Traditional linguistic models, on the other
hand, only describe whether or not a string is grammatical.  This
information is too coarse-grained for use in practical tasks; for
instance, both ``Grill doctoral candidates'' and ``Grill doctoral
updates'' are valid sentences, and yet we know that the first string
is far more likely than the second.

Perhaps the simplest statistical approach to language modeling is the
maximum likelihood estimate (MLE), which simply counts the number of
times that the string of interest occurs in the training sample $S$ and
normalizes by the  sample size.  For ``Grill doctoral
candidates'', this estimate takes the form
\begin{equation}
\pmle(\mbox{``Grill doctoral candidates''}) = \frac{\counts(\mbox{``Grill doctoral candidates'')}}{|S|},
\end{equation}
where $\counts(\mbox{``Grill doctoral candidates''})$ is the number of
times the phrase occurred in $S$.  The quantity $|S|$ might be the
number of word triples in $S$, or the number of sentences in $S$, or
some other relevant measure.

Notice that if the event of interest is {\em unseen}, that is, does
not occur in $S$, then the maximum likelihood estimate assigns it a
probability of zero.  In terms of practicality, this turns out to be a
fatal flaw because of the {\em sparse data} problem: even if $S$ is
quite big, a large number of possible events will not appear in $S$.
Assigning all unseen events a probability of zero, as the MLE does,
amounts to declaring many perfectly reasonable strings to have zero
probability of occurring, which is clearly unsatisfactory.

To illustrate the pernicious nature of the sparse data problem, we
present the following example.  Consider the set $S$ to be the text
contained in all the pages indexed by AltaVista, Digital's web search
engine \cite{altavista}.  Currently, this set consists of 31
million web pages, which, at an extremely conservative estimate,
means that $S$ contains at least a billion words.  Yet at the time of
this writing, the phrase ``Grill doctoral candidates'' does not occur
at all among those billion words, so  the MLE would rule out
this sentence as absolutely impossible.  

Although the sparse data problem affects low-frequency events, it is
incorrect to infer that it therefore is not important.  One might
attempt to claim that if an event has such a low probability that it
does not occur in a very large sample, then actually estimating its
probability to be zero will not be a major error.  However, the
aggregate probability of unseen events can be a big percentage of the
test data, which means that it is quite important to treat unseen
events carefully.  Brown et al. \shortcite{Brown+al:class}, for
instance, studied a 350 million word sample of English text, and
estimated that in any new sample drawn from the same source
distribution, 14\% of the trigrams (sequences of three consecutive
words) would not have occurred in the large text.  A speech recognizer
that refused to accept 1 out of every 7 sentences would be completely
unusable.

As a historical aside, we observe that Noam Chomsky famously declared
that sparse data problems are insurmountable.
\begin{quote}
It is fair to assume that neither sentence (1) [Colorless green
ideas sleep furiously] nor (2) [Furiously sleep ideas green
colorless] ... has ever occurred .... Hence, in any statistical model
... these sentences will be ruled out on identical grounds as equally
`remote' from English. Yet (1), though nonsensical, is grammatical,
while (2) is not.''\footnote{Ironically, this remark is now so
well known that it has become false:  use of AltaVista reveals that at the
time of this writing, 40 web pages contain the first sentence, whereas
only three contain the second.}  \cite[pg. 16]{Chomsky:57a} 
\end{quote}
This thought experiment helped ``[disabuse] the field once and for all
of the notion that there was anything of interest to statistical
models of language'' \cite{Abney:96a}.  

However, in the years since Chomsky wrote this remark, some progress
on ameliorating the sparse data problem has been made.  Indeed,
Chomsky's statement is based on the false assumption that ``any
statistical model'' must be based on the maximum likelihood
estimate. This is certainly not the case.
Two standard language modeling techniques used in speech recognition,
Jelinek-Mercer smoothing and Katz
\backoff\, smoothing, make use of an estimator guaranteed to be
non-zero.  In the case where the probability of an unseen word pair
$(w_1,w_2)$ is being estimated, these methods incorporate the
probability of word $w_2$ (details can be found in section
\ref{sec:smoothing}).  But this is not always adequate: for example,
the word ``updates'' appears on more web pages indexed by AltaVista
than ``candidates'' does.

The key idea in this thesis is that we can use {\em
similarity} information to make more sophisticated
probability estimates when sparse data problems occur.  This idea is
intuitively appealing, for if we know that the word ``candidates'' is
somehow similar to the word ``nominees'', then the occurrence of the
sentence ``Grill doctoral nominees'' would lead us to believe that
``Grill doctoral candidates'' is also likely.

The notion of similarity we explore is that of {\em distributional
similarity}, since we will represent words as distributions over the
contexts in which they occur (as implied by the quotation that opens
this chapter).  We will thus be concerned with measures of the
``distance'' between probability mass functions.  We discuss several
such measures in chapter \ref{ch:sim}, but our main focus will be on
using the Kullback-Leibler divergence, an information-theoretic
quantity.

The work presented in this thesis can be divided into two parts.  The
first is the development of a {\em distributional clustering} method
for grouping similar words.  This method builds probabilistic,
hierarchical clusters: objects belong to each cluster with some
probability, and clusters are broken up into subclusters so that a
hierarchy results.  We derive the method, exhibit clusters found by
our method in order to provide a qualitative sense of how the method
performs, and show that effective language models can be constructed
from the clusters produced by our method.  To our knowledge, this is
the first probabilistic clustering method to be applied to natural language processing.

The second part is the development of a more computationally efficient
way to take incorporate similarity information: a nearest-neighbor (or
``most similar neighbor'') language model that combines estimates from
specific objects rather than from classes.  We compare several
different implementations of this type of model against standard
smoothing methods and find that using similarity information leads to far
better estimates.

This thesis is organized as follows.  Chapter \ref{ch:sim}
describes many of the theoretical results employed in later
chapters.  We discuss standard language modeling techniques and study
the properties of several distributional similarity functions.  Chapter
\ref{ch:clust} presents our distributional clustering method.
Chapter \ref{ch:wsd} develops the nearest-neighbor
approach and compares the performance of several implementations on a
pseudo-word-disambiguation task.  Chapter \ref{ch:pp} considers an
extension of our nearest-neighbor approach and studies its performance
on more realistic tasks.  We conclude with a brief summary of the
thesis and indicate directions for further work in chapter
\ref{ch:concl}.

\setcounter{chapter}{1}

\chapter{Distributional Similarity}
\label{ch:sim}

\newtheorem{theorem}{Theorem}[chapter]

This chapter presents background material underlying the work in
this thesis.  In section \ref{sec:representation} we argue that
representing objects as distributions is natural and useful.  Section
\ref{sec:smoothing} reviews common methods for estimating
distributions from a sample.  We use these methods to
provide initial distributions to our algorithms and also as
standards against which to compare the performance of our
similarity-based estimates.  Section \ref{sec:measures} studies
various functions measuring similarity between distributions.  We pay
particular attention to the Kullback-Leibler divergence
\cite{Cover+Thomas:91a}, which plays a central role in our work.


\section{Objects as Distributions}
\label{sec:representation}

The first issue we must address is what representation to use for
the objects we wish to cluster and compare.  
For the moment, we will be vague about what sorts of objects we will
be considering; researchers have clustered everything from documents
\cite{Salton:68a,Cutting+al:92a} to irises
\cite{Fisher:36a,Cheeseman+al:88a}.

We want the representation we choose to satisfy two requirements.
First, the representation should be general enough to apply
to many different types of objects.  Second, 
any particular object's representation should be easy
to calculate from samples alone; we do not want to use outside sources of
information such as on-line dictionaries. 
This second condition expresses our preference for algorithms that
are adaptable; if we rely on knowledge that is hard for computers to
derive from training data, then we cannot use our
algorithms on new domains without expending considerable effort on
re-acquiring the requisite knowledge.  Furthermore, large samples that
have few or no annotations\footnote{In the following chapters, we use
either unannotated data or data that has been tagged with parts of
speech.}  are far more common and readily obtainable than large
highly-annotated samples, and thus working with representations
adhering to the second condition tends to be much more convenient.

Many clustering schemes represent objects in terms of a set $\set{A_1,
A_2, \ldots, A_N}$ of attributes
\cite{Kaufman+Rousseeuw:90}.  Each object is associated
with an attribute vector $(a_1, a_2, \ldots, a_N)$ of values for the
attributes.  Some attributes can take on an infinite
number of values; for example, the mean of a normal distribution can
be any real number. Other attributes, such as the sex of a patient, range
only over a finite set.  Usually, no assumptions are made about the
relationship between different attributes.

In this thesis, we use a restricted version of the attribute
representation.  Objects are equated with probability mass functions,
or distributions: each attribute must have a nonnegative real value, and
we require that all object attribute vectors $(a_1, a_2, \ldots, a_N)$
satisfy the constraint that $\sum_{i=1}^N a_i = 1$.  We can think of $a_i$ as
the probability that the object assigns to $A_i$.  This distributional
representation for objects is particularly appropriate for situations
arising in unsupervised learning, where a learning algorithm must
infer properties of events from a sample of unannotated data.  In such
situations, we can define the attributes to be the contexts in which
events can occur; the value $a_i$ for a particular
event is then the proportion of the time the event occurred in context
$i$.  

For example, suppose we wish to learn about word usage from the
following (small) sample of English text: ``A rose is a rose is not a
nose''.  Our events are therefore words.  If we define the context of
a word to be the following word, then the possible contexts are ``a'',
``is'',``nose'', ``not'', and ``rose''.  The attribute vector for the
word ``a'' is $(0, 0, 1/3,0,2/3)$, since ``a'' occurs before ``nose''
(the third attribute) one out of three times, and before ``rose'' (the
last attribute) twice.

In accordance with our first requirement for representations, the
distributional representation is fairly general.  For instance, we
have demonstrated that words can be represented as distributions over
subsequent words, and we can just as easily represent documents as
distributions over the words that occur in them, or customers as
distributions over the products they buy, and so on.  Indeed, it is a
reasonable representation whenever the data consists of a set of
events (e.g., words occurring together) rather than measurements or
properties (e.g., a list of each word's part of speech).  Also,
in compliance with our second requirement for distributions, the
distributional representation for any object is trivial to calculate
as long as the contexts are easily recognizable.  Furthermore, we wish
to apply our techniques to language modeling, a task for which
probability distributions must be produced.  Finally, the constraint
that the components of attribute vectors sum to unity is of use to us
in our calculations, as will be seen in chapter
\ref{ch:clust}.


\section{Initial Estimates for Distributions}
\label{sec:smoothing}

The remainder of this thesis will be concerned with object
distributions that have been estimated from object-context pairs.
More formally, let $\objects$ be the set of objects under
consideration and $\contexts$ be the set of possible contexts,
$\contexts = \set{\context_1, \context_2, \ldots, \context_N}$.
Assume that the data consists of pairs $(\obj,\context) \in \objects
\times \contexts$ along with counts $\counts(\obj,\context)$ of how many
times $(\obj,\context)$ occurred in some training sample.  Counts for
individual objects and contexts  are readily attained from counts
for the pairs:  $\counts(\obj) = \sum_{\context}
\counts(\obj,\context)$ and $\counts(\context) = \sum_{\obj}
\counts(\obj,\context)$;  without loss of
generality, assume that every object and every context occurs at least
once.  We wish to represent object $\obj$ by the conditional
distribution $P(\context|\obj)$ all $\context \in \contexts$. This
distribution must be estimated from the data pairs.  Of course, the
goal of this thesis is to develop good estimates for $P(\context
|\obj)$, but we need some initial distributions to start with.

A particularly simple estimation method is the {\em maximum likelihood
estimate} (MLE) $\pmle(\context| \obj)$:
\begin{equation}
\pmle(\context |\obj) = \frac{\counts(\obj,\context)}{\counts(\obj)}.
\label{MLE-general}
\end{equation}
Notice that if the joint event $(\obj,\context)$ never occurs, then
$P_{MLE}(\context| \obj) = 0$, which is equivalent to saying that any
event that does not occur in the training sample is impossible.  As
noted in \chap \ref{ch:intro}, using the maximum
likelihood estimate tends to grossly underestimate the
probability of low-frequency events.

Many alternatives to the MLE
\cite{Good:53a,Jelinek:80a,Katz:87a,Church+Gale:91a} take the MLE as
an initial estimate and adjust it so that the total estimated
probability of pairs occurring in the sample is less than one, leaving
some probability mass for unseen pairs. These techniques are known as
{\em smoothing} methods, since they ``smooth over'' zeroes in
distributions.  Typically, the adjustment
involves either {\em interpolation}, in which the new estimator is a
weighted combination of the MLE and an estimator  guaranteed to
be nonzero for unseen pairs, or {\em discounting}, in which the MLE is
decreased to create some leftover probability mass for unseen pairs.

The work of \namecite{Jelinek:80a} is the classic interpolation
method.  They produce an estimate by linearly interpolating the MLE
for the conditional probability of an object-context pair $(\obj,\context)$
with the maximum likelihood estimate $\pmle(\context) =
\counts(\context)/\sum_{\context} \counts(\context)$  for the probability of
context $y$:
\begin{equation}
P_{{\sc JM}}(\context | \obj) =  \lambda(\obj) \pmle(\context | \obj)
+ (1 -  \lambda(\obj)) \pmle(\context).
\label{interp}
\end{equation}
The function $\lambda(\obj)$ ranges between 0 and 1,  
and reflects our confidence in the available data
regarding $\obj$.  
If $\obj$ occurs relatively frequently, then we have reason to believe
that the MLE for the pair $(\obj,\context)$ is reliable.  We then 
give
$\lambda(\obj)$ a high value,  so that $P_{{\sc JM}}$ depends mostly on 
$\pmle(\context | \obj)$.  On the other hand, if $\obj$ is relatively rare,
then $\pmle(\context | \obj)$ is unlikely to be very accurate.  In this case, we
decide to rely more on $\pmle(\context)$, since counts for the
single event $\context$ are higher than counts for the
joint event ($\obj,\context$).  We therefore set $\lambda(\obj)$
to a relatively  low value.
A method for
training $\lambda$ is described by
\namecite{Bahl:83a}.  

A popular alternative in the speech recognition literature is the
\backoff\, discounting method of \namecite{Katz:87a}.  It provides a clear
separation between frequent events, for which observed frequencies are
reliable probability estimators, and low-frequency events, whose
prediction must involve additional information sources. Furthermore,
the \backoff\, model does not require complex estimation calculations
for interpolation parameters such as $\lambda(x)$, as is the case for
Jelinek and Mercer's method.

Katz first uses the Good-Turing formula \cite{Good:53a}
to replace the actual
frequency $\counts(\obj,\context)$ of an object-context pair
with a discounted frequency $\counts^{*}(\obj,\context)$.  Let 
$n_{m}$ denote the number of pairs that occurred $m$ times in
the sample.  The Good-Turing estimate then defines $\counts^{*}(\obj,\context)$  as
\begin{displaymath}
\counts^{*}(\obj,\context) = (\counts(\obj,\context) + 1) \frac{n_{\counts(\obj,\context) + 1}}{n_{\counts(\obj,\context)}}.
\end{displaymath}
This discounted frequency is used in the same way the true
frequency is used in the MLE (equation (\ref{MLE-general})):
\begin{displaymath}
\label{gt}
P_{d}(\context|\obj) = \frac{\counts^*(\obj,\context)}{\counts(\obj)}.
\end{displaymath}
As a consequence, the estimated conditional probability of an unseen
pair $(\obj',\context')$ is 
\begin{displaymath}
P_{d}(\context'|\obj') =\frac{(n_1/n_0)}{\counts(\obj')}.
\end{displaymath}
Thus, the probability mass assigned to unseen
pairs involving object $\obj'$ is distributed 
uniformly. 
The total mass assigned to unseen pairs
involving  $\obj'$ is simply the complement of the mass assigned to
seen pairs involving $\obj'$:
\begin{displaymath}
\tilde{\beta}(\obj') = 1 - \sum_{\context:\counts(\obj',\context) > 0}
P_d(\context | \obj').
\end{displaymath}
For more details, see \namecite{Nadas:85a}, who
presents three different derivations (two empirical-Bayesian and one
empirical) of the Good-Turing estimate.

Katz alters the Good-Turing treatment by not using $P_{d}$ for unseen
pairs.  Rather, he bases his estimate of the conditional probability of an
unseen pair $(\obj',\context')$ on an 
estimate of the probability of $\context'$.  This amounts to 
assuming that the behavior of $(\obj',\context')$ is independent of
the behavior of $\obj'$; \namecite{Jelinek:80a} make a similar
assumption when they set $\lambda$ to a low
value in equation (\ref{interp}).  

More formally, we write the estimate
for an arbitrary pair $(\obj,\context)$ in
the following form, which is not Katz's original presentation but will
be convenient for us in chapters \ref{ch:wsd} and \ref{ch:pp} (note
the asymmetrical treatment of seen and unseen pairs):
\begin{equation}
\hat{P}(\context| \obj) = \left\{\!\!\!\!
\begin{array}{l@{\hspace{0.6ex}}l}
P_d(\context | \obj) & \mbox{if $\counts(\obj,\context) > 0$} \\
\alpha(\obj)P_r(\context | \obj) & \mbox{otherwise  ($(x,y)$ is unseen)}
\end{array}\right.\;.\label{genmodel}
\end{equation}
$P_r$ is the model for probability redistribution among 
unseen pairs. Katz (implicitly)
defines $P_r$ as the probability of the context:
\begin{displaymath}
P_r(\context|\obj) = P(\context), 
\label{ubo}
\end{displaymath}
so that 
\begin{equation}
\pbo(\context| \obj) = \left\{\!\!\!\!
\begin{array}{l@{\hspace{0.6ex}}l}
P_d(\context | \obj) & \mbox{if $\counts(\obj,\context) > 0$} \\
\alpha(\obj)P(\context) & \mbox{otherwise}
\end{array}\right.\;.\label{backoff}
\end{equation}
In later chapters, we will take advantage of the placeholder $P_r$ to
insert our own similarity-based probability redistribution models.
The quantity $\alpha(\obj)$ is a normalization factor required to ensure that 
$\sum_{\context}\pbo(\context| \obj) = 1$:
\begin{eqnarray*}
\alpha(\obj) & = &\frac{\tilde{\beta}(\obj)}{\sum_{\context:\counts(\obj,\context)= 0} P_r(\context | \obj)} \\
& = & \frac{\tilde{\beta}(\obj)}{1-\sum_{\context:\counts(\obj,\context)>0} P_r(\context | \obj)}.
\end{eqnarray*}
The second formulation of the normalization is computationally
preferable because it is generally the case that the total number of
possible pairs far exceeds the number of observed pairs.

Should we use Jelinek-Mercer smoothing, Katz \backoff\, smoothing, or
perhaps some other technique?  A thorough study by
\namecite{Chen+Goodman:96} showed that \backoff\, and Jelinek-Mercer
smoothing 
perform consistently well, with \backoff\, generally yielding better
results for modeling pairs.  Since the \backoff\, formulation also
contains a placeholder for us to apply similarity-based estimates, we
will use Katz's estimation method whenever smoothed distributions are required.

\section{Measures of Distributional Similarity}
\label{sec:measures}

In this section, we consider theoretical and computational properties
of several functions measuring the ``similarity'' between
distributions.  
We refer to these functions as distance functions,
rather than similarity functions, since most of them achieve their
{\em minimum} when the two distributions being compared are {\em
maximally} similar (i.e., identical).  The work described in chapters
\ref{ch:wsd} and \ref{ch:pp} 
uses negative exponentials of distance functions when true similarity
functions (that is, functions that increase as similarity increases)
are required.  

We certainly do not intend to give an exhaustive listing of all
distance functions. (See
\namecite{Anderberg:73a} for an extensive survey.) Our purpose is simply
to examine important properties of functions that we use or that are
commonly employed by other researchers in natural language processing
and machine learning.

We discuss the KL divergence in section
\ref{sec:KL} in detail, as it forms the basis for most of the work in
this thesis.  We also describe several other distance functions,
including the total divergence to the mean (section \ref{sec:KL+}),
various geometric norms (section \ref{sec:geometric}), and some
similarity statistics (section \ref{sec:statistics}).  We will pay
particular attention to the computational requirements of these
functions.  In view of the fact that we wish to use very large data
sets, we will require that the time needed to calculate the distance
between any two distributions be linear or near-linear in the number
of attributes.  This demand is not strictly necessary for the work
described in this thesis -- the clustering work of chapter
\ref{ch:clust} depends on the use of the KL divergence, and the
similarity computations of chapters \ref{ch:wsd} and \ref{ch:pp} are
done in a preprocessing phrase.  However, one of our future goals is
to find adaptive versions of our algorithms, in which case we must use
functions that can be computed efficiently.

We defer discussion of the {\em confusion probability}, defined by
\namecite{Essen+Steinbiss:92a}, until chapter \ref{ch:wsd}.  This
function is of great importance to us because Essen and Steinbiss's
{\em co-occurrence smoothing} method is quite similar to our own work
on language modeling.  The reason we do not include the confusion
probability in this chapter is that it is not a function of two
distributions :  each object $\obj$ is described both by the conditional
probability $P(\context| \obj)$ and the marginal probability
$P(\obj)$, so that comparing two objects involves four
distributions.

For the remainder of this section, let $\obj_1$, $\obj_2$, and
$\obj_3$ be three objects with associated distributions
$P(\cdot|\obj_1)$, $P(\cdot |\obj_2)$, and $ P(\cdot |
\obj_3)$, respectively.  It doesn't matter how these distributions
were estimated.  For notational convenience, we will call these
distributions $\pone$, $\ptwo$, and $\pthree$.  We will occasionally
refer to a distribution $p$ by its corresponding attribute vector
$(p(\x_1), p(\x_2),
\ldots, p(\x_N))$.

\subsection{KL Divergence}
\label{sec:KL}
  We define the function
 $D(\pone || \ptwo)$ as

\begin{equation}
D(\pone || \ptwo) \defas \sum_{\x \in \alph} \pone(\x) \log
\frac{\pone(\x)}{\ptwo(\x)}
\label{KL}
\end{equation}

\noindent (we will not specify the base of the logarithm).  Limiting
arguments lead us to set $0 \log \frac{0}{\ptwo} = 0$, even if $\ptwo
= 0$,
  and $\pone \log
\frac{\pone}{0} = \infty$ when $\pone$ is not zero.

Function (\ref{KL}) goes by
many names in the literature, including information gain
\cite{Renyi:70a}, error \cite{Kerridge:61a}, relative entropy, cross
entropy, and Kullback Leibler distance \cite{Cover+Thomas:91a}.
Kullback himself refers to the function as information for
discrimination, reserving the term ``divergence'' for the symmetric
function $D(\pone||\ptwo) + D(\ptwo||\pone)$ \cite{Kullback:59a}.  We
will use the name {\em Kullback-Leibler (KL) divergence} throughout
this thesis.

The KL divergence is a standard infor\-mation-theoretic ``measure'' of
the dissimilarity between two probability mass functions, and has been
applied to natural language processing (as described in this thesis),
machine learning, and statistical physics.  It is not a metric in the
technical sense, for it is not symmetric and does not obey the
triangle inequality (see, e.g., theorem 12.6.1 of
\namecite{Cover+Thomas:91a}).  However, it is non-negative, as shown
in the following theorem.

\begin{theorem}[Information inequality]  $D(\pone || \ptwo) \geq 0$, with equality holding if
and only if $\pone(\x) = \ptwo(\x)$ for all $\x \in \alph$.
\label{th:info-ineq}
\end{theorem}
\proof{
Most authors prove this theorem using {\em Jensen's inequality}, which
deals with expectations of convex functions (notice that $D(\pone
|| \ptwo)$ is the expected value with respect to $\pone$ of the quantity $\log
(\pone /\ptwo)$).
However, we present here a short proof attributed to Elizabeth Thompson
\cite{Green:96a}.

Let $\ln$ denote the natural logarithm, and let $b > 0$ be the base of
the logarithm in (\ref{KL}).  First observe that for any $z
\geq 0$, $\ln(z) \leq z - 1$, with equality holding if and only if $z
= 1$.  Then, we can write
\begin{eqnarray*}
-D(\pone || \ptwo) & = & \sum_{\x \in \alph} \pone(\x) \log_b
\frac{\ptwo(\x)}{\pone(\x)} \\
& = & \frac{1}{\ln(b)} \sum_{\x \in \alph} \pone(\x) \ln
\frac{\ptwo(\x)}{\pone(\x)} \\
& \leq & \frac{1}{\ln(b)} \sum_{\x \in \alph} \pone(\x) \left(
\frac{\ptwo(\x)}{\pone(\x)} - 1 \right) \\
& = & \frac{1}{\ln(b)} \left(\sum_{\x \in \alph} \ptwo(\x) - \sum_{\x \in \alph}
\pone(\x) \right) \\
& = &  \frac{1}{\ln(b)} ( 1 - 1) = 0,
\end{eqnarray*}
with equality holding if and only if $\frac{\ptwo(\x)}{\pone(\x)} = 1$
for all $\x \in \alph$.
} 

Since the KL divergence is $0$ when the two distributions are exactly
the same and greater than 0 otherwise, it is really a measure of
dissimilarity, as mentioned above, rather than similarity.  This
yields an intuitive explanation of why we should not expect the KL
divergence to obey the triangle inequality: as
\namecite{Hatzivassiloglou+McKeown:93a} observe, dissimilarity is not
transitive.

What motivates the use of the KL divergence, if it is not a true
distance metric?  We appeal to statistics,
information theory, and the maximum entropy principle.

The statistician \namecite{Kullback:59a} derives the KL divergence from a Bayesian
perspective.  Let $\X$ be a random variable taking values in $\alph$.
Suppose we are considering exactly two hypotheses about $\X$:  $H_\pone$ is the hypothesis that
$\X$ is distributed according to $\pone$, and
$H_\ptwo$ is the hypothesis that $\X$ is distributed according to 
$\ptwo$.  Using Bayes' rule, we can write the posterior probabilities
of the two hypotheses as
$$P(H_\pone | \x) = \frac{P(H_\pone) \pone(\x)}{P(H_\pone) \pone(\x) +
P(H_\ptwo) \ptwo(\x)},$$
and $$P(H_\ptwo | \x) = \frac{P(H_\ptwo) \ptwo(\x)}{P(H_\pone) \pone(\x) +
P(H_\ptwo) \ptwo(\x)}.$$
Taking logs of both equations and subtracting, we obtain
\begin{displaymath}
\log \frac{\pone(\x)}{\ptwo(\x)} = \log
\frac{P(H_\pone|\x)}{P(H_\ptwo|\x)} - \log
\frac{P(H_\pone)}{P(H_\ptwo)}.
\end{displaymath}
\noindent We can therefore consider $\log \left( \pone(\x) / \ptwo(\x)
\right)$ to be the information $\x$ supplies for choosing $H_\pone$
over $H_\ptwo$:  it is the difference between the logarithms of the
posterior odds ratio and the prior odds ratio. $D(\pone || \ptwo)$ is then
the average information for choosing $H_\pone$
over $H_\ptwo$.  Thus, the KL divergence does indeed measure the dissimilarity
between two distributions, since the greater their divergence is, the easier it
is, on average, to distinguish between them.

Another statistical rationale for using the KL divergence is given by
\namecite{Cover+Thomas:91a}.  Let the {\em empirical frequency distribution}
of a sample {\bf \x} of length $n$ be the probability mass function
$p_{{\bf \x}}$, where $p_{{\bf \x}}(\x)$ is simply the number of times
$\x$ showed up in the sample divided by $n$.  
\begin{theorem}
Let $\ptwo$ be a hypothesized source distribution.  The probability
according to $\ptwo$ of observing a sample of length $n$ with
empirical frequency distribution $\pone$ is approximately
$b^{-nD(\pone || \ptwo)}$, where $b$ is the base of the logarithm function.
\label{th:types}
\end{theorem}
Therefore, we see that if we are trying to
decide between hypotheses $\ptwo_1, \ptwo_2, \ldots, \ptwo_k$ when $\pone$ is the empirical
frequency distribution of the observed sample, then $D(\pone || \ptwo_i)$
gives the relative weight of evidence in favor of hypothesis
$\ptwo_i$.  

The KL divergence arises in information theory as a measure of coding
inefficiency.  If $\X$ is distributed according to $\pone$, then the average codeword length of the best code for $\X$
is the {\em entropy} $H(\pone)$ of $\pone$: 
\begin{displaymath}
H(\pone) \defas - \sum_{\x \in \alph} \pone(\x) \log \pone(\x).
\end{displaymath}
However, if distribution $\ptwo$ were (mistakenly) used to encode
$\X$, then the average codeword length of the resulting code would
increase by $D(\pone || \ptwo)$.  Therefore, if the divergence between
$\pone$ and $\ptwo$ is large, then $\pone$ and $\ptwo$ must be
dissimilar, since it is inefficient (on average) to use $\ptwo$ in
place of $\pone$.

Finally, we look at the maximum entropy argument.  The entropy of a
distribution can be considered a measure of its uncertainty;
distributions for which many outcomes are likely (so that one is
``uncertain'' which outcome will occur) can only be described by
relatively complicated codes.  The {\em maximum entropy principle},
first stated by
\namecite{Jaynes:57a}, is to assume that the distribution underlying some
observed data is the distribution with the highest entropy among all
those consistent with the data -- that is, one should pick the
distribution that makes the fewest assumptions necessary.  If one accepts the
maximum entropy principle, then one can use it to motivate the use of
the KL divergence in the following manner.  The distribution $\tilde{\ptwo}(\x) =
1/\vert \alph \vert$ is certainly the a priori maximum entropy
distribution.  We can write 
\begin{eqnarray*}
D(\pone || \tilde{\ptwo}) & =& \sum_{\x \in \alph} \pone(\x) \log \pone(\x) -
\sum_{\x \in \alph} \pone(\x) \log \tilde{\ptwo}(\x) \\
 & = & - H(\pone) - \log \frac{1}{\vert \alph \vert} \\
 & = & \log \vert \alph \vert - H(\pone).
\end{eqnarray*}
Maximizing entropy is therefore equivalent to minimizing the KL
divergence to the prior $\tilde{\ptwo}$ given above, subject to the constraint that one must
choose a distribution that fits the data.

To summarize, we have described three motivations for using the KL
divergence.  For the sake of broad acceptability, we have given both
Bayesian arguments (those that refer to priors) and non-Bayesian
ones.\footnote{ The often heated debates between Bayesians and
non-Bayesians are well known.  For example,
\namecite[pg. 24]{Skilling:91a} writes, ``there is a valid defence [sic] of
using non-Bayesian methods, namely incompetence.''}  These are by no
means the only reasons.  For further background, see
\namecite{Cover+Thomas:91a} and \namecite{Kullback:59a} for general
information, \namecite{Aczel+Daroczy} for an axiomatic development, and
\namecite{Renyi:70a} for a description of information theory that uses
the KL divergence as a starting point.

Some authors
\cite{Brown+al:class,Church+Hanks:90a,Dagan+Marcus+Markovitch:95a,Luk:95a} use
the {\em mutual information}, which is the KL divergence between the
joint distribution of two random variables and their product
distributions.  Let $A$ and $B$ be two random variables with
probability mass functions $f(A)$ and $g(B)$, respectively, and let
$h(A,B)$ be their joint distribution function.  Then
\begin{equation}
I(A,B) = D(h||f\cdot g) = \sum_{a \in {\cal A}} \sum_{b \in {\cal
B}} h(a,b) \log \frac{h(a,b)}{f(a) \cdot g(b)}
\label{mutual-info},
\end{equation}
where ${\cal A}$ and ${\cal B}$ denote the sets of possible values for
$A$ and $B$, respectively.  The mutual information measures the
dependence of $A$ and $B$, for if $A$ and $B$ are independent, then $h
=f\cdot g$, which implies that the KL divergence between $h$ and
$f\cdot g$ is zero by the information inequality (theorem
\ref{th:info-ineq}).  We will not give the mutual information further
consideration because we do not wish to attempt to estimate joint
distributions.  Indeed, \namecite{Church+Hanks:90a} consider two words
to be associated if the words occur near each other in some sample
of text; but \namecite{\vhref} note that the occurrence of two
adjectives in the same noun phrase means that the adjectives cannot be
similar.  Thus, the information that joint distributions carry about
similarity varies too widely across different applications for it to
be a generally useful notion for us.

While there are many theoretical reasons justifying the use of the KL
divergence, there is a problem with employing it in practice.  Recall
that for distributions $\pone$ and $\ptwo$, $D(\pone || \ptwo)$ is
infinite if there is some $\x' \in \alph$ such that $\ptwo(\x') = 0$
but $\pone(\x')$ is nonzero.  If we know $\pone$ and $\ptwo$ exactly,
then this is sensible, since the value $\x'$ allows us to distinguish
between $\pone$ and $\ptwo$ with absolute confidence.  However, often
it is the case that we only have estimates $\hat{\pone}$ and
$\hat{\ptwo}$ for $\pone$ and $\ptwo$.  If we are not careful with our
estimates, then we may erroneously set $\hat{\ptwo}(\x)$ to zero for
some $\x$ for which $\pone(\x) > 0$, with the effect that $D(\hat{\pone} || \hat{\ptwo})$ can be 
infinite when $D(\pone || \ptwo)$ is not.

There are several ways around this problem.  One is to use smoothed
estimates, as described above in section \ref{sec:smoothing}, for
$\pone$ and $\ptwo$; this is the approach taken in chapter
\ref{ch:pp}.  Another is to only calculate the KL divergence between
distributions and average distributions.  The work described in
chapter \ref{ch:clust} computes divergences to cluster centroids,
which are created by averaging a whole class of objects.  Chapter
\ref{ch:wsd} describes experiments where we calculate the total
divergence of  $\pone$ and $\ptwo$ to their average; we examine some
properties of the total divergence
in the next subsection.

\subsection{Total Divergence to the Mean}
\label{sec:KL+}

Equation (\ref{total-divergence}) gives the definition of the {\em total
(KL) divergence to the mean}, which appears in
\namecite{Dagan+Lee+Pereira:comp}  ($A$
stands for ``average''):
\begin{equation}
A(\pone,\ptwo) \defas D(\pone || \frac{\pone + \ptwo}{2}) + D(\ptwo ||
\frac{\pone + \ptwo}{2}),
\label{total-divergence}
\end{equation}
where $((\pone + \ptwo)/2)(\x) = (\pone(\x) + \ptwo(\x))/2$.
If $\pone$ and $\ptwo$ are two empirical
frequency distributions (defined just above theorem \ref{th:types}),
then $A(\pone, \ptwo)$ can be used as a test statistic for the
hypothesis that $\pone$ and $\ptwo$ are drawn from the same
distribution.

Using
theorem \ref{th:info-ineq}, we see that $A(\pone,\ptwo) \geq 0$, with
equality if and only if $\pone = \ptwo$.  $A(\pone,\ptwo)$ is
clearly a symmetric function, but does not obey the triangle
inequality, as will be shown below.

We can write $A(\pone,\ptwo)$ in a more convenient form by
observing that
\begin{eqnarray*}
D(\pone \Vert \frac{\pone + \ptwo}{2}) & = & \sum_{\x \in \alph} \pone(\x) \log
\frac{2 \pone(\x)}{\pone(\x) + \ptwo(\x)}
\\
 & = & \log 2 + \sum_{\x \in \alph} \pone(\x) \log \frac{ \pone(\x)}{\pone(\x) + \ptwo(\x)}.
\end{eqnarray*}
The sum over $\x \in \alph$ may be broken up into two parts, a sum
over those $\x$ such that both $\pone(\x)$ and $\ptwo(\x)$ are 
greater than zero, 
 and a sum
over those $\x$ such that $\pone(\x)$ is greater than zero but
$\ptwo(\x) = 0$.  We call these sets $\Both$ and $\Poneonly$,
respectively:  $\Both \defas \set{\x : \pone(\x) > 0, \ptwo(\x) > 0}$ and
$\Poneonly \defas \set{\x: \pone(\x) > 0, \ptwo(\x) = 0}$.  Then,
\begin{eqnarray*}
D(\pone \Vert \frac{\pone + \ptwo}{2}) & = & \log 2 + \sum_{\x \in
\both} \pone(\x) \log \frac{ \pone(\x)}{\pone(\x) + \ptwo(\x)} +
\sum_{\x \in \poneonly} \pone(\x) \log \frac{ \pone(\x)}{\pone(\x) +
\ptwo(\x)} \\
  & = &  \log 2 + \sum_{\x \in
\both} \pone(\x) \log \frac{ \pone(\x)}{\pone(\x) + \ptwo(\x)} + 
\sum_{\x \in \poneonly} \pone(\x) \log \frac{ \pone(\x)}{\pone(\x)} \\
 & = & \log 2 + \sum_{\x \in
\both} \pone(\x) \log \frac{ \pone(\x)}{\pone(\x) + \ptwo(\x)}.
\end{eqnarray*}
A similar
decomposition of $D(\ptwo \Vert \frac{\pone + \ptwo}{2})$ into two
sums over $\Both$ and $\Ptwoonly \defas \set{\x: \ptwo(\x) > 0, \pone(\x) =
0}$ holds.  Therefore, we can write
\begin{equation}
A(\pone, \ptwo) = 2 \log2 + \sum_{\x \in \both} \left\lbrace \pone(\x)
\log \frac{\pone(\x)}{ \pone(\x) + \ptwo(\x)} 
+ \ptwo(\x) \log \frac{\ptwo(\x)}{\pone(\x) + \ptwo(\x)}\right\rbrace.
\label{avg-both}
\end{equation}
Equation (\ref{avg-both}) is computationally convenient, for it
involves sums only over elements of $\Both$, as opposed to over all
the elements in $\alph$.  We will typically consider situations in
which $\Both$ is (estimated to be) much smaller than $\alph$.

Since the two ratios in (\ref{avg-both}) are both less than one, the
sum over elements in $\Both$ is always negative.  $A(\pone,\ptwo)$
therefore reaches its maximum when the set $\Both$ is empty, in which
case $A(\pone,\ptwo) = 2 \log 2$.  This observation makes it easy to
see that $A(\pone,\ptwo)$ does not obey the triangle inequality.  Let
$\alph = \set{\x_1, \x_2}$.  Consider distributions $\tilde{\pone}$,
$\tilde{\ptwo}$, and $\tilde{\pthree}$, where
\begin{displaymath}
\tilde{\pone}(\x_1) = 1,\, \tilde{\pone}(\x_2) = 0; \qquad
\tilde{\ptwo}(\x_1) = \tilde{\ptwo}(\x_2) = \frac{1}{2}; \qquad
\tilde{\pthree}(\x_1) = 0, \, \tilde{\pthree}(\x_2) = 1.
\end{displaymath}
Then $A(\tilde{\pone},\tilde{\ptwo}) +
A(\tilde{\ptwo},\tilde{\pthree}) = \log 2 + \log(2/3) + 2 \log(4/3) =
\log 2 + \log(32/27) < 2 \log 2$, whereas
$A(\tilde{\pone},\tilde{\pthree}) = 2 \log 2$, since the supports for
$\tilde{\pone}$ and  $\tilde{\pthree}$ are disjoint.  Therefore,
$A(\tilde{\pone},\tilde{\ptwo}) + A(\tilde{\ptwo},\tilde{\pthree})
\not\ge A(\tilde{\pone}, \tilde{\pthree})$, violating the triangle inequality.

\subsection{Geometric Distances}
\label{sec:geometric}

If we think of probability mass functions as vectors, so that
distribution $p$ is associated with the vector $(p(\x_1),
p(\x_2), \ldots, 
p(\x_N))$ in $\Re^N$, then we can measure the distance between distributions
by various geometrically-motivated functions, including the $L_1$ and
$L_2$ norms and the cosine function.  All three of these functions
appear quite commonly in the clustering literature
\cite{Kaufman+Rousseeuw:90,Cutting+al:92a,Schutze:93a}.  The first two
functions are true metrics, as the name ``norm'' suggests.

The {\em $L_1$ norm} (also called the ``Manhattan'' or ``taxi-cab''
distance) is defined as
\begin{equation}
\Lone(\pone,\ptwo) \defas \sum_{\x \in \alph} | \pone(\x) - \ptwo(\x)|.
\label{L1}
\end{equation}
Clearly, $\Lone(\pone,\ptwo) = 0$ if and only if $\pone(\x) =
\ptwo(\x)$ for all $\x$.  Interestingly, $\Lone(\pone,\ptwo)$ bears the following relation,
discovered independently by Csisz{\'a}r
and Kemperman, to $D(\pone||\ptwo)$:
\begin{equation}
\Lone(\pone, \ptwo) \leq  \sqrt{ D(\pone || \ptwo) \cdot 2 \ln b}\,,
\label{Lone-bound}
\end{equation}
where $b$ is the base of the logarithm function.  Consequently,
convergence in KL divergence implies convergence in the $L_1$ norm.  However, we can find
a much tighter bound, as follows.
By 
dividing up the sum in equation (\ref{L1}) into sums over $\Both$,
$\Poneonly$, and $\Ptwoonly$
as defined in section \ref{sec:KL+}, we obtain
\begin{displaymath}
\Lone(\pone, \ptwo) = \sum_{\x \in \poneonly} \pone(\x) + \sum_{\x \in
\ptwoonly} \ptwo(\x) + \sum_{\x \in \both} |
\pone(\x) - \ptwo(\x) |.
\end{displaymath}
Since 
\begin{displaymath}
\sum_{\x \in \poneonly} \pone(\x) = 1 - \sum_{\x \in \both} \pone(\x)
\qquad \mbox{and} \qquad 
\sum_{\x \in \ptwoonly} \ptwo(\x) = 1 - \sum_{\x \in \both} \ptwo(\x),
\end{displaymath}
we can express $\Lone(\pone,\ptwo)$ in a form depending only on
the elements of $\Both$:
\begin{equation}
\Lone(\pone, \ptwo) = 2 + \sum_{\x \in \both} \left(|\pone(\x) -
\ptwo(\x) | - \pone(\x)  - \ptwo(\x)  \right).
\label{Lone-both}
\end{equation}
Applying the triangle inequality to (\ref{Lone-both}),
we see that
 $\Lone(\pone, \ptwo) \leq 2$, with equality
if and only if the set $\Both$ is empty.
Also, (\ref{Lone-both}) is a convenient expression from a computational
point of view, since we do not need to sum over all the elements of
$\alph$.  We describe experiments using $\Lone$ as distance function
in chapter \ref{ch:wsd}.

The {\em $L_2$ norm} is the  Euclidean distance between vectors.
Let $|| \cdot ||$ denote the usual norm function, $||\pone(\x)|| = \sqrt{\sum_{\x} \pone(\x)^2}$. Then,
\begin{displaymath}
L_2(\pone,\ptwo) = ||\pone(\x) - \ptwo(\x)|| \defas \left( \sum_{\x \in \alph} (\pone(\x) -
\ptwo(\x))^2 \right)^{\frac{1}{2}}.
\end{displaymath}
Since the $L_1$ norm bounds the $L_2$ norm, the inequality
of equation (\ref{Lone-bound}) also applies to the $L_2$ norm.

Although the $L_2$ norm appears quite often in the literature,
\namecite{Kaufman+Rousseeuw:90} write that
\begin{quote}
In many branches of univariate and multivariate statistics it has 
been known for a long time that methods based on the minimization of
sums (or averages) of dissimilarities or absolute residuals (the
so-called $L_1$ methods) are much more robust than methods based on
sums of squares (which are called $L_2$ methods).  The computational
simplicity of many of the latter methods does not make up for the fact
that they are extremely sensitive to the effect of one or more
outliers. (pg. 117)
\end{quote}
We therefore will not give further consideration to the $L_2$ norm in
this thesis.

Finally, we turn to  the {\em cosine function}.  This symmetric function is
related to the angle between two vectors; the ``closer'' two vectors
are, the smaller the angle between them.   
\begin{equation}
\cos(\pone,\ptwo) \defas \frac{\sum_{\x \in \alph} \pone(\x)
\ptwo(\x)}{||\pone|| ||\ptwo||}
\label{cosine}
\end{equation}
Notice that the cosine is an inverse distance function, in that it
achieves its maximum of 1 when $\pone(\x) = \ptwo(\x)$ for all $\x$,
and is zero when the supports of $\pone$ and $\ptwo$ are disjoint.
For all the other functions described above, it is just the opposite:
they are zero if and only if $\pone(\x) = \ptwo(\x)$ for all
$\x$, and are greater than zero otherwise.  Further analysis of
geometric properties of the cosine function and other geometric
similarity functions used in information retrieval can be found in
\namecite{Jones+Furnas:87a}.

The cosine function is not as efficient to compute as the other
functions we have discussed.  While the numerator in (\ref{cosine})
requires only summing over elements of $\Both$, the elements of
$\Poneonly$ and $\Ptwoonly$ must be taken into account in calculating
the denominator.  It may be desirable to calculate the norms of all
distributions as a preprocessing step (we cannot just normalize the
vectors because we would violate the constraint that attribute vector
components sum to one).

\subsection{Similarity Statistics}
\label{sec:statistics}

There are many correlation statistics for measuring the association
between random variables \cite[Chapter 4.2]{Anderberg:73a}.  The most
well-known of these is the Pearson correlation coefficient; some
non-parametric measures are the gamma statistic, Spearman's
correlation coefficient, and Ken\-dall's $\tau$ coefficient
\cite{Gibbons:93a}.  The Spearman statistic was used by
\namecite{Finch+Chater:92a} to find syntactic categories, and
Kendall's statistic appears in work  by
\namecite{Hatzivassiloglou+McKeown:93a} (henceforth \vh) on clustering
adjectives.  We concentrate on the latter statistic since we will
discuss \vh's work in some detail in the next chapter.

Kendall's $\tau$ coefficient is based on pairwise comparisons.  For
every pair of contexts ($\x_i$, $\x_j$), we consider the quantities
$\alpha^{ij}_\pone=\pone(\x_i) - \pone(\x_j)$ and $\alpha^{ij}_\ptwo =
\ptwo(\x_i) - \ptwo(\x_j)$.  The pair is a {\em concordance}
if both $\alpha^{ij}_\pone$ and $\alpha^{ij}_\ptwo$ have the same sign, and a
{\em discordance} if their signs differ (if either of these quantities
is zero, then the pair is a tie, which is neither a concordance nor a discordance).
$\tau(\pone,\ptwo)$ is the difference between the probability of
observing a concordance and the probability of observing a discordance,
and so ranges between $-1$ and $1$.  A value of $1$ corresponds to
perfect concordance (but not necessarily equality) between $\pone$ and
$\ptwo$, $-1$ corresponds to perfect
discordance, and $0$ to  no correlation.
An unbiased estimator of $\tau(\pone,\ptwo)$ is
\begin{displaymath}
\hat{\tau}(\pone,\ptwo) \defas \frac{\mbox{number of observed concordances} - \mbox{number of
observed discordances}}{{|\alph| \choose 2}}.
\end{displaymath}

In terms of computational efficiency, $\tau(\pone,\ptwo)$ is slightly
more expensive then the total divergence to the mean or the $L_1$
norm.  In order to calculate the number of discordances,
\vh \, first order the $\x$'s in $\alph$ by their
probabilities as assigned by $\pone$.  Then, they rerank the $\x$'s
according to the probabilities assigned by $\ptwo$.  The number of
discordances is then exactly the number of discrepancies between the
two orderings.  
Since we need to sort the set $\alph$ and calculate the
number of discrepancies between the two orderings, we spend 
$O(|\alph| \log_2 |\alph|)$ time 
to calculate the similarity between $\pone$ and
$\ptwo$.  An optimization not noted by \vh\, is that for all
$\x' \in \Both \cup \Poneonly \cup \Ptwoonly$ and $\x'' \notin
\Both \cup \Poneonly \cup \Ptwoonly$
(that is, $\pone(\x'') = \ptwo(\x'') = 0$), the pair $(\x',
\x'')$ cannot be a discordance -- it is a concordance if $\x' \in
\Both$ and a tie otherwise.  Therefore, we actually only need to sort
$\alph' = \Both \cup \Poneonly \cup \Ptwoonly$, a $O(|\alph'| \log_2
|\alph'|)$ operation.  In the case of sparse
data, this would be a significant time savings, although we would
still be using more than linear time.

\subsection{An Example}

\begin{figure}
\epsfscaledbox{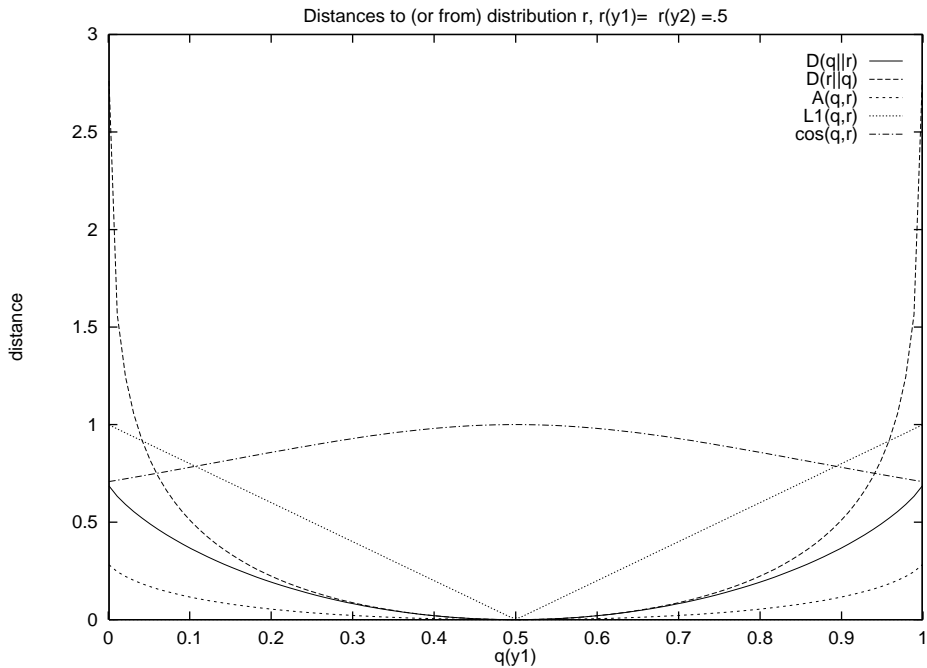}{5in}
\caption{Comparison of distance functions}
\label{fig:dist-functions}
\end{figure}

To aid in visualizing the behavior of the salient functions described
above, we consider a two-dimensional example where $\alph =
\set{\x_1,\x_2}$.  In this situation, $\pone(\x_2) = 1 - \pone(\x_1)$ for any
distribution $\pone$, so we only need to know the value of a distribution at
$\x_1$.  In figure \ref{fig:dist-functions}, we have plotted the
values of various distance functions with respect to a fixed
distribution $\ptwo =(.5,.5)$.  The horizontal axis represents the probability
of $\x_1$, so that $.75$ on the  horizontal axis means the distribution $\pone =
(.75,.25)$.  The fixed distribution $\ptwo$ is at $.5$ on the
horizontal axis.

As observed above, the KL divergences, the total divergence to the
mean, and the $L_1$ norm are all zero at $\ptwo\,$ and increase as one
travels away from $\ptwo$.  The cosine function, on the other hand, is
$1$ at $\ptwo$ and decreases as one travels away from $\ptwo$.

Figure
\ref{fig:dist-functions} demonstrates that the KL divergence is not
symmetric, for the curve $D(\ptwo||\pone)$ lies above the curve
$D(\pone||\ptwo)$.  In general, the KL divergence from a sharp to a
flat distribution is less than the divergence from a flat to a sharp
distribution -- a sharp distribution (such as $(.9,.1)$) is one with
relatively high values for some of the attributes, whereas a flat
distribution resembles the uniform distribution.  The intuition behind
this behavior is as follows.  If we assume that the source
distribution (the second argument to $D(\cdot||\cdot)$) is flat, then
it would be somewhat odd to observe a sharp sample distribution.
However, it would be even more surprising to observe a flat sample if
we believe that the source distribution is sharp.  For instance,
suppose the source distribution were (.5,.5).  Then, the probability
of observing 9 $\context_1$'s and $1$ $\context_2$ in a sample of
length 10 (i.e., a sharp empirical distribution) would be
\begin{displaymath}
{10 \choose 9} (.5)^9(.5)^1 \approx .01.
\end{displaymath}
However, if the source distribution were (.9,.1), then the probability
of observing 5 $\context_1$'s and 5 $\context_2$'s (i.e., a flat
empirical distribution) would be
\begin{displaymath}
{10 \choose 5} (.9)^5 (.1)^5 \approx .001.
\end{displaymath}

An interesting feature to note is that the curve for $A(\pone,\ptwo)$,
the total divergence to the mean, is lower than the KL divergence
curves, and that these, in turn, are for the most part lower than the
$L_1$ curve.  We speculate that the flatness of $D(\pone|| \ptwo)$ and
$A(\pone,\ptwo)$ relative to $L_1(\pone,\ptwo)$ around the point
$\pone = \ptwo$ indicates that these two functions are somewhat more
robust to sampling error, for using $\pone = \ptwo + \epsilon$ (for
small $\epsilon$) instead of $\pone = \ptwo$ results in a much greater
change in the value of the $L_1$ norm than in the value of the KL
divergence or the total divergence to the mean.


\section{Summary and Preview}

We have now established the groundwork for the results of
this thesis.  We have explained why we want to use distributions to
represent objects, and have described ways to estimate these
distributions and to measure the similarity between distributions.  

We have been working with conditional probabilities induced by objects
over contexts.
As mentioned above, ``objects'' and ``contexts''
are fairly general notions; for instance, an object might be a
document and the contexts might be the set of words that can occur in
a document.  We will confine our attention to modeling
pairs of words, so that $\objects$ and $\contexts$ are sets of words.  
In chapters \ref{ch:clust} and \ref{ch:wsd}, $\objects$ is a set of nouns
and $\contexts$ is a set of transitive verbs;  $\counts(\obj, \context)$
indicates the number of times $\obj$ was the direct object of
verb $\context$.  Chapter \ref{ch:pp} considers the bigram case, where
$\objects$ is the set of all possible words, $\contexts = \objects$, and
$\counts(\obj,\context)$ denotes the number of times word $\obj$
occurred immediately before the word $\context$.


\newcommand{\beq}[1]{\begin{equation}\label{#1}}
\newcommand{\eeq}{\end{equation}}
\newcommand{\bseq}[1]{\begin{displaymath}}
\newcommand{\eseq}{\end{displaymath}}
\newcommand{\beqa}[1]{\begin{equation}\label{#1}\begin{eqalign}}
\newcommand{\eeqa}{\end{eqalign}\end{equation}}
\newcommand{\bsubeq}[1]{\begin{subequations}\label{#1}\begin{eqalignno}}
\newcommand{\esubeq}{\end{eqalignno}\end{subequations}}
\newcommand{\defeq}{\stackrel{\rm def}{=}}
\def\base{e}
\sloppy

\setcounter{chapter}{2}
\chapter{Distributional Clustering}
\label{ch:clust}

\newtheorem{lemma}{Lemma}[chapter]

This chapter describes the first of our similarity-based methods for
estimating probabilities.  The probabilistic, hierarchical
distributional clustering scheme detailed here is a model-based
approach, where the behavior of objects is modeled by class behavior.
The following two chapters describe a nearest-neighbor approach, where 
we base our estimate of an object's behavior on the behavior of objects
most similar to it, so that no class construction is involved.

\section{Introduction}

Much attention has been devoted to the study of clustering techniques, and indeed whole books have been written on the subject
\cite{Anderberg:73a,Hartigan:75a,Kaufman+Rousseeuw:90}.  Traditional
applications of clustering include discovering structure in data and
providing summaries of data.  We propose to use clustering as a solution to
sparse data problems: by grouping data into similarity classes, we
create new, generalized sources of information which may be consulted
when information about more specific events is lacking.  That is,
if we wish to estimate the probability of an event ${\cal E}$ that occurs
very rarely in some sample, then we can base our estimate on the
average behavior of the events in ${\cal E}$'s class(es); since a class
encompasses several data points, estimates of class probability are
based on more data than estimates of the probability of a single
event.  For example, suppose we wish to estimate the graduation rate
of Asian-American females enrolled at Westlake High School in Westlake, Ohio.
If there is only one Asian-American female at WHS,
then we will not have enough data to infer the right rate (we would
probably have to guess either 100\% or 0\%). 
Suppose, however, that we consider a group of high schools that are
similar to WHS (e.g., public high schools in suburban areas in Ohio).
Then, we can average together
information about Asian-American females attending schools in that
group to make a better estimate.

To our knowledge, all clustering algorithms in the
natural language processing literature create ``hard'' or Boolean
classes, with every data point belonging to one and only one class.
In other words, these algorithms build partitions of the data space.
The combinatorial demands of such hard clustering schemes are
enormous, as there are $\left\{{n \atop k} \right\}$ ways to group $n$
observations into $k$ non-empty sets, where
\begin{displaymath}
\left\{ {n \atop k} \right\} = \frac{1}{k!} \sum_{i=0}^{k}
(-1)^{k-i}{k \choose i} i^n
\end{displaymath}
is a Stirling number of the second kind \cite{Knuth:vol1}.  There are
a huge number of possible groupings even for small values of $k$ and
$n$: \namecite{\vhref} observe that one can divide twenty-one points
into nine sets in approximately $1.23 \times 10^{14}$ ways.  As it
turns out, the problem of finding a partition that minimizes some
optimization function is NP-complete \cite{Brucker:78a}, so,
not surprisingly, most hard clustering algorithms resort to greedy or
hill-climbing search to find a good partition.

Greedy and hill-climbing approaches all first create an initial
clustering and then iteratively make local changes to the clustering
in order to improve the value of some optimization function.  Let $k$
be the desired number of clusters.  {\em Update} methods begin with
$k$ initial classes chosen in some fashion, and repeatedly move data
points from one class to another.  The number of clusters therefore
stays (about) the same from one iteration to the next.  Two special
cases of update methods are {\em medoid} and {\em centroid} methods,
both of which represent clusters by data points. Medoids are actual
data points, whereas centroids are ``imaginary'' data points created
by averaging together object distributions
\cite{Kaufman+Rousseeuw:90}.  Cluster membership is decided by
assigning each object to the closest cluster representative, where
``closeness'' is measured by some distance function.  Each iteration
step consists of first moving some representative in order to improve
the value of the optimization criterion, and then updating cluster
memberships.

Non-update methods, where the number of clusters varies during the
course of the clustering, include {\em divisive} and {\em
agglomerative} clustering.  Divisive algorithms start with one
universal class to which all the data points belong; each iteration
involves choosing one of the current set of classes to split into two
new classes.  Agglomerative algorithms, in contrast, begin with each
data point belonging to its own class; then, in each iteration step,
some pair of current classes is merged to form a new, larger class.
In either case, the choice of which class to split or which classes to
merge is generally made by picking the class or classes whose
division or combination results in the largest improvement in the
optimization function, and the process stops once $k$ clusters have
been formed.

Both divisive algorithms and agglomerative algorithms, if allowed to
run until all classes have been merged into one, readily yield {\em
hierarchical} clusterings, which can be represented by {\em
dendrograms} (essentially, binary trees).  At the root of the
dendrogram is the class containing all the data points (the first
class considered in the divisive case and the last class formed in the
agglomerative case).  Each node $\eta$ in the dendrogram represents a class,
denoted by ${\rm class}(\eta)$.  Nodes $\eta_1$ and $\eta_2$ are children
of node $\eta'$ if at some iteration step either ${\rm
class}(\eta')$ was divided into ${\rm class}(\eta_1)$ and ${\rm
class}(\eta_2)$, or ${\rm class}(\eta_1)$ and ${\rm class}(\eta_2)$
were agglomerated into ${\rm class}(\eta')$, depending on which type of
clustering algorithm was used.  

While the class hierarchy produced may of course itself be of
interest, an appealing aspect of hierarchical clustering is that it provides
an attractive solution to the  problem of deciding
on the right number of clusters.  The partitioning methods mentioned
above  generally take the number of clusters $k$ as an input parameter
rather than deciding what the right number of clusters is.
As
\namecite{Anderberg:73a}  writes, 
``Hierarchical clustering methods give a configuration for every
number of clusters from one (the entire data set) up to the number of
entities (each cluster has only one member)'' (pg. 15).  However, both
\namecite{Anderberg:73a} and
\namecite{Kaufman+Rousseeuw:90} 
express reservations about hierarchical
methods:  
\begin{quote}
A hierarchical method suffers from the defect that it can
never repair what was done in previous steps.  Indeed, once an
agglomerative algorithm has joined two objects, they cannot be
separated....Also, whatever a divisive algorithm has split up cannot
be reunited.  The rigidity of hierarchical methods is both the key to
their success (because it leads to small computation times) and their
main disadvantage (the inability to correct erroneous decisions).''
\cite[pp. 44-45]{Kaufman+Rousseeuw:90} 
\end{quote}

We propose a novel ``soft'' (probabilistic) hierarchical clustering
method that overcomes this rigidity problem.  Instead of each data
point belonging to one and only one class, we assign probabilities of
class membership, with every data point belonging to every class with
positive probability.  Since we reestimate membership probabilities at
each iteration, there is no sense in which data points can be
permanently assigned to the same or separate classes.

\begin{figure}
\unitlength1in
\begin{minipage}[t]{7in}
\hspace{2.41in}
\epsfscaledbox{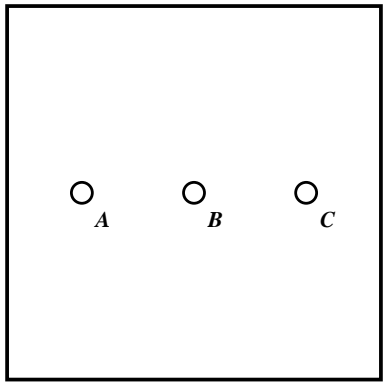}{2in} 
\caption{An ambiguous case}
\label{fig:ambig}
\end{minipage}
\end{figure}

Probabilistic clusterings have another advantage:  they provide a more
descriptive summary of the data.  Consider the situation depicted in
figure \ref{fig:ambig}, where circle $B$ is halfway between 
$A$ and $C$.  Suppose that two clusters are desired.  A hard clustering
is forced  to associate $B$ with only one of the
other circles, say, $A$.  It then reports that the partition found is
$\set{\set{A,B},\set{C}}$, which does not convey the information that $B$
could just as well have been grouped with $C$.  A soft clustering, on
the other hand, can state that $B$ belongs to $A$'s cluster and $C$'s
cluster with equal probability, and so can express the ambiguity of
the situation.

In brief, our clustering method is a centroid-based, probabilistic,
divisive, hierarchical algorithm for associating abstract objects by
learning their distributions.  Each class is represented by a
centroid, which is placed at the cluster's (weighted) center of mass.  For
each object $\obj$ and each centroid $c$, we calculate a membership
probability $P(c|\obj)$ that $\obj$ belongs to $c$.  Our method begins
by creating a single centroid, with each object belonging to that
centroid with probability one, and then iteratively splits one of the
current centroids and reestimates membership probabilities.  The
creation of child centroids from parent centroids creates a hierarchy
of classes in the obvious way.

As decided upon in section \ref{sec:representation}, objects (both
data points and centroids) will be represented by distributions over a
set $\contexts$ of contexts.  We will use the KL divergence, discussed
at length in section \ref{sec:KL}, as distance function.  Our
optimization function is the {\em free energy}, a quantity motivated
by statistical physics; the algorithm uses deterministic annealing to
find phase transitions of the free energy, and splits cluster centroids
at these transitions.  Each time we update the annealing
parameter, we reestimate the location of the cluster centroids and the
membership probabilities for each object.  

We shall be especially interested in the problem of clustering words,
although our theoretical results will be described in a general
fashion.  We re-emphasize that our clustering method can be used for
clustering any objects that can be described as distributions, and
indeed future work involves employing our techniques for clustering documents.
We evaluate our method on tasks involving the prediction of
object-verb pairs, and find that it greatly reduces error rate,
especially in cases where traditional methods such as Katz's \backoff\,
method (see section \ref{sec:smoothing}) would fail.


\section{Word Clustering}
\label{sec:problem}

Methods for automatically classifying words according to their
contexts are of both scientific and practical interest. The
scientific questions arise in connection with distributional views of
linguistic (particularly lexical) structure and also in relation to
the question of lexical acquisition. From a practical point of view,
word classification addresses issues of data sparsity and
generalization in statistical language models, especially 
models used to decide among alternative analyses proposed by a grammar.

It is well known that a simple tabulation of frequencies of certain
words participating in certain configurations  (for example,
frequencies of pairs of transitive main verbs and head nouns of
the verbs' direct objects), cannot be reliably used for comparing the
likelihoods of different alternative configurations. The problem is
that for large samples, the number of possible joint events is
much larger than the number of event occurrences in the sample, so
many events occur rarely or even not at all. Frequency counts thus yield
unreliable estimates of their probabilities.

\namecite{Hindle:90a} proposed dealing with the data sparseness
problem by estimating the likelihood of unseen events from that of
``similar'' events that have been seen. For instance, one may estimate
the likelihood of a particular adjective modifying a noun from the
likelihoods of that adjective modifying similar nouns. This requires a
reasonable definition of noun similarity and a method for
incorporating the similarity into a probability estimate.  In Hindle's
proposal, words are similar if there is strong statistical evidence
that they tend to participate in the same events. His notion of
similarity seems to agree with our intuitions in many cases, but it is
not clear how to use this notion to construct word classes and
corresponding models of association.

In this chapter, we build a similarity-based probability model out of
two parts: a model of the association between words and certain hidden
classes, and a model of the behavior of these classes.  Some researchers
have built such models from preexisting sense classes constructed by
humans; for example, \namecite{Resnik:92a} uses WordNet, and
\namecite{Yarowsky:92b} works with Roget's thesaurus.  As mentioned in
chapter \ref{ch:sim}, however, we are interested in ways to derive
classes directly from distributional data.  Resnik's thesis contains a
discussion of the relative advantages of the two approaches
\cite{Resnik:93a}.

In what follows, we will consider two sets of words, the set $\objects$
 of nouns, and the set
$\contexts$ of transitive verbs.  We are interested in the
object-verb relation:  the pair $(\obj,\context)$ denotes the
event that noun $\obj$ occurred as the head noun of the direct object
of verb $\context$. Our raw knowledge about the relation
consists of the frequencies $\counts(\obj,\context)$ of
particular pairs $(\obj,\context)$ in the required configuration in a
training corpus. Some form of text analysis is required to collect
these pairs. The counts used in our first experiment
were derived from newswire text automatically parsed by Hindle's parser
Fidditch \cite{Hindle:fidditch}. Later, we constructed
similar frequency tables with the help of a statistical part-of-speech tagger
\cite{Church:88a} and  tools for regular expression pattern-matching
on tagged corpora \cite{Yarowsky:92a}. We have not compared the
accuracy and coverage of the two methods or studied what biases they
introduce, although we took care to filter out certain systematic
errors (for instance, subjects of complement clauses for report verbs
like ``say'' were  incorrectly parsed as direct objects).

We  only consider the problem of classifying nouns according
to their distribution as direct objects of verbs; the converse problem
is formally similar.  For the noun classification problem, the
empirical distribution of a noun $\obj$ is given by the
conditional density 
\begin{displaymath}
\pmle(\context | \obj)
= \frac{\counts(\obj,\context)}{\sum_\context \counts(\obj,\context)} =
\frac{\counts(\obj,\context)} {\counts(\obj)},
\end{displaymath} where $\counts(z)$ denotes the number of times
event $z$ occurred in the training corpus. The problem we study is how
to use the $\pmle(\cdot | \obj)$ to classify the $\obj \in
\objects$. Our classification method will construct a set $\clusters$
of clusters $c$ and cluster membership probabilities $\pmem$. Each cluster
$c$ is associated with a cluster  centroid distribution
$\pcen$, which is a discrete density over $\contexts$
obtained by computing a weighted average of the noun distributions
$\pmle(\cdot | \obj)$. We will move freely between describing a noun
(or centroid) as $\obj$ (or $c$) and as $\pmle(\cdot|\obj)$ (or $\mp(\cdot|c)$).

To cluster nouns $\obj$ according to their conditional verb
distributions $\pmle(\cdot|\obj)$, we need a measure of similarity
between distributions. We use for this purpose the KL divergence from
section \ref{sec:KL}:
\begin{displaymath}
D(\pone || \ptwo) = \sum_{\context \in \contexts} \pone(\context) \log \frac{\pone(\context)}{\ptwo(\context)}.
\end{displaymath}
The KL divergence is a natural choice for a variety of reasons, most
of which we have already discussed in section \ref{sec:KL}.  As
mentioned there, $D(\pone || \ptwo)$ measures how inefficient on
average it would be to use a code based on $\ptwo$ to encode a
variable distributed according to $\pone$. With respect to our
problem, $D(\pmle(\cdot | \obj) || \mp(\cdot|c))$ thus gives us the
loss of information in using the centroid distribution $\mp(\cdot|c)$
instead of the empirical distribution $\pmle(\cdot| \obj)$ when
modeling noun $\obj$.  Furthermore, minimizing the KL divergence yields
cluster centroids that are a simple weighted average of member
distributions, as we shall see.

One technical difficulty  is that $D(\pone ||
\ptwo)$ is infinite when $\ptwo(\context) = 0$ but $\pone(\context) > 0$.  Due to
sparse data problems, it is often the case that
$\pmle(\context|\obj)$ is zero for a particular pair
$(\obj,\context)$.  We could sidestep this problem by smoothing zero
frequencies, perhaps using one of the methods described in section \ref{sec:smoothing}.  However, this is not very
satisfactory because one of the goals of our work is precisely to
avoid data sparsity problems by grouping words into classes. As it
turns out, the difficulty is avoided by our clustering technique: instead
of computing the KL divergence between individual word distributions,
we only calculate divergences between word distributions and cluster
centroids.  Since centroids are average distributions, they are
guaranteed to be nonzero whenever the word distributions are.  This is
a useful advantage of our method over techniques that need to compare
pairs of individual objects, since estimates for individual objects
are prone to inaccuracies due to data sparseness.

The organization of the rest of this chapter is as follows.  
We develop the theoretical basis for our clustering algorithm in
section \ref{sec:clust-gen}.  We present some example clusterings in
section \ref{sec:examples} in order to get a sense of the qualitative
performance of our algorithm.  Section \ref{sec:clust-eval} presents two
evaluations of the ability of our cluster-based probability estimation method
to estimate word pair probabilities, especially in situations where
data is sparse; we
show that indeed, our method does a good job of modeling.  Finally, in
section \ref{sec:clust-related} we
review other work in the NLP community on clustering words, and
briefly touch upon soft clustering methods from other fields.


\section{Theoretical Basis}
\label{sec:clust-gen}

Our general problem can be seen as that of learning the joint
distribution $P(\obj,\context)$ of pairs in $\objects \times
\contexts$ from a large sample.  The training
data is a sample $S$ of $n$ independently drawn pairs
\bseq{sentence}
S_i = ( \obj_{j_i} , \context_{l_i} ), \qquad 1\leq i \leq n.
\eseq
We assume that each $\obj_j \in \objects$ and $\context_l \in \contexts$
occurs in the sample at least once (we cannot train a model for
$\obj$ or $\context$ if we have no information about them).

The line of argument in this section proceeds as follows.  We first
set up the general form of our cluster-based probability model.  We
determine the two principles, minimum distortion and maximum entropy,
that guide our search for the proper parameter settings for the model,
and combine these two principles into the {\em free energy} function.
Sections \ref{sec:membership} and \ref{sec:centroids} go into the
details of how we set the parameters by maximizing entropy and
minimizing distortion.  Finally, section \ref{sec:hierarchical}
describes how searching for phase transitions of the free energy yields
a hierarchical clustering.  

In order to estimate the likelihood of the sample, we need a
probability model $\mp(\obj,\context) = \mp(\obj)\mp(\context|\obj)$.
We would like to find a set of clusters $\clusters$, each represented
by a cluster  centroid $c$, such that each conditional
distribution $\mp(\context|\obj)$ can be decomposed as
\begin{equation}
\mp(\context|\obj) =
   \sum_{c \in \clusters} \pmem \pcen.
\label{cond-prob}
\end{equation}
$\pmem$ is the {\em membership probability} that $\obj$ belongs to $c$,
and $\pcen$ is $\context$'s probability according to the
{\em centroid distribution}
for $c$:  as stated above, centroids are representative objects, and so form a
distribution over $\contexts$ just like objects do.  Ideally, the
objects that belong most strongly to a given cluster would be similar to
one another.

According to equation (\ref{cond-prob}), then, we estimate the
probability of $\context$ given $\obj$ by taking an average of the
centroid distributions, weighting each $\pcen$ by the probability
that $\obj$ belongs to $c$.  We thus make a Markovian assumption that
the association of $\obj$ and $\context$ is made solely through the
clusters, that is, that $\context$ is conditionally independent
of $\obj$ given $c$.  The cluster model drastically reduces the
dimension of the model space, since the number of $(c,\obj)$ and
$(c,\context)$ pairs should be much lower than the number of
possible $(\obj,\context)$ pairs.

Given the decomposition of $\mp(\context | \obj)$ in equation
(\ref{cond-prob}), we can write the likelihood assigned by our model to
a pair as
\begin{equation}
\mp(\obj,\context) =  \sum_{c \in \clusters } \mp(\obj) \pmem \pcen. 
\label{likelihood}
\end{equation}
We will assume that the marginals for $\obj \in \objects$ are not part
of our model and so can be considered fixed; to indicate this, we will
write $P(\obj)$ instead of $\mp(\obj)$.  Without loss of generality,
we assume that $P(\obj)$ is greater than zero for all $\obj$.
In order to flesh equation
(\ref{likelihood}) out, then, we need only  find suitable forms for the cluster
membership distributions $\pmem$ and centroid distributions $\pcen$.
We will be guided by two principles: first, that our model should fit
the data well (otherwise, our model is not useful), and second, that
our model should make as few assumptions as possible (otherwise, our
model is not general).

Goodness of fit is determined by the {\em distortion} of the model.
Equation (\ref{cond-prob}) estimates the probability of $\context$
given $\obj$ by
randomly selecting a cluster $c$ according to distribution
$\pmem$, and then using $\pcen$ to estimate the (conditional)
probability of $\context$. 
Recall from section \ref{sec:KL} that $D(\pmle(\cdot|\obj) ||
\mp(\cdot|c))$ measures the inefficiency of using $c$'s distribution
rather than $\obj$'s maximum likelihood distribution to code
for $\obj$.
 The distortion ${\cal D}$ is the average coding loss incurred by our model:
\begin{equation}
{\cal D} = \sum_{\obj} P(\obj) \sum_{c} \pmem d(\obj,c),
\label{distortion}
\end{equation}
where $d(\obj,c)$ is notational shorthand for
 $D(\pmle(\cdot|\obj) ||\mp(\cdot|c))$.  

As it turns out, the distortion equation does not give us enough
information to find good closed-form expressions for the membership
probabilities.  In fact, without any other constraints, the cluster
system that minimizes distortion is the one in which there is one
centroid placed on top of each object, with each object belonging only
to the centroid it coincides with.  Therefore, we add the requirement
that the membership assignments make the fewest assumptions possible,
that is, that the probability that an object belongs to a centroid
should not be any higher than it needs to be.  This requirement
corresponds to the maximum entropy principle, described in section
\ref{sec:KL}.  Therefore, we wish to maximize the {\em configuration
entropy}
\begin{equation}
H = - \sum_{\obj} P(\obj) \sum_{c} \pmem \log \pmem,
\label{cond-entropy}
\end{equation}
which is the average entropy of the membership probabilities.

We can combine distortion and entropy into a single function, the {\em
free energy}, which appears in work on statistical
mechanics \cite{Rose+al:phase}:
\begin{equation}
F = {\cal D} - H/\beta.
\label{free}
\end{equation}
This function is not arbitrary; indeed, at maximum entropy points (see
section \ref{sec:membership}), we
can show that 
\begin{eqnarray}
  H &=& - \frac{\partial F}{\partial T} \label{deriv-H}  \,\:\: \mbox{and} \\ 
D &=& \frac{\partial \beta F}{\partial \beta}, \label{deriv-D}\\
\end{eqnarray}
where $T = 1/\beta$.
The {\em
minima} of $F$ are of special interest to us, since such points
represent a balance between the ``disordering'' force
of maximizing entropy and the ``ordering'' force of minimizing
distortion. 
In fact, in statistical mechanics, the probability of
finding a system in a given configuration is a negative exponential in
$F$, so the system is most likely to be found in its minimal free
energy configuration.  $\beta$ is a free parameter whose
interpretation we will leave for later.

Suppose we fix the number of clusters $|\clusters|$.  Clearly, (local)
minima of $F$ occur when the entropy is at a (local) maximum and,
simultaneously, the  distortion is at a (local) minimum (although
critical points of $F$ need not correspond to critical points of
${\cal D}$ and $H$).  However, it is difficult to jointly maximize
entropy and minimize distortion, since the location of cluster
centroids affects the membership probabilities, and vice versa; that
is, the $\pcen$ and the $\pmem$ are not independent.  We therefore
simplify the search for minima of $F$ by breaking up the estimation
process into two steps.  First, we hold the distortion and centroid
distributions fixed, and maximize the entropy subject to these
constraints.  Since the distortion is regarded as constant in this
step, maximizing entropy corresponds to a reduction in free energy.
Second, we fix the membership probabilities at the values derived in
the first step, and thus can treat the entropy as a constant.  We then
find a critical point of $F$ with respect to the centroid
distributions; it turns out that this critical point is in fact a
minimum of the distortion, and therefore free energy is reduced once
again.  Moving the centroid distributions may change the values of the
membership probabilities that maximize entropy, though, and so we
repeat these two steps until a stable configuration is reached.  This
two-step estimation iteration is reminiscent of the EM
(Estimation-Maximization) algorithm
\cite{Dempster+al:77a} commonly
used to find maximum likelihood solutions.

Before we continue, we review the notation that will be used in the
following sections.  Model probabilities are always marked with a
tilde ($\mp$).  The model parameters are the membership probabilities
$\pmem$ and the centroid distributions $\pcen$.  
The object marginal probabilities $\mp(\obj) = P(\obj)$ are not
considered part of the model, and so are regarded as positive constants
throughout.\footnote{Our implementation sets $P(\obj) =
1/|\objects|$ instead of $\pmle(\obj)$, since we are interested in
distributional modeling without regard to the frequencies of
particular nouns.}
The centroid
marginals $\mp(c)$ are given by
$\mp(c) = \sum_{\obj}
\pmem P(\obj)$; this form ensures that $\sum_{\obj} \mp(\obj|c) = 1$.  
Empirical frequency
distributions are denoted by $\pmle$ and are considered fixed by
the data.  By assumption, for all $\context$ there exists an object
$\obj$ such that $\pmle(\context|\obj) > 0$.
The quantity $d(x,c)$ is shorthand for the KL divergence $D(\pmle(\cdot |\obj)
||\mp(\cdot|c))$.
We summarize this information in table \ref{table:parameters}.
\begin{table}[ht]
\begin{center}
\begin{tabular}{lll}
Quantity & Value & Notes \\ \hline
$\pmem$ & ? & (to be determined) \\
$\pcen$ & ? & (to be determined) \\
$\mp(\obj)$ & $P(\obj)$ & fixed at positive values\\
$\mp(c)$ & $\sum_{\obj} \pmem P(\obj)$ & determined by $\pmem$ \\
$\mp(\obj | c)$ & $\pmem P(\obj)/\mp(c)$ & determined by $\pmem$  \\
$\pmle(\context | \obj)$ & $\counts(\obj,\context)/\counts(\obj)$ &
fixed by data; $\forall \context, \exists \obj: \pmle(\context|\obj) > 0$ \\
$d(\obj,c)$ & $D(\pmle(\cdot|\obj) || \mp(\cdot|c))$ & determined by $\pcen$
\end{tabular}
\end{center}
\caption{\label{table:parameters} Summary of common quantities}
\end{table}

We will use natural logarithms in this chapter, so that the base of
the logarithm function is $\base$; using another base would not not
substantially alter our results, but we would have extra constant
factors in most of our expressions.  The next two subsections assume
that the number of clusters has been fixed.


\subsection{Maximum-Entropy  Cluster Membership}
\label{sec:membership}

This section addresses the first parameter estimation step of finding
the cluster membership probabilities $\pmem$ that maximize the
configuration entropy, and hence reduce the free energy, assuming that
the distortion and the centroid distributions are fixed (it does not
suffice simply to hold the centroid distributions fixed, since we see
from equation (\ref{distortion}) that the distortion depends on the
membership probabilities, too).  We will make the further assumption
that for all centroids $c$, $\pcen > 0$ for all $\context$; this
assumption is justified in the next section.

Recall the definition of the configuration entropy $H$ from
equation (\ref{cond-entropy}):
\begin{displaymath}
H = - \sum_{\obj} P(\obj) \sum_{c} \pmem \log \pmem.
\end{displaymath}
We wish to maximize this quantity subject to two constraints: the normalization
constraint that  $\sum_{c} \pmem = 1$ for all $\obj$, and the distortion
constraint that ${\cal D} = K$ for some constant $K$.  We
therefore take the variation of the function $\constrained{H}$:
\begin{displaymath}
\constrained{H} = H - \sum_{\obj} \alpha_{\obj} \left(\sum_c \pmem-
1\right) - \beta \left( \sum_{\obj} P(\obj) \sum_c \pmem d(\obj,c) - K \right),
\end{displaymath}
where $\alpha_{\obj}$ and $\beta$ are Lagrange multipliers.  It is
important to note that we are using
$\beta$ both here as a multiplier and as a normalization term
in the free energy (\ref{free}).

We now calculate the partial derivative  of
$\constrained{H}$ with respect to a given membership probability
$\pmemp$, since fixing the centroid distributions means that the
$\pmem$ are independent (except for their association through the
fixed distortion).
\begin{eqnarray*}
\delmemp{\constrained{H}} & = &
\delmemp{H} - \alpha_{\obj} - \beta P(\obj)d(\obj,c)  \\ 
& = & - \left( P(\obj)\pmem \frac{1}{\pmem} + P(\obj) \log \pmem
\right)- \alpha_{\obj} - \beta P(\obj)d(\obj,c)  \\ 
&=& -P(\obj) \left( 
1 + \log \pmem + \frac{\alpha_{\obj}}{P(\obj)} + \beta d(\obj,c) 
 \right) 
\end{eqnarray*}
(there is no problem with division by $P(\obj)$ since we assumed all
object marginals are positive).  

At critical points of $\constrained{H}$, we have that $\delmempslash{\constrained{H}} = 0$.  This allows us to solve for $\pmem$:
\begin{displaymath}
\pmemp = \base^{- \beta d(\obj,c)} \base^{-(1 + \alpha')},
\end{displaymath}
where 
$\alpha' = \alpha_{\obj}/P(\obj)$.  Since $\alpha'$ is
meant to insure the normalization of $\pmemp$, 
$\alpha'$ must be set to a value such that the following is satisfied:
\begin{displaymath}
\base^{1 + \alpha'} = \sum_{c} \base^{-\beta d(\obj,c)} \stackrel{def}{=} Z_{\obj}
\end{displaymath}
($Z$ is standard notation for partition (normalization) functions; the
name comes from the German {\em Zustandsumme}).
We therefore have a closed-form solution for the 
membership probabilities:
\begin{equation}
\pmem = \frac{\base^{-\beta d(x,c)}}{Z_x}.
\label{membership}
\end{equation}
It was shown by \namecite{Jaynes:63a} that the exponential form
(\ref{membership}) gives not just a critical point but the maximum of
the entropy, and so we have a maximum entropy estimate of membership
probability, as desired. The expression (\ref{membership}) is
intuitively satisfying because it makes the membership probabilities
dependent on distance (in the KL divergence sense): the farther $\obj$
is from $c$, the less likely it is that $\obj$ belongs to $c$.
Furthermore, given that the centroid distributions were fixed at
positive values for all $\context$, $d(\obj,c)$ is always defined,
which means that all membership probabilities are positive; each
object has some degree of association with each cluster.  For each
$\pmem$, we need to calculate $d(\obj,c)$ which is a sum over all
$\context \in \contexts$, so the time to update all the membership
probabilities is $O(|\objects||\clusters||\contexts|)$.  However, if
the object distributions are sparse, then the  computation of
$d(\obj,c)$ will be significantly faster.

There is a pleasing relationship between
expression (\ref{membership}) for $\pmem$
and an estimate given by theorem \ref{th:types}, restated here:

\vspace{.2cm}
\noindent {\bf Theorem \ref{th:types}}
{\em Let $\ptwo$ be a hypothesized source distribution.  The
probability according to $\ptwo$ of observing a sample of length $n$
with empirical frequency distribution $\pone$ is approximately
$b^{-nD(\pone || \ptwo)}$, where $b$ is the base of the logarithm
function.}

\vspace{.2cm}
\noindent Thus, the maximum entropy membership probability $\pmem =
\base^{-\beta d(x,c)}/Z_\obj$ corresponds to the probability of
observing object distribution $\pmle(\context | \obj)$ if the source
distribution is assumed to be the centroid $\pcen$, except that
$\beta$ has replaced  the sample size $n$.  Therefore, if
we regard $\beta$ not as a Lagrange multiplier but as a free
parameter, we can in some sense control the sample size ourselves.  If
we use a high value of $\beta$, then we express strong belief in the
maximum likelihood estimate $\pmle$ (as would be the case for a very
large sample), so that the probability that $\obj$ belongs to a
centroid $c$ is negligible unless $d(\obj,c)$ is very small.
Conversely, a low value of $\beta$ is equivalent to a small sample, in
which case we do not trust the MLE and so allow $\pmem$ to be high
even if $\obj$ is relatively distant from $c$.  Section
\ref{sec:hierarchical} describes how we vary $\beta$ in order to
derive a hierarchical clustering.

We conclude this section by observing that at the maximum entropy membership probabilities, the free
energy can be rewritten as follows:
\begin{eqnarray}
F & =& {\cal D} - \frac{1}{\beta} H \nonumber \\
 & = & \sum_{\obj} \sum_c P(\obj) \pmem \left( d(\obj,c) +
\frac{1}{\beta} \log \pmem \right) \nonumber\\
 & = & \sum_{\obj} \sum_c P(\obj) \pmem \left( d(\obj,c) - 
d(\obj,c) - \frac{1}{\beta}\log Z_{\obj} \right)\:\: \mbox{(substitution of (\ref{membership}))} \nonumber\\
 & = & - \frac{1}{\beta} \sum_{\obj} P(\obj) \log Z_{\obj} \sum_c \pmem \nonumber\\
 & = & - \frac{1}{\beta} \sum_{\obj}P(\obj) \log Z_{\obj},
\label{free-maxent}
\end{eqnarray}
where the last step is justified since we ensured the normalization of
the maximum-entropy membership probabilities.  By simple
differentiation, it is easy to see that if we set $T = 1/\beta$, then
$\partial F/ \partial T = - \beta F - \beta {\cal D} = -H$, and
$\partial (\beta F)/ \partial \beta = {\cal D}$.  This gives us
equations (\ref{deriv-H}) and (\ref{deriv-D}), as desired.


\subsection{Minimum-Distortion Cluster Centroids}
\label{sec:centroids}

We now proceed with the second estimation step.
We fix the  membership probabilities $\pmem$ at their maximum entropy
values, calculated above, so that the configuration entropy can now be
considered a constant, and the expression for the free energy is given
by equation (\ref{free-maxent}).

Now that the membership probabilities have been fixed, the individual
centroid distributions are all independent and we just need to
find values for them that minimize $F$, subject
to the constraint that $\sum_{\context} \pcen = 1$ for all centroids.
What we will do is first find a critical point of $F$ (equations
(\ref{var-f-part}) through (\ref{centroid})), and then prove in
lemma \ref{th:min-distortion}
that this critical point is in fact a minimum by showing that it
minimizes the distortion ${\cal D}$.

In order to find a critical point of $F$,  we take partial derivatives of 
\begin{displaymath}
\constrained{F}  =  - \frac{1}{\beta} \sum_{\obj}P(\obj) \log Z_{\obj}
-  \sum_{c}\gamma_c  \left( \sum_{\context} \pcen - 1 \right),
\end{displaymath}
where $\gamma_c$ is yet another Lagrange multiplier and we use
expression (\ref{free-maxent}) for $F$.

The partial derivative of $\constrained{F}$ with respect to a given $\pcen$ is
calculated as follows:
\begin{eqnarray}
\delpcenshort{\constrained{F}} &=&  
- \frac{1}{\beta} \sum_{\obj}P(\obj) \cdot \frac{1}{Z_{\obj}} \cdot \delpcen{\left( \sum_{c'} \base^{-\beta
d(\obj,c')} \right)} 
- \gamma_c \nonumber \\
 & = & - \frac{1}{\beta} \sum_{\obj}P(\obj) \frac{\base^{ -\beta d(\obj,c)}}{Z_{\obj}}
\left( - \beta \delpcenshort{d(\obj,c)} \right) - \gamma_c \nonumber \\
 & = & \sum_{\obj} P(\obj) \pmem \delpcenshort{d(\obj,c)} - \gamma_c.
\label{var-f-part}
\end{eqnarray}
The variation of $d(\obj,c)$ with respect to $\pcen$ is
\begin{eqnarray}
\delpcenshort{d(\obj,c)} & = & \delpcen{ \left(\sum_{\context'} \pmle(\context'|\obj)
\log \frac{\pmle(\context'|\obj)}{\mp(\context'|c)}\right)} \nonumber \\
 &= & - \frac{\pmle(\context|\obj)}{\pcen}, \label{var-d}
\end{eqnarray}
so the centroid distribution term $\pcen$ reappears.
Substituting (\ref{var-d}) into (\ref{var-f-part}), we have
\begin{eqnarray*}
\delpcenshort{\constrained{F}} & = & \sum_{\obj} P(\obj) \pmem \left(-
\frac{\pmle(\context|\obj)}{\pcen} \right) - \gamma_c \\
& = & - \frac{1}{\pcen} \sum_{\obj} P(\obj) \pmem \pmle(\context|\obj)
- \gamma_c.
\end{eqnarray*}
At a critical point of $\constrained{F}$, the partial derivative of
$\constrained{F}$ must be 0, which allows us to solve for $\pcen$:
\begin{equation}
\pcen = \frac{1}{\gamma_c} \sum_{\obj} P(\obj) \pmem
\pmle(\context|\obj)
\label{centroid-nonnorm}
\end{equation}
The multiplier ${\gamma_c}$ is meant to enforce the constraint that
$\sum_{\context} \pcen = 1$, so
\begin{eqnarray*} 
1 &=& \sum_{\context} \frac{1}{\gamma_c}  \sum_{\obj} P(\obj) \pmem
\pmle(\context|\obj) \\
 & = &  \frac{1}{\gamma_c} \sum_{\obj} P(\obj) \pmem.
\end{eqnarray*}
Therefore, $\gamma_c =  \sum_{\obj} P(\obj) \pmem = \mp(c)$; upon
substitution of this into (\ref{centroid-nonnorm}), we finally obtain
the centroid distributions:
\begin{equation}
\pcen = \sum_{\obj} \mp(\obj|c) \pmle(\context | \obj).
\label{centroid}
\end{equation}
We thus have a natural expression for a cluster centroid $c$: it is an
average over all data points $\obj$, weighted by the Bayes inverse of
the probability that $\obj$ belongs to $c$.  The Bayes inverses are
all positive since the maximum-entropy membership probabilities are,
so the centroid distribution cannot be zero for any $\context$ since
we assume that $\pmle(\context|\obj)$ is nonzero for at least one
$\obj$.  It is clear that the time required to update all the centroid
distributions is $O(|\contexts| |\clusters| |\objects|)$ in
the worst case; again, however, the computation is much faster if the
object distributions are sparse.

Now, expression (\ref{centroid}) gives us the unique critical point of
$F$ when the entropy is held fixed; but is the free energy actually reduced at this
point?  Our goal, after all, is to look for minima of $F$.  Since the
entropy was held fixed, it suffices to show that the centroid
distributions (\ref{centroid}) yield a minimum of the distortion,
which we do in the following lemma.
\begin{lemma}
If the distortion ${\cal D}$ has exactly one critical point with
respect to centroid distributions $\pcen$, $0 \leq \pcen \leq 1$, then
that critical point is the unique
minimum of ${\cal D}$, assuming the cluster membership probabilities $\pmem$
are fixed.
\label{th:min-distortion}
\end{lemma}
\proof{
Since 
\begin{eqnarray*}
{\cal D}  &= & \sum_{\obj} P(\obj) \sum_c \pmem d(\obj,c) \\
& = & \sum_{\obj} P(\obj) \sum_{c} \pmem \sum_{\context}
\pmle(\context | \obj) \log \pmle(\context| \obj) - \\
& & \qquad
\sum_{\obj} P(\obj) \sum_{c} \pmem \sum_{\context} \pmle(\context | \obj) \log
\pcen,
\end{eqnarray*}
and the centroid distributions are independent when the membership
probabilities are fixed, it is sufficient to maximize for each
centroid $c$ the quantity
\begin{eqnarray*}
{\cal D}_c & =& \sum_{\obj} P(\obj) \pmem \sum_{\context} \pmle(\context | \obj) \log
\pcen \\
& = & \sum_{\context} \left( \log \pcen \right) \sum_{\obj}  P(\obj) \pmem
\pmle(\context | \obj) \\
& = & \sum_{\context} \log \left( \pcen^{Q(c,\context)} \right) \\
& = & \log \left( \prod_{\context}  \pcen^{Q(c,\context)} \right),
\end{eqnarray*}
where $Q(c, \context) = \sum_{\obj} P(\obj) \pmem \pmle(\context | \obj)$ does
not depend on $\pcen$.  But since the logarithm is a strictly
increasing function, we need only find a maximum of the product 
\begin{equation}
\prod_{\context}  \pcen^{Q(c,\context)}.
\label{d-prod}
\end{equation}
Observe that this product (unlike the logarithm, which is why we 
had to do all this equation rewriting) is continuous on the domain
$\set{\pcen : 0 \leq \pcen \leq 1}$, which is closed and bounded.
Therefore, we know from analysis that (\ref{d-prod}) achieves both its
maximum and its minimum on its domain.  Since clearly every point on
the boundary of the domain yields a 
minimum value (zero), the unique critical
point must be the maximum of (\ref{d-prod}) and thus the minimum of
${\cal D}$.  
} 

Now, since the fixed membership probabilities determine the entropy,
any critical point of $F$ must also be a critical point of the
distortion 
because $\partial F = \partial {\cal D}$ if $H$ is a constant.  Therefore,
the centroid distributions (\ref{centroid}) define the unique critical
point of the distortion, and application of
lemma \ref{th:min-distortion} tells us that this is indeed the minimum
of $\cal D$.  Thus, we have succeeded in finding centroid
distributions which minimize distortion and therefore reduce the free energy.


\subsection{Hierarchical Clustering}
\label{sec:hierarchical}

In the previous two sections, we developed maximum entropy estimates
for membership probabilities and minimum distortion estimates for
centroid distributions:
\begin{eqnarray*}
\pmem & = & \exp(-\beta d(\obj,c))/Z_{\obj}\qquad ~~~(\ref{membership}),\, \mbox{and} \\
\pcen & = & \sum_{\obj} \mp(\obj|c) \pmle(\context|\obj)\qquad (\ref{centroid}).
\end{eqnarray*}
Our search for minima of $F$
at a fixed $\beta$ is a two-step iteration described in section
\ref{sec:clust-gen}.  First, we set the membership probabilities at
their maximum entropy values (\ref{membership}), using the current
centroid distributions.  Then, we plug these membership
probabilities into (\ref{centroid}) to update the centroid
distributions.  We repeat this two-step cycle until the parameters
converge to steady states.  

Now, this two-step iteration lets us find cluster centroids and
membership probabilities for a fixed number of clusters.  However, we
have not yet shown how the number of clusters is chosen.  The
inclusion of the parameter $\beta$ in the free energy expression 
\begin{displaymath}
F = {\cal D} - H/\beta
\end{displaymath}
suggests
the use of a {\em deterministic annealing} procedure for clustering
\cite{Rose+al:phase}, in which the number of clusters is determined
through a sequence of phase transitions by continuously increasing
$\beta$ according to an {\em annealing schedule}.

As discussed in section \ref{sec:membership}, $\beta$ plays a role
similar to sample size and thus controls the importance of the
distance function $d(\obj,c)$.  However, it will now be fruitful to
think of $\beta$ as the inverse of temperature.  At the high
temperature limit (low $\beta$), the entropy $H$ has the biggest role
in minimizing the free energy, so a system consisting of only one
cluster centroid is preferred.  At the low temperature limit (high
$\beta$), the distortion dominates and the minimum-energy
configuration is then the one where we have one centroid placed on top
of every data point, with each data point belonging with probability
one to the centroid it coincides with.  Thus, the system has ``cooled
down'' to the point where the freedom of objects to associate with
distant centroids has disappeared.  Between these two extremes, there
must be critical values of $\beta$ at which {\em phase transitions} occur; that
is, when the natural solution involves including more centroids.  

We find these phase transitions by taking a cluster $c$ and a {\em
twin} $c^*$ of $c$ such that the centroid $\mp(\cdot|c^*)$ is a small
random perturbation of $\mp(\cdot|c)$. Below the critical $\beta$ at
which $c$ splits, the membership and centroid iterative reestimation
procedure will make $P(\cdot|c)$ and $P(\cdot|c^*)$ converge, from
which we infer that $c$ and $c^*$ are really the same cluster. But if
$\beta$ is above the critical value for $c$, the two centroids will
diverge, giving rise to two children of $c$.

A sketch of our clustering procedure appears in figure \ref{fig:proc}.  We start with very low
$\beta$ and a single cluster whose centroid is the average of all noun
distributions (and so is guaranteed to be nonzero for all $\context$). For any given $\beta$, we have a current set of {\em leaf}
clusters corresponding to the current free energy minimum. To
refine such a solution, we search for the lowest $\beta$ that causes
some  leaf cluster to split.  Ideally, there
is just one split at that critical value, but for practical
performance and numerical accuracy reasons we may have several splits
at the new critical point. The splitting procedure can then be
repeated to achieve the desired number of clusters or model
cross-entropy.  

\begin{figure}[ht]
\begin{quote}
\begin{code}
$\beta \leftarrow \beta_0$
create initial centroid
\medskip
REPEAT until $\beta = \beta_{MAX}$ or enough clusters:
    For each centroid $c$, create twin $c^*$
\medskip   
    REPEAT until twins $(c,c^*)$ split or too many iterations:
        Estimate membership probs by (\ref{membership})
        Estimate centroids by (\ref{centroid})
\medskip     
    IF more than one centroid split
    THEN [raised $\beta$ too quickly]
        lower $\beta$
    ELSE IF no centroid split
        raise $\beta$
    ELSE [one centroid split]
        raise $\beta$
    delete extra twins $c^*$
\end{code}
\end{quote}
\caption{\label{fig:proc}Clustering algorithm}
\end{figure}


\section{Clustering Examples}
\label{sec:examples}

\begin{quote}
The properties that the child can detect in the input -- such as the
serial positions and adjacency and co-occurrence relations among words
-- are in general linguistically irrelevant. \cite[pg. 50]{Pinker:84a}
\end{quote}

\begin{figure}
\epsfscaledbox{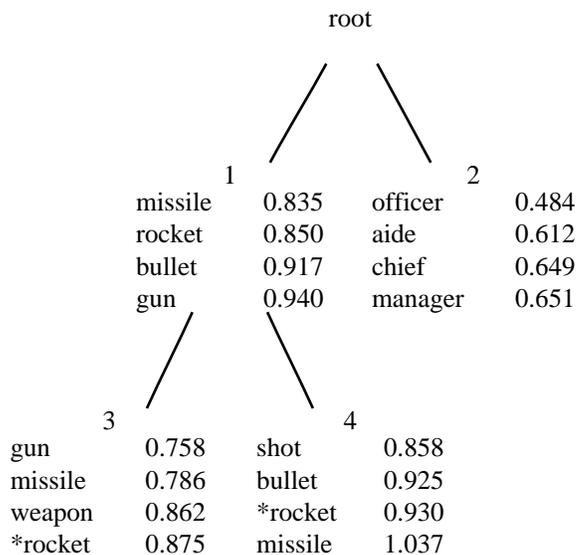}{3in}
\caption{Direct object clusters for verb {\em fire}}
\label{fire-clusters}
\end{figure}

In this section, we describe experiments with clustering words using
the procedure described in the previous section. As explained there,
our clustering procedure yields for each value of $\beta$ a set ${\cal
C}_\beta$ of clusters minimizing the free energy $F$, with the model
estimate for the conditional probability of a verb $\context$ given a
noun $\obj$ being
\begin{displaymath}
 \mp(\context|\obj) = \sum_{c\in \cal C_\beta} \pmem \pcen,
\end{displaymath}
where $\pmem$ depends on $\beta$.  Recall that the pair
$(\obj,\context)$ means that $\obj$ occurred as the head noun of the
direct object of verb $\context$; for example, the pair (thesis,write)
might be extracted from the sentence ``You should write your thesis''.

In our first experiment, we wanted to choose a small set of nouns that
we could be sure bore some relation to one another.  Therefore, we
chose the set $\objects$ to consist of the 64 nouns appearing most
frequently as heads of direct objects of the verb ``fire'' in 
the Associated Press newswire for 1988. In this corpus, the chosen nouns
appeared as direct object heads of a total of 2147 distinct verbs, so
each noun was represented by a density over  2147 verbs.

Figure \ref{fire-clusters} shows the five words most similar to the
cluster centroid for the four clusters resulting from the first
two cluster splits, along with the KL divergences from the centroids.
It can be seen that the first split separates the objects corresponding to
the weaponry sense of ``fire'' (cluster 1) from the ones corresponding
to the personnel action (cluster 2). The second split then further
refines the weaponry sense into a projectile sense (cluster 4) and a
projector (of projectiles) sense (cluster 3). That split is somewhat
less sharp, perhaps because not enough distinguishing contexts occur
in the corpus.  Notice that ``rocket'' is close to both centroids 3
and 4 and therefore has a high probability of belonging to {\em both}
classes: our ``soft'' clustering scheme allows this type of ambiguity.
Note that the ``senses'' we refer to are our own designations for the
clusters -- the algorithm does not decide what the sense(s) of a
cluster actually are.

\begin{figure*}
\epsfscaledbox{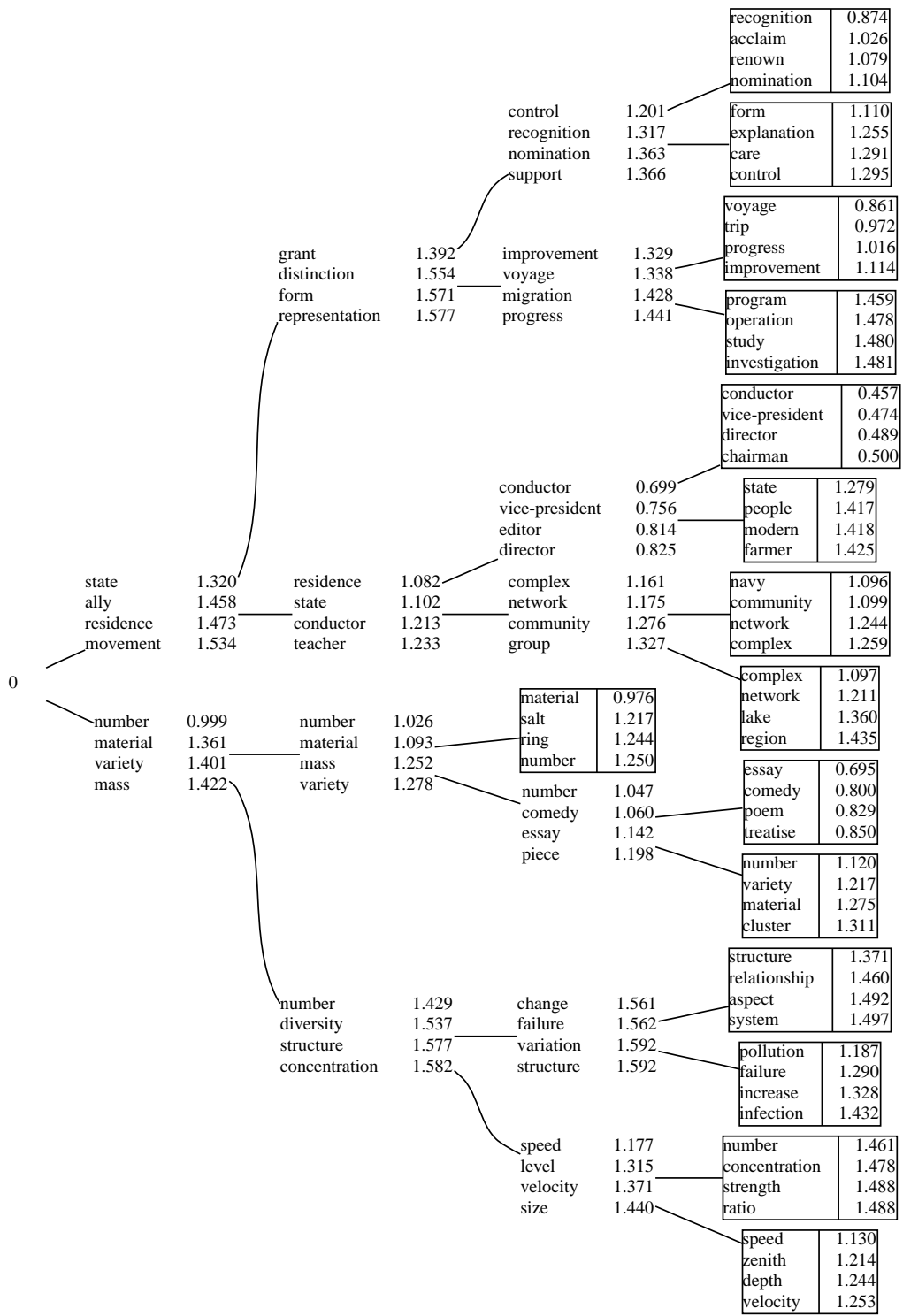}{5.5in}
\caption{Noun clusters for Grolier's encyclopedia}
\label{grol}
\end{figure*}

Our second experiment was performed on a bigger data set: we used
object-verb pairs involving the 1000 most frequent nouns in the June
1991 electronic version of Grolier's Encyclopedia (10 million words).
Figure \ref{grol} shows the four closest nouns for each centroid in a
set of hierarchical clusters derived from this corpus.  Again, we
notice that the clusters and cluster splits often seem to correspond
to natural sense distinctions.  We also observe that a general word
like ``number'' is close to quite a few cluster centroids.

\section{Model Evaluation}
\label{sec:clust-eval}

The preceding qualitative discussion provides some indication of what
aspects of distributional relationships may be discovered by
clustering. However, we also need to evaluate clustering more
rigorously as a basis for models of distributional relationships. 
We now look at two kinds of measurements of model quality: (i)
KL divergence between held-out data and the asymmetric model, and
(ii) performance on the task of deciding which of two verbs is more
likely to take a given noun as direct object when the data relating
one of the verbs to the noun has been withheld from the training data.

\begin{figure*}
\epsfscaledbox{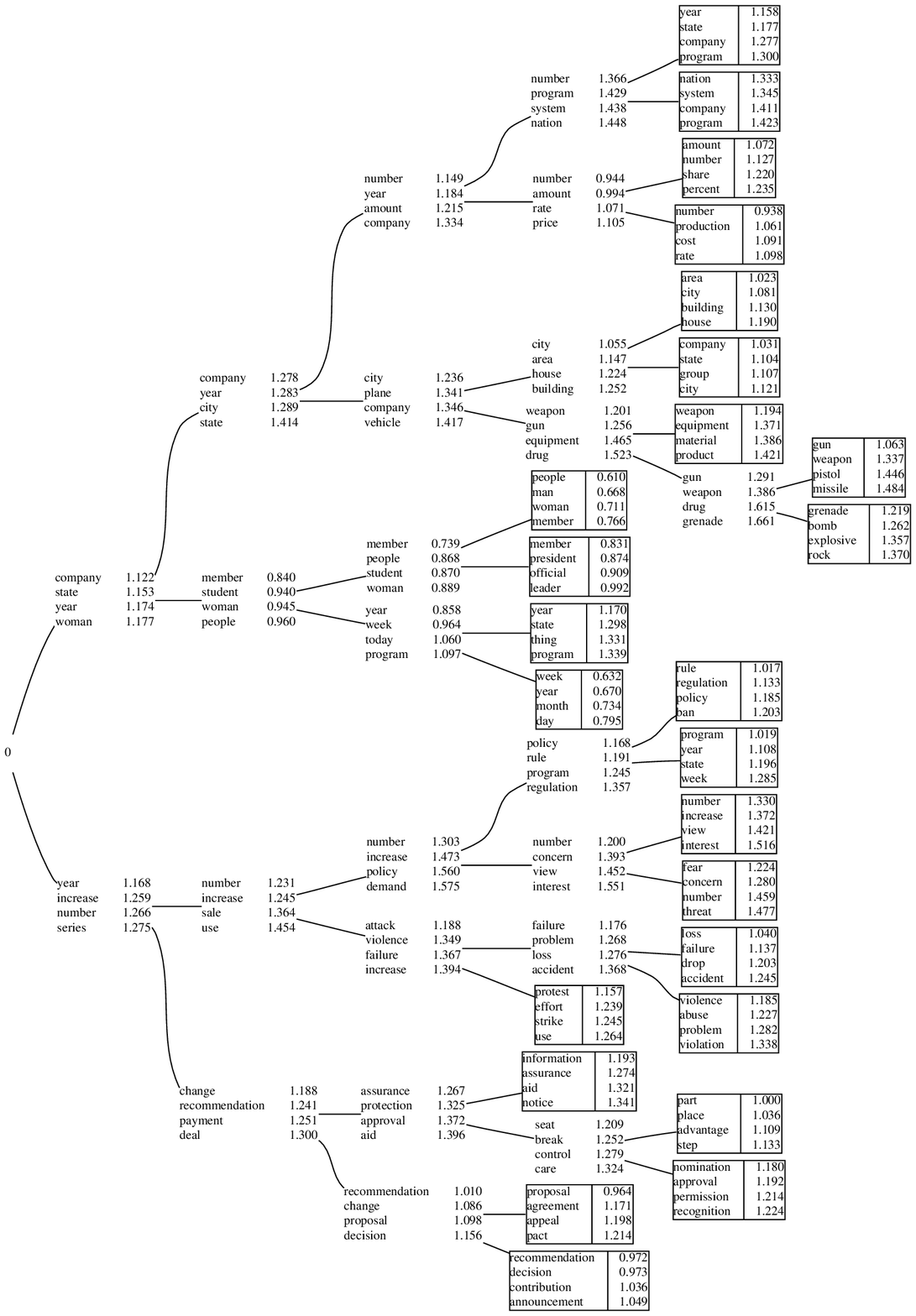}{5.5in}
\caption{Noun clusters for 1988 Associated Press newswire}
\label{ap}
\end{figure*}

The evaluation described below was performed on a data set extracted
from 44 million words of 1988 Associated Press newswire by using the
pattern-matching techniques mentioned earlier. This collection process
yielded 1112041 verb-object pairs. We then selected the subset
involving the 1000 most frequent nouns in the corpus for clustering,
and randomly divided it into a training set of 756721 pairs and a test
set of 81240 pairs.  Figure
\ref{ap} shows the closest nouns to the cluster centroids in an early
stage of the hierarchical clustering of the training data.

\subsection{KL Divergence}

\begin{figure}
\epsfscaledbox{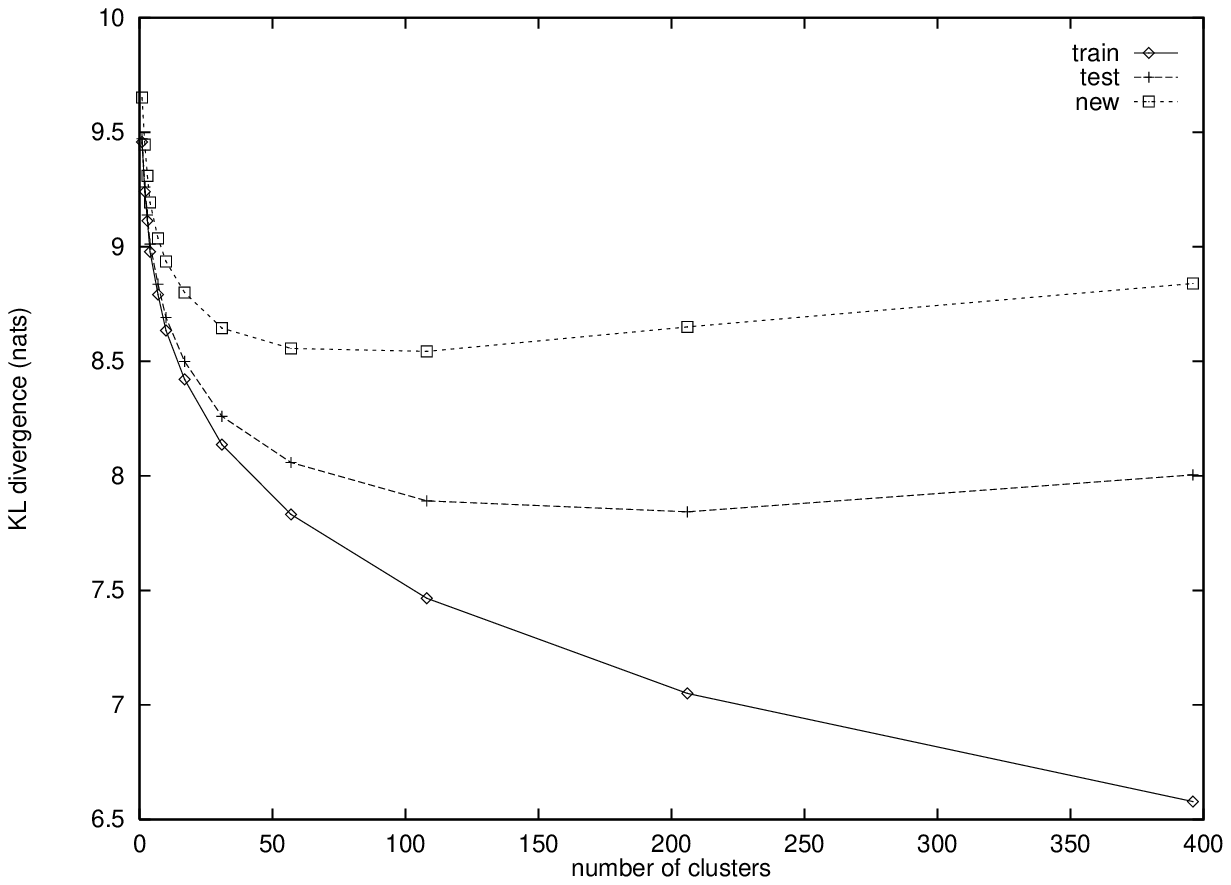}{4in}
\caption{Model evaluation, 1988 Associated Press  object-verb pairs}
\label{rel-entropy}
\end{figure}

Figure \ref{rel-entropy} plots the aggregate KL divergence of several
data sets to cluster models of different sizes; the higher the KL
divergence, the worse the coding inefficiency of using the cluster
model.   The aggregate KL divergence is given by
\begin{displaymath}
\sum_{\obj} D(\pmle(\cdot| \obj)||\mp(\cdot | \obj)). 
\end{displaymath}
For each critical value of $\beta$, we
show the aggregate KL divergence with respect to the cluster model
based on $C_\beta$ for three sets:  the training set (set {\em train}), a
randomly selected held-out test set (set {\em test}), and a set of held-out
data for a further 1000 nouns that were not clustered (set {\em new}).

Not surprisingly, the training set
aggregate divergence decreases monotonically. The test set aggregate
divergence decreases to a minimum at 206 clusters and then starts
increasing, which suggests that the larger models are overtrained.

The new noun test set is intended to evaluate whether clusters based on
the 1000 most frequent nouns are useful classifiers for the
selectional properties of nouns in general.  We characterize each new
noun $\obj$ by its maximum likelihood distribution
$\pmle^{\mbox{new}}(\cdot|\obj)$ as estimated from the new sample 
(we can't use the training data since the new nouns by definition
don't appear there).  The corresponding cluster membership probabilities
for a new noun then have the form
\begin{displaymath}
\pmem = \exp\left( - \beta D(\pmle^{\mbox{new}}(\cdot|\obj) ||
P(\cdot|c)) \right)/Z_{\obj}
\end{displaymath}
and the model probability estimate is calculated as before.
As the figure shows, the
cluster model provides over one nat of information about the
selectional properties of the new nouns, although the overtraining effect
is even more pronounced than for the held-out data involving the 1000
clustered nouns.

\subsection{Decision Task}
\begin{figure}
\epsfscaledbox{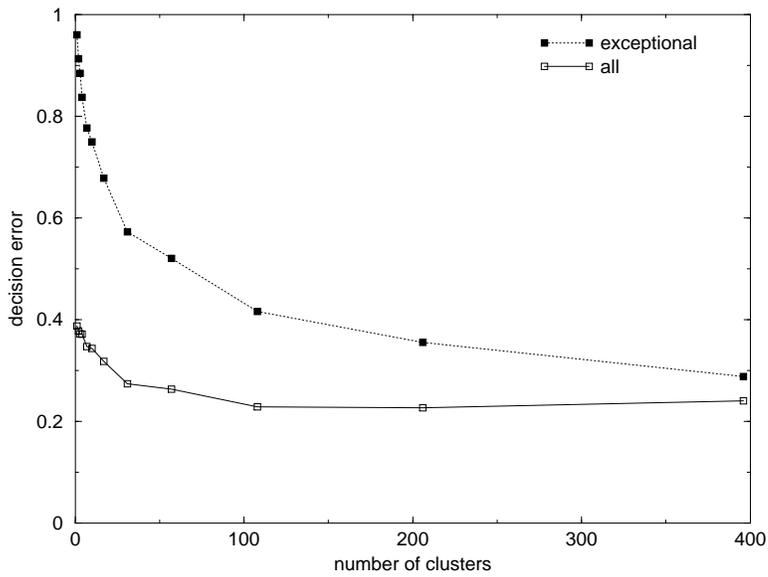}{4in}
\caption{Pairwise verb comparisons, 1988 Associated Press object-verb pairs}
\label{pairwise}
\end{figure}

We also evaluated our cluster models on a verb decision task related to
applications in disambiguation in language analysis. The task
consists of judging which of two verbs $\context$ and $\context'$ is more
likely to take a given noun $\obj$ as object when all occurrences of
$(\obj,\context)$ in the training set were deliberately deleted. Thus,
this test evaluates how well the models reconstruct missing data from
the cluster centroids, since we are interested in cluster models that
can help solve sparse data problems.

The data for this test was built from the training data for the
previous one in the following way, based on an experiment by Dagan,
Marcus, and Markovitch \shortcite{Dagan+Marcus+Markovitch:95a}. We
randomly picked 104 object-verb pairs $(\obj,\context)$ such that verb
$\context$ appeared fairly frequently (between 500 and 5000
occurrences), and deleted all occurrences of such pairs
from the training set.  The resulting training set was
used to build a sequence of cluster models as before.  To create the
test set, for each verb $\context$ in a deleted pair, a {\em confusion
set} $\set{\context,\context'}$ was created.  Then, each model was
presented with the triple $(\context, \obj, \context')$, and was asked
to decide which of $\context$ and $\context'$ is more likely to appear
with a noun $\obj$.

Of course, we need some way of judging correctness without having
access to the true pair probabilities, since the source distribution
for natural language is presumably unknown.  We fall back on the
empirical frequencies to give us a rough estimate of the correct
answer.  Since these frequencies are known not to be entirely accurate
(otherwise, we would have no need of cluster models!), we choose to
create confusion sets for a noun $\obj$ out of pairs of verbs
$\context$ and $\context'$ such that one of the verbs occurred at
least twice as often with $\obj$ than the other in the original data
set (prior to the pair deletion).  Thus, we can be reasonably sure that
whichever verb occurred with $\obj$ more often in the training set
truly has a higher probability of co-occurrence.

In order to evaluate performance, we compare the sign of $\log \left(
\mp(\context|\obj)/\mp(\context'|\obj) \right)$ with that of $\log
\left( \pmle(\context|\obj)/\pmle(\context'|\obj) \right)$ on the
initial data set. The error rate for each model is simply the
proportion of sign disagreements over the test corpus. Figure
\ref{pairwise} shows the error rates for each model on all the
selected $(\context,\obj,\context')$ ({\em all}) and for just those
{\em exceptional} triples in which the log frequency ratio of
$(\obj,\context)$ and $(\obj,\context')$ differs from the log marginal
frequency ratio of $\context$ and $\context'$.

The exceptional cases are especially interesting in that estimation
methods (such as Katz's \backoff\, method) based just on the marginal
frequencies, which the initial one-cluster model represents, would be
consistently wrong.  We see that the cluster model tremendously
outperforms classic estimation methods in the exceptional cases, and
thus has the potential to provide a much better solution to the sparse
data problem.  Furthermore, while some overtraining effects can be
observed for the largest models considered, these effects do not appear
for the exceptional cases.


\section{Related Work}
\label{sec:clust-related}

It is  beyond the scope of this thesis to provide a review
of the entire body of clustering literature; data clustering has
been discussed in fields ranging from statistics to biology.  One list
of journals that publish papers on the subject contains
987 entries \cite{CSNA:journals}; indeed, a summary of various clustering
methods is a thesis in itself
\cite[``substantially this same text was submitted as a
dissertation'', pg. xiii]{Anderberg:73a}.  We therefore narrow our
focus to two subjects: clustering methods appearing in the
natural language-processing  literature (section
\ref{sec:word-clustering}), and other probabilistic clustering
methods (section \ref{sec:sh-clustering}).

\subsection{Clustering in Natural Language Processing}
\label{sec:word-clustering}

Quite a few methods for distributional clustering have appeared in the
literature of the natural language processing community, although to
the best of our knowledge, our work is the first to use soft
clustering in a language-processing context.  The algorithms we will
describe here algorithms fall into two categories: those that seek to
find classes corresponding to human concepts, and those that create
classes for the purpose of improving language modeling.

As an aside, we note that these two categories correspond to two
orthogonal trends in clustering work in general.  The first trend,
readily apparent in recent work on data mining and knowledge
discovery, is to find clusters that are somehow well-formed.  Work in
this vein uses optimization criteria concerning cluster structure; for
instance, our distortion function ${\cal D}$ measures the average
distance between objects and centroids.  The other trend is to find
clusters that aid in the performance of some task; work in this area
uses optimization criteria based on likelihood or some other
performance measure.

\subsubsection{Clustering for Clusters' Sake}
Most of the methods whose end goal is the production of clusters (and
therefore do not test whether the clusterings aid in the performance of some
task) are geared towards
finding either semantic or syntactic classes.  The work of
\namecite{\vhref} (henceforth \vh) is notable because they provide a
way to evaluate the goodness of semantic clusterings; many other
papers (for example, \namecite{Finch+Chater:92a} or
\namecite{Schutze:93a}) merely present example clusters and state that
the derived classes seem to correspond to intuition.

\vh\, describe a hard clustering scheme for grouping
semantically-related adjectives.  They treat adjectives as distributions
over the nouns they modify,  and use Kendall's $\tau$ coefficient
(studied in section \ref{sec:statistics}) to measure the distance between
these distributions.
Their optimization function is one of well-formedness: it rewards
partitions that minimize the average distance between adjectives in
the same cluster.  They carefully delineate a rigorous evaluation
method for comparing clusterings produced by their algorithm against
clusterings produced by human judges,\footnote{One minor criticism:
the number of clusters to create was a parameter given to the system,
whereas the humans were free to choose whatever number of clusters
they wished.}  computing precision, recall, fallout and F-measure
results with respect to an average of the responses given by the
judges, thereby taking into account the fact that humans do not always
agree with each other.

An interesting feature of their work is that they incorporate negative
linguistic similarity information.  By simply observing that
adjectives in the same noun phrase should not, for a variety of
linguistic reasons, be placed in the same class, they get dramatically
better results (17-50\% improvement across the various performance
metrics).  

Some superficial similarities with our clustering work are readily
apparent.  The distributional similarity component of \vh's system
treats adjectives as distributions over nouns, while we treat nouns as
distributions over verbs.  Also, we and \vh\, both used Associated
Press newswire as training data, although \vh\, only used 8.2 million
words, as opposed to our 44 million.  However, our results are
incomparable because our goals differ.  \vh\, explicitly aim to create
classes of semantically related words, and so must solicit human
judgments.  They were therefore constrained by human limitations to
clustering only 21 adjectives. We, on the other hand, are more
interested in clusterings that improve performance and so make use of
a great deal more data.

An independent body of work seeking to build classes 
corresponding to human intuitions is the field of language clustering.
Many researchers in comparative lexicostatistics study the problem of
how to create hierarchical clusterings that correspond to the
evolution and splitting off of languages over time.
\namecite{Black+Kruskal:97a} give a short history and bibliography of
the field.

\subsubsection{Clustering for Language Modeling}

A large number of papers have been written on using class-based models
to improve language modeling (five such papers appear in the
1996 ICASSP proceedings alone
\cite{ICASSP96}).  A common approach is to group words by
their parts of speech.  However, there is no reason to believe that
classifications based on parts of speech are optimal with respect to
language modeling performance,  so we look at papers which present
novel clustering techniques.  Since the methods we discuss all attempt
to create probabilistic models with strong predictive power, it is
not surprising that they are all guided by the maximum likelihood
principle.

The most well-known class-based method is the work by
\namecite{Brown+al:class}.  In their setting, the set of objects and
the set of contexts are the same ($\objects = \contexts = \vocab$);
the pair $(\word_1,\word_2)$ denotes the appearance of the two-word
sequence $\word_1 \word_2$ in the training sample.  Brown et al.
assume a Boolean clustering of the data, so that each word $\word$
belongs only to the class $\class{\word}$, where $\class{\cdot}$ is
the membership function.  Then, their class-based probability
estimate takes the form
\begin{equation}
\mp(\word_2|\word_1) = \mp(\word_2 | \class{\word_2})
\mp(\class{\word_2} | \class{\word_1}).
\label{Brown-prob}
\end{equation}
Given the membership function, the parameters $\mp(\word |
\class{\word})$ and $\mp(\class{\word_2} |
\class{\word_1})$ are determined by sample frequencies, so only the
function $\class{\cdot}$ needs to be estimated.  This is
done by attempting to find class assignments that maximize the average
mutual information $\langle I \rangle$ of the clusters, which in the
limit is equivalent to maximizing the likelihood: if $t_1 t_2 \ldots
t_n$ is the training text, then
\begin{eqnarray*}
L_c & =&  \frac{1}{n-1} \log \mp(t_2 \ldots t_n | t_1) \\
  & \approx & -H(t)+  \langle I \rangle,
\end{eqnarray*}
where $H(t)$ is the entropy of the unigram (single word) distribution,
which we can consider to be fixed.

A serious problem Brown et al. face is that they do not have a way to
calculate good estimates for $\class{\cdot}$.  Therefore, in each
iteration step of their agglomerative clustering algorithm, they are
forced to try many different merges of classes to find the one
yielding the best improvement in $\langle I \rangle$.  After some
amount of care, they are able to derive an algorithm that takes
$O(|\vocab|^3)$ time in each iteration step, whereas in the same
setting our iteration steps would take $O(|\clusters||\vocab|^2)$ time,
which is a significant savings if the number of clusters is small
relative to the number of words.  Also, once the desired number of
clusters has been achieved, Brown et al. shift words from cluster to
cluster in order to compensate for premature groupings of words in the
same class -- this is the rigidity problem referred to in the
quotation from Kaufman and Rousseeuw earlier in this chapter.  Since
we create a soft clustering, we never have to compensate for words
being incorrectly classed together.  At any rate, Brown et al.'s
method potentially involves much wasted computation
since both good and bad merges and shifts must be tried, whereas we
are guaranteed that each step we take reduces the free energy.

Brown et al. do present an alternative algorithm which spends
$O(|\clusters|^3)$ time in each iteration.  This algorithm sorts the
words by frequency and puts the top $k$ into their own classes.  Each
iteration step consists of adding the next most frequent word yet to
be clustered as a new class and then finding the best merge among the
new set of classes; when this merge is taken, the system once again
has $k$ clusters. On the
other hand, it is possible that this heuristic narrows the search down
so much that good classings are missed.  This may well explain the
small perplexity reduction achieved by Brown et al.'s method on the
Brown corpus (from 244 to 236 using a model that interpolates the
class-based model with word-based estimators).

Another commonly-cited class-based language-modeling method, that of
\namecite{Kneser+Ney:93a}, is presented by \namecite{Ueberla:94a}.  In many
respects, Kneser and Ney's work is quite similar to that of Brown et
al.  The same probability model (\ref{Brown-prob}) is used, and some
of the heuristics employed to speed up calculations are the same as
well.  However, their optimization criterion differs, although it,
too, is derived via the maximum likelihood principle.  Instead of an
agglomerative clustering algorithm, Kneser and Ney start with the
desired number of clusters, so that the only operation undertaken to
improve the clustering is to move words from one
cluster to another, searching for the move that makes the biggest
improvement. The running time of each such iteration step is
$O(|\vocab|\cdot(|\vocab| + |\clusters|^2))$.

Ueberla reports that Kneser and Ney's method achieves 
perplexity improvements of up to 25\% on Wall Street Journal data
with respect to Katz's \backoff\, method.  This is a rather
stunning result.  However, the class-based model uses a smoothing
method known as absolute discounting
\cite{Ney+Essen:93a}.  An interesting question is how much of the
performance is due to the smoothing method and how much is due to
the clustering (Brown et al. did not smooth the data); no comparison
was done between the class-based method and the absolute discounting
method.

\subsection{Probabilistic Clustering Methods}
\label{sec:sh-clustering}

One of the first papers to discuss the notion of probabilistic
clustering is that of
\namecite{Ruspini:70a}, who was inspired
by Zadeh's work on fuzzy sets \cite{Zadeh:65a}.  His method attempts
to find membership probabilities (which he calls ``degrees of
belongingness'') that optimize certain well-formedness conditions
similar to our distortion function ${\cal D}$.  However, he does not
attempt to mathematically derive good estimates, so the search for
good parameter settings consists of repeatedly altering one membership
probability while keeping all the others fixed.  Furthermore, his
method relies on distances between objects, rather than between
objects and average distributions.  This poses no problem in his case
because he only considers artificial problems where the true distances
are known.  In practice, however, estimates of inter-object distances
can be quite sensitive to noise; centroid methods overcome this
problem by averaging together many points.

The {\em fuzzy $k$-means} method \cite{Bezdek:81a}, a generalization
of the $k$-means approach, bears some resemblance to our procedure.
It is a centroid method using the Euclidean distance ($L_2$) as
distance function.  The centroid distributions depend on the 
squares of the membership probabilities:
\begin{displaymath}
\pcen = \frac{\sum_{\obj}  \left(\pmem\right)^2 \pmle(\context|\obj)}{
\sum_{\obj} \left(\pmem\right)^2   },
\end{displaymath}
and the membership probabilities in turn depend on the positions of the
centroids.
The optimization function rewards clusterings
that minimize the distance between objects and
centroids:
\begin{displaymath}
F_{{\sc OPT}} = \sum_{\obj} \sum_{c} \left(\pmem\right)^2 (L_2(\obj,c))^2. 
\end{displaymath}
This is a well-formedness condition rather than a maximum likelihood
criterion, and in fact fuzzy $k$-means is not meant to produce
probability estimates.  It is also not meant to produce a hierarchical
clustering; the number of centroids is kept constant throughout the
iterated estimation process.

The clustering procedures most similar to our own are the
deterministic annealing approaches; these include the work of
\namecite{Rose+al:phase} (which influenced our approach) and
\namecite{Hofmann+Buhmann:97a}.  These both find clusters that
minimize the free energy (\ref{free}).  An important difference is
that they use the squared Euclidean distance ($L_2$), whereas we use the KL
divergence as distance function.  In the distributional setting we
have been considering, using the KL divergence is well-motivated,
whereas it is not entirely clear why the $L_2$ norm would be
meaningful.

Bayesian methods 
\cite{Wallace+Dowe:94a,Cheeseman+Stutz:96a}
combine well-formedness constraints and performance criteria.  They
seek to find the model with the maximum posterior probability given
the data, where the posterior probability is based on the product of
the model prior and the likelihood the model assigns to the data.  The
prior is based on the structure of the cluster system, and in general
encodes a bias for fewer clusters; such a prior serves to balance out
the tendency of maximum likelihood criteria to reward systems that
have a large number of clusters.  This is analogous to our inclusion
of a maximum entropy condition in the derivation of our method, since
the maximum entropy criterion also tends to favor having fewer
clusters.

The methods of \namecite{Wallace+Dowe:94a} and
\namecite{Cheeseman+Stutz:96a} do not yield cluster hierarchies
because the number of clusters is allowed to fluctuate from iteration
step to iteration step.  The ``class hierarchy'' described by
\namecite{Hanson+Stutz+Cheeseman:91a} does not consist of classes but
rather of attributes: each node in the dendrogram represents a
collection of parameter settings inherited by all the descendents of
that node.

\section{Conclusions}

We have described a novel clustering procedure for probability
distributions that can be used to group words according to their
participation in particular grammatical relations with other words.
Our method builds a hierarchy of probabilistic classes using an
iterative algorithm reminiscent of EM.  The resulting clusters are
intuitively informative, and can be used to construct class-based word
coocurrence models with substantial predictive power.

While the clusters derived by the proposed method seem in many cases
semantically significant, this intuition needs to be grounded in a
more rigorous assessment. In addition to evalutions of predictive power 
of the kind we have already carried out, it might be worthwhile to compare
automatically-derived clusters with human judgements in a suitable
experimental setting, perhaps the one suggested by \namecite{\vhref}.  In
general, however, the development of methods for directly measuring
cluster quality is an open research area; the problem is compounded
when one takes hierarchical clusterings into account.

Another possible direction to take would be to move to other domains.
For instance, document clustering has been studied by many researchers
in the field of information retrieval.  Recently, there has been
renewed interest in using document clustering as a {\em browsing} aid
rather than a search tool (see \namecite{Cutting+al:92a} for a short
discussion), and also as a way to organize documents (Yahoo!
\shortcite{yahoo} provides a hierarchical clustering in which documents can
appear in more than one class).  In situations where $|\objects|$ and
$|\contexts|$ are very large, our clustering algorithm may be somewhat
slow; however, the extra descriptive power provided by our {\em
probabilistic} clustering may well be worth the extra computational
effort.

\setcounter{chapter}{3}
\chapter{Similarity-Based Estimation}
\label{ch:wsd}

In the previous chapter we looked to cluster centroids as a source of
information when data on a specific event ${\cal E}$ was lacking.  
This chapter introduces an alternative  model,
where we instead
look to
the events most similar to ${\cal E}$.  For convenience, we will refer
to the new type of model as {\em similarity-based}, although our
clustering method of the preceding chapter also made use of the notion
of similarity.

\section{Introduction}

In the previous chapter, we described a method for automatically
clustering distributional data, and showed that we can use the
clusters so derived to construct effective models for predicting
probabilities in situations where data is lacking.  The clustering
method was divisive: the system started with just one cluster
centroid, and as the temperature was slowly lowered, phase transitions
caused cluster centroids to split.  This splitting of centroids meant
that the number of clusters $k$ did not have to be determined
beforehand; rather, all possible numbers of clusters could be
considered in a fairly efficient fashion, and the best configuration
could be chosen via cross-validation.  For clustering algorithms that
keep the number of clusters constant throughout the estimation
process, the only way to try out many different numbers of clusters is
to re-run the algorithm with a different value of $k$ each time.  But
it is generally the case for these algorithms that the results of the
computation for one $k$ cannot be used to aid the computation for a
different $k$, so the search for the right $k$ is not very efficient.

An interesting alternative is a nearest-neighbor approach, where given
an event ${\cal E}$ whose probability we need to estimate, we consult
only those events that are most similar to ${\cal E}$.  In a sense,
we allow each event to form the centroid of its own class, and thus
avoid having to find the right number of clusters.  While this
approach does not reduce the size of the model parameter space as
class-based approaches do, it avoids the over-generalization that
class-based models can fall prey to. As 
\namecite{Dagan+Marcus+Markovitch:95a} argue, using class information
to model specific events may lead to too much loss of
information.  Probabilistic clusterings ameliorate this problem
somewhat by combining estimates from different classes, using
membership probabilities to weight the class estimates appropriately,
but the concern about over-generalization is still valid.

\begin{figure}
\unitlength1in
\hspace{2.3in}
\epsfscaledbox{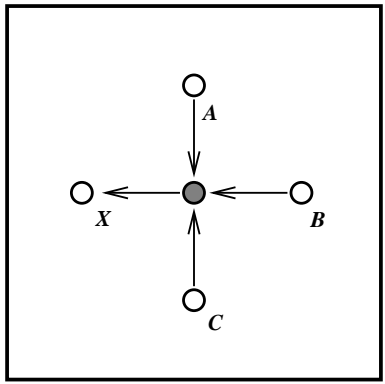}{2in} 
\caption{Centroid overgeneralization}
\label{fig:over}
\end{figure}

We present a small example to make the over-generalization problem
clearer.  Figure \ref{fig:over} shows a situation in which there are
four objects (the empty circles) and 
only one centroid (the grey circle in the middle).  Let us assume we are
trying to model the behavior of $X$.
In a cluster-based centroid model, the estimate for the behavior of
$X$ depends on the behavior of the centroid; this dependence is
indicated by the arrow from the centroid to $X$.  However, 
the behavior of the centroid
is an average of the behaviors of all the other points, including $A$,
$B$, and $C$, as indicated by the arrows pointing to the centroid.
Therefore, an estimate for $X$ depends not only on $A$ and $C$, which
are relatively close to it, but also on $B$, which is much farther
way.  It might make more sense to use only points $A$ and $C$ in
trying to estimate $X$.

We therefore turn our attention in this and the next chapter to
similarity-based language modeling techniques that do not require
building general classes.  While our cluster model, described in the
previous chapter, estimated the conditional probability
$P(\context|\obj)$ of an object-context pair by averaging together
class estimates $P(\context|c)$, weighting the evidence of each class
by the degree of association $P(c|x)$ between $x$ and $c$:
\begin{displaymath}
\mp(\context|\obj) =
   \sum_{c \in \clusters} \pmem \pcen,
\end{displaymath}
our new object-centered model replaces the centroids by other objects:
\begin{equation}
\mp(\context|\obj) =
   \sum_{\obj'} f(\obj,\obj') P(\context|\obj'),
\label{gen-sim}
\end{equation}
where $f(\obj,\obj')$ depends on the similarity between $\obj$ and
$\obj'$.

We are not the originators of equation (\ref{gen-sim}).
Similarity-based estimation was first used for language
modeling in the {\em cooccurrence smoothing} method of
\namecite{Essen+Steinbiss:92a}, derived from work on acoustic model
smoothing by \namecite{Sugawara+al:85a}.  \namecite{Karov+Edelman:96a}
develop a similarity-based disambiguation method that also can be fit
into the framework of equation (\ref{gen-sim}); however, since their
method does not estimate probabilities and relies on a similarity
function that is calculated via an iterative process, we will not give
further consideration to their work here.

In this chapter we establish proof of concept: we discuss and compare
ways to instantiate equation (\ref{gen-sim}), using a simple decision
task for evaluation purposes.  The KL divergence will once again prove
to be an effective measure of dissimilarity.  In the next chapter we
evaluate a similarity-based model on more true-to-life tasks that test
the utility of our method for speech
recognition; we use a more complicated version of the model presented
here, incorporating several heuristics in order to speed up the
computation.

\section{Chapter Overview}

As in the previous chapter, our goal is to estimate the (conditional)
probability of object-context pairs $(\obj,\context) \in \objects
\times \contexts$. Our first concern in this chapter is to describe
similarity-based estimation methods in general.  In section
\ref{sec:format} we develop a common framework for these methods, so
that the only parameter that varies from method to method is the
similarity function used.  In the following section (\ref{sec:simfns})
we describe various similarity functions.  The majority are based on
distance functions studied in chapter
\ref{ch:sim}, but we also discuss the {\em confusion probability},
which appears in the work of
\namecite{Essen+Steinbiss:92a}.  

The second part of this chapter describes our evaluation of
similarity-based methods.  In section \ref{sec:task-descr}, we
introduce the problem of {\em pseudo-word disambiguation}, a task
which is related to the usual word sense disambiguation problem, but
presents many advantages in terms of ease of experimentation.  After a
discussion of the data used to construct basic language models and a
comparison of these basic models (\ref{sec:wsd-data}), we look at a
few examples to get a qualitative sense for how the different
similarity functions perform (\ref{sec:sampleclose}).  Finally,
section \ref{sec:wsd-eval} presents five-fold
cross-validation results for the similarity-based methods and for several
baseline models.  Our tests show that indeed, similarity information
can be quite useful in sparse data situations.  In particular, we
found that all the similarity-based methods performed almost 40\%
better than \backoff\, if unigram frequency was eliminated from being
a factor in the decision.

An interesting phenomenon we observe  is that the effect of
removing extremely rare events from the training set is quite
dramatic when similarity-based methods are used.  We found that, contrary
to a claim made by Katz that such events can be discarded without
hurting language model performance \cite{Katz:87a}, similarity-based
smoothing methods suffer noticeable performance degradation when
singletons (events that occur exactly once) are omitted.

Throughout this chapter, the base of the logarithm function is 10.
\section{Distributional Similarity Models}
\label{sec:format}

A similarity-based language model consists of three parts: a scheme
for deciding when to use similarity-based information to determine the
probability of a word pair, a method for combining information from
similar words, and, of course, a function measuring the similarity
between words.  We give the details of each of these three parts in
the following three sections.  

\subsection{Discounting and Redistribution}
\label{sec:redistribute}

We hold that it is best to always use the
most specific information available.  While the maximum likelihood
estimate (MLE)
\begin{displaymath}
\pmle(\context |\obj) = \frac{\counts(\obj,\context)}{\counts(\obj)}
\end{displaymath}
(equation (\ref{MLE-general}) from chapter \ref{ch:sim}, where
$\counts(z)$ is the number of times event $z$ occurred in the training
data) yields a terrible estimate in the case of an unseen word pair,
it is pretty good when sufficient data exists.  Therefore, Katz's
\shortcite{Katz:87a} implementation of the Good-Turing discounting
method, described in chapter \ref{ch:sim}, provides an attractive
framework for similarity-based methods; it uses the (discounted) MLE when
the pair $(\obj,\context)$ occurs in the data, and a different
estimate if the pair does not occur:
\begin{displaymath}
\hat{P}(\context| \obj) = \left\{\!\!\!\!
\begin{array}{l@{\hspace{0.6ex}}l}
P_d(\context | \obj) & \mbox{if $\counts(\obj,\context) > 0$} \\
\alpha(\obj)P_r(\context | \obj) & \mbox{otherwise  ($(x,y)$ is unseen)}
\end{array}\right.\;. \qquad \qquad  \qquad  \qquad (\ref{genmodel}) 
\end{displaymath}
Recall that equation (\ref{genmodel})
actually represents a modification of Katz's formulation: we have
written $P_r(\context|\obj)$ where Katz has $P(\context)$.  This
allows us to use similarity-based estimates for unseen word pairs,
rather than simply backing off to the probability of the context
$\context$.  Observe that this formulation means that we will use the
similarity estimate for unseen word pairs only, as desired.

We next investigate estimates for $P_r(\context| \obj)$ derived by
averaging information from objects that are distributionally similar to
$\obj$.

\subsection{Combining Evidence}
\label{sec:combine}

The basic assumption of a similarity-based model is that if object
$\obj'$ is ``similar'' to object $\obj$, then the behavior of $\obj'$
can yield information about the behavior of $\obj$.  When data on
$\obj$ is lacking, then, we average together the distributions of
similar objects, weighting the information furnished by a particular
$\obj'$ by the similarity between $\obj'$ and $\obj$.

More precisely, let $W(\obj,\obj')$ denote an increasing function of
the similarity between $\obj$ and $\obj'$; that is, the more similar
$\obj$ and $\obj'$ are, the larger $W(\obj,\obj')$ is.  Let ${\cal
S}(\obj)$ denote some set of objects that are most similar to $\obj$
(we discuss the exact form of ${\cal S}(\obj)$ in the next paragraph).
Then, the general form of similarity model we consider is a
$W$-weighted linear combination of predictions of similar objects:
\begin{equation}
\psim(\context|\obj) = \sum_{\obj' \in {\cal S}(\obj)}
\frac{W(\obj,\obj')}{\sum_{\obj' \in {\cal S}(\obj)}W(\obj,\obj')}{P(\context|\obj')}.
\label{sim-formula}
\end{equation}
Observe that according to this formula, we predict that $\context$ is
likely to occur with $\obj$ if it tends to occur with objects that are
very similar to $\obj$.

Considerable latitude is allowed in defining the set ${\cal S}(\obj)$.
\namecite{Essen+Steinbiss:92a} and Karov and
Edelman \shortcite{Karov+Edelman:96a} (implicitly) set ${\cal S}(\obj) =
\objects$.  However, if $\objects$ is very large, it is desirable to
restrict ${\cal S}(\obj)$ in some fashion, so that summing over all
$\obj' \in \objects$ is not too time-consuming.  In the next chapter,
we will consider various heuristics for choosing a small set of
similar words.  These heuristics include setting a limit on the
maximum size of ${\cal S}(\obj)$, and only allowing an object $\obj'$
to belong to ${\cal S}(\obj)$ if the dissimilarity between $\obj$
and $\obj'$ is less than some threshold value.  We will show some
evidence at the end of this chapter that limiting the size of the set
of closest objects does not greatly degrade performance, at least for
the best similarity-based models.

The approach taken in this chapter is to use $\psim$ as the probability
redistribution model in equation (\ref{genmodel}), i.e.,
$P_r(\context | \obj) = \psim(\context | \obj)$.
In the next chapter we discuss a variation in which $P_r$ is a linear
combination of $\psim$ and another estimator. 

\section{Similarity Functions}
\label{sec:simfns}

The final step in defining a similarity-based model is to choose what
similarity function to use.  We first look in section \ref{sec:dist-based}
at three functions from chapter \ref{ch:sim} that measure the distance
between distributions.  For each of these functions, it is necessary
to define a weight function $W(\obj,\obj')$ that ``reverses the
direction'' of the distance function, since we need weights that have
larger values when the distributions are less distant.  Section
\ref{sec:confusion} 
describes some of the properties of the {\em confusion probability},
which was used to achieve good performance results by
\namecite{Essen+Steinbiss:92a}.
Section \ref{sec:base} discusses the base language models from which
object distributions are computed, and also summarizes some properties
of the four similarity functions we will compare.

Regardless of which similarity function is chosen, in order to make
the computation of equation (\ref{sim-formula}) efficient it is
useful to compute the $|\objects| \times |\objects|$ matrix of
similarities $W(\obj_i,\obj_j)$ or distances $d(\obj_i,\obj_j)$ (for arbitrary
distance functions $d$) beforehand.
\subsection{Distance Functions}
\label{sec:dist-based}

In chapter \ref{ch:sim}, we studied several functions measuring the
distance between probability distributions.  These included the KL
divergence (section \ref{sec:KL})
\begin{displaymath}
D(\obj || \obj') = \sum_{\context} P(\context | \obj) \log
\frac{P(\context | \obj)}{P(\context | \obj')},
\end{displaymath}
the total divergence to the mean (section \ref{sec:KL+})
\begin{displaymath}
A(\obj,\obj') \defas D(\obj || \frac{\obj + \obj'}{2}) + D(\obj' ||
\frac{\obj + \obj'}{2})
\end{displaymath}
($(\obj + \obj')/2$ denotes the probability mass function
$(P(\cdot|\obj) + P(\cdot|\obj'))/2$), and the $L_1$ norm (section \ref{sec:geometric})
\begin{displaymath}
\Lone(\obj,\obj') \defas \sum_{\context} | P(\context |\obj) -
P(\context| \obj')|.
\end{displaymath}
Since these functions are all distance functions, they {\em decrease}
when the similarity between $\obj$ and $\obj'$ increase.  However, we
desire weight functions $W(\obj, \obj')$ that are {\em increasing} in
the similarity between $\obj$ and $\obj'$.  

In the case of the KL divergence $D$, we set
$W(\obj,\obj')$ to be 
\begin{displaymath}
W_D(\obj,\obj')=10^{-\beta D(\obj||\obj')}.
\end{displaymath}
$\beta$ is an experimentally-tuned parameter controlling the relative
influence of the objects closest to $\obj$: if $\beta$ is high, then
$W(\obj,\obj')$ is non-negligible only for those $\obj'$ that are
extremely close to $\obj$,
whereas if $\beta$ is low, objects that are somewhat distant from
$\obj$ also contribute to the estimate.
The choice of a negative exponential form
is motivated by the fact that the probability of drawing an
i.i.d. sample of size $n$ with empirical distribution $P$ from a
multinomial $Q$ is $10^{-nD(P||Q)}$ to first order in the exponent --
this is theorem \ref{th:types} from section
\ref{sec:KL}.  

When the distance function is the total divergence to the mean, we
also use a negative exponential:
\begin{displaymath}
W_A(\obj,\obj')=10^{-\beta A(\obj,\obj')}.
\end{displaymath}
Again, $\beta$ controls the relative importance of the most similar
objects and is determined experimentally.

Finally, we define the weight function for the $L_1$ norm to be
\begin{displaymath}
W_{L_1}(\obj,\obj')= (2 - \Lone(\obj,\obj'))^\beta,
\end{displaymath}
with $\beta$ playing the same role as in $W_D$ and $W_A$ above
(we tried using the exponential form $10^{-\beta L_1(\obj,\obj')}$,
but $(2 - \Lone(\obj,\obj'))^\beta$ yielded better performance results).

We have made no attempt to normalize these various weight functions,
so they take on different sets of values; for example, $W_D(\obj,
\obj)= W_A(\obj, \obj)= 1$, but $W_L(\obj, \obj) =2^\beta$.
Normalization is not necessary because our evaluation task ignores
scale factors.

\subsection{Confusion probability}
\label{sec:confusion}

Essen and Steinbiss \shortcite{Essen+Steinbiss:92a} introduced {\it
confusion probability} in the context of cooccurrence smoothing for
language modeling. Cooccurrence smoothing was also applied by
\namecite{Grishman+Sterling:93a} to the problem of estimating the likelihood of
selectional patterns.

Of the four similarity-based models Essen and Steinbiss consider, we
choose to describe and implement  model 2-B (equivalent to
model 1-A) because it was found to be the best performer of the four.
Indeed, Essen and Steinbiss report test-set perplexity reductions of
up to 14\% on small corpora.  Although they used an interpolation
framework, where the similarity-based estimate was linearly
interpolated with other estimators for seen as well as unseen events,
we will for the sake of uniformity incorporate the confusion
probability into the \backoff-like framework of equation
(\ref{genmodel}).

The confusion probability represents the likelihood that object
$\obj'$ can be substituted for object $\obj$; it is based on the
probability  that $\obj$ and
$\obj'$ are found in the same contexts:
\begin{equation}
\conf(\obj' | \obj) = \sum_{\context} \frac{P(\obj
| \context) P(\obj' | \context) P(\context)} {P(\obj)}
\label{conf}
\end{equation}
(the term $P(\obj)$ is required to ensure that $\sum_{\obj'}
\conf(\obj'|\obj) = 1$).
Since this expression incorporates both conditional probabilities and
marginal probabilities, it is not a measure of the distance between
two distributions as are the functions described in section
\ref{sec:measures}.

The confusion probability is symmetric in the sense that
$\conf(\obj'|\obj)$ and $\conf(\obj|\obj')$ are identical up to frequency
normalization: $\frac{\conf(\obj'|\obj)}{\conf(\obj|\obj')} =
\frac{P(\obj)}{P(\obj')}$. Unlike the measures described above, $\obj$ may not
be the ``closest'' object to itself, that is, there may
exist an object $\obj'$ such that $\confarg  > \conf(\obj|\obj)$, as
we shall see in section \ref{sec:sampleclose}.

Further insight into the behavior of $\conf$ is gained by using
Bayes' rule to rewrite expression (\ref{conf}):
\begin{eqnarray*}
\confarg & = & \sum_{\context} \frac{1}{P(\obj)} 
\left(\frac{P(\context|\obj) P(\obj)}{P(\context)} \right)
\left( \frac{P(\context|\obj') P(\obj')}{P(\context)} \right) P(\context)\\
&=& \sum_{\context} \frac{P(\context|\obj)}{P(\context)} 
P(\context| \obj') P(\obj').
\end{eqnarray*}
\noindent This form reveals another important difference between the
confusion probability and the functions $D$, $A$, and $\Lone$ described
above.  The latter three functions rate $\obj'$ as similar to $\obj$ if, roughly,
$P(\context|\obj')$ is high when $P(\context | \obj)$ is.  $\confarg$,
however, is greater for those $\obj'$ for which $P(\obj',
\context)$ is large when $P(\context|\obj)/P(\context)$ is.  Notice
that the case when the ratio
$P(\context|\obj)/P(\context)$ is large
contradicts the back-off assumption that $P(\context)$ is a good
estimate of $P(\context|\obj)$ when the pair $(\obj,\context)$ is unseen.

While the fact that $\conf$ is called a probability
implies that it ranges between $0$ and $1$, some elementary
calculations show that in fact its maximum value is $\frac{1}{2}
\max_{\context} P(\context)$. 
Following Essen and Steinbiss, we
choose the weight function $W(\obj,\obj')$ to be the confusion
probability itself without including the scale parameter $\beta$.

\subsection{Base Language Models}
\label{sec:base}

Throughout the above discussion, we have blithely referred to
the quantities $P(\context|\obj)$, $P(\obj)$, and $P(\context)$
without explaining where these quantities actually come from.  
These must be provided by some base language model $P$, but it turns out
that there is some subtlety as to the form the base language model may
take.

As discussed in section \ref{sec:KL}, the KL divergence
$D(\obj||\obj')$ is undefined if there exists a context $y$ such that
$P(\context|\obj)$ is greater than zero but $P(\context|\obj')$ is
zero.  This argues for a language model that is smoothed so that
$P(\context|\obj')$ cannot be zero.  A natural choice is to use the 
\backoff\, estimate, so that
$P(\context|\obj) = \pbo(\context|\obj)$, where $\pbo$ is given by
equation (\ref{backoff}).

However, the normalization of the confusion probability (\ref{conf})
requires that the base language model be consistent with respect to
joint and marginal probabilities, that is, that
\begin{displaymath}
P(\obj) =\sum_{\context} P(\context|\obj) P(\obj).
\end{displaymath}
Unfortunately, the \backoff\, estimate does not have this property,
since it discounts conditional probabilities without altering the
marginals.  Therefore, we use the maximum likelihood estimate as the
base language model for $\conf$: $P(\context|\obj) =
\pmle(\context|\obj)$

Thus, we cannot directly compare the performances of all four of the
similarity-based models defined above because they require different
base language models.  In the experimental results section of this chapter,
then, we will evaluate the total divergence to the
mean, the $\Lone$ norm, and the confusion probability, using $\pmle$
as the base language model.  Chapter \ref{ch:pp} describes experiments
where the KL divergence is used as the distance function and the
\backoff\, estimate is used as the base language model.

Several features of the measures of similarity listed above are
summarized in table \ref{table:simsum}. ``Base LM constraints'' are
conditions that must be satisfied by the probability estimates of the
base language model.  The last column indicates whether the weight
$W(\obj, \obj')$ associated with each similarity function depends on a
parameter that needs to be tuned experimentally.

\begin{table}[ht]
\begin{center}
\begin{tabular}{l|l|c|c}
distance &  range &  base LM constraints & tune? \\ \hline
$D$ & $[0, \infty]$ & $P(\context | \obj') \neq 0$ if $P(\context|\obj) \neq 0$ &
yes \\
$A$ & $[0, 2 \log 2]$ &none & yes\\
$\Lone$ & $[0,2]$ & none & yes\\
$\conf$ & $[0, \frac{1}{2} \max_{\context} P(\context)]$ & Bayes
consistency & no
\end{tabular}
\end{center}
\caption{\label{table:simsum} Summary of similarity function properties} 
\end{table}

\section{Experimental Results}

We evaluated three of the similarity measures described above on a
word sense disambiguation task.  Each method is presented with a noun
and two verbs, and must decide which verb is more likely to have the
noun as a direct object.  Thus, we do not measure the absolute quality
of the assignment of probabilities, as would be the case in a standard
language model evaluation such as  perplexity reduction (defined in the
next chapter) but merely ask that a method be able to distinguish
between two alternatives.  We are therefore able to ignore constant
factors, and so need neither normalize the similarity
measures to lie between $0$ and $1$ nor  calculate
the denominator in equation (\ref{sim-formula}).


\subsection{Pseudo-word Sense Disambiguation}
\label{sec:task-descr}
In the usual word sense disambiguation task, the method to be tested
is presented with an ambiguous word in some context, and is asked to
use the context to identify the correct sense of the word.  For example,
a test instance might be the sentence fragment ``robbed the bank'';
the disambiguation method must decide whether ``bank'' refers to a
river bank, a savings bank, or perhaps some other alternative.

While sense disambiguation is clearly an important task, it presents
numerous experimental difficulties.  First of all, the very notion of
``sense'' is not clearly defined; for instance, dictionaries may
provided sense distinctions that are too fine or too coarse for the
data at hand.  Also, one needs to have training data for which the
correct senses have been assigned, which can require considerable
human effort.

To circumvent these and other difficulties, we set up a pseudo-word
disambiguation experiment \cite{Schutze:92a,Gale+al:92b}, the general
format of which is as follows.  We first construct a list of {\em
pseudo-words}, each of which is the combination of two different words
in $\contexts$.  Each word in $\contexts$ contributes to exactly one
pseudo-word.  Then, we replace each $\context$ in the test set with
its corresponding pseudo-word.  For example, if we choose to create a
pseudo-word out of the words ``make'' and ``take'', we would change
the test data like this:
\begin{center}
\begin{tabbing}
make \= plans \= $\Rightarrow$ \= \{make, take\} plans \\
take \> action \>  $\Rightarrow$ \> \{make, take\} action \\
\end{tabbing}
\end{center}
The method being tested must choose between the two
words that make up the pseudo-word.

The pseudo-word set-up has two attractive features.  First, the
alternative ``senses'' are under the control of the experimenter.
Each test instance presents exactly two alternatives to the
disambiguation method, and the alternatives can be chosen to be of the
same frequency, the same part of speech, and so on.  Secondly, the
pre-transformation data yields the correct answer, so that no
hand-tagging of the word senses is necessary.  These advantages make
pseudo-word experiments an elegant and simple means to test the
efficacy of different language models.


\subsection{Data}
\label{sec:wsd-data}

We ran our evaluation on the same Associated Press newswire data that
we used for the clustering evaluation described in the previous
chapter.  To review, we set $\objects$ to be the 1000 most frequent
nouns in the data; $\contexts$ was the set of transitive verbs
$\context$ that were observed to take a noun in $\objects$ as direct
object.  The extraction of object-verb pairs was performed via regular
pattern matching and concordancing tools \cite{Yarowsky:92a} from 44
million words of 1988 Associated Press newswire, which had been
automatically tagged with parts of speech \cite{Church:88a}.
Admittedly, regular expressions are inadequate for this task; although
we filtered the results somewhat, some bad pairs doubtless
remained.

\begin{table}
\begin{center}
\begin{tabbing}
\hspace*{3cm}\= \quad \=without singletons \hspace*{.1cm} \= 587833 \= para
\= para \kill
\>Training data for base language models \\
\>\> with singletons: \> 587833 \\
\>\> without singletons: \> 505426 \\
\>Parameter tuning and test data \\
\pushtabs
\hspace*{3cm}\=\quad \=$T_1$: \hspace*{.05cm}\= 3434 \hspace*{.2cm}\=
(tuning: \= 13718) \\
\>\>$T_2$: \> 3434 \>(tuning: \>13718) \\
\>\>$T_3$: \> 3434 \>(tuning: \>13718) \\
\>\>$T_4$: \> 3434 \>(tuning: \>13718) \\
\>\>$T_5$: \> 3416 \>(tuning: \>13736) \\
\poptabs
\end{tabbing}
\end{center}
\caption{\label{table:bigramcounts} Number of bigrams in the training,
parameter tuning, and test sets.}
\end{table}

We used $80\%$, or $587833$, of the pairs so derived, for building
base bigram language models, reserving $20\%$ for testing purposes.
As some of the similarity measures to be compared require smoothed
language models, while others do not, we calculated both a Katz
\backoff\, language model ($P = \pbo$) and a maximum likelihood model ($P =
\pmle$).  Furthermore, we wished to investigate
Katz's claim that one can delete {\em singletons}, word pairs that
occur only once, from the training set without affecting model
performance \cite{Katz:87a}; our training set contained $82407$
singletons.  We therefore built four base language models, summarized
in table \ref{table:langmod}.

\begin{table}[ht]
\begin{center}
\begin{tabular}{c|cc}
 	& with singletons (587833 pairs)	& omit singletons
(505426 pairs) \\ \hline
MLE	&  \MLE1			& \MLEo1	\\
\backoff	& \bo1			& \boo1			\\
\end{tabular}
\end{center}
\mycap{\label{table:langmod} Base language models}
\end{table}

Since we wished to test the effectiveness of using similarity
information for unseen word cooccurrences, we removed from the test
set any object-verb pairs that occurred in the training set; this
resulted in $17152$ {\em unseen} pairs (some occurred multiple times).
The unseen pairs were further divided into five equal-sized parts,
$T_1$ through $T_5$, which formed the basis for five-fold
cross-validation: in each of the five runs, one of the $T_i$ was used
as a performance test set, with the other 4 sets combined into one set
used for tuning parameters (if necessary) via a simple grid search.
Finally, test pseudo-words were created from pairs of verbs with
similar frequencies, so as to control for word frequency in the
decision task.  Our measure of performance was the {\em error rate},
defined as
\[
\frac{1}{n} (\mbox{number of incorrect choices } + (\mbox{number of
ties})/2)
\]
where $n$ was the size of the test corpus.  A tie occurs
when the two words making up a pseudo-word are deemed equally likely.

We first look at the performance of the base language models
themselves.  Their error rates are summarized in
table \ref{table:baseline}.  \MLE1 and \MLEo1 both have
error rates of exactly $.5$ because the test sets consist of unseen
bigrams, which are assigned a probability of $0$ by the maximum likelihood
estimate.  Since we chose to form pseudo-words out of verbs of similar
frequencies, the back-off models \bo1 and \boo1
also perform poorly.

\begin{table}[ht]
\begin{center}
\begin{tabular}{l|lllll}
 & $T_1$ & $T_2$ & $T_3$ & $T_4$ & $T_5$ \\ \hline
\MLE1 & .5 & .5 & .5 & .5 & .5 \\
\MLEo1 & \H{ } & \H{ } & \H{ } & \H{ } & \H{ } \\
\bo1 & 0.517 & 0.520 & 0.512 & 0.513 &0.516 \\
\boo1 & 0.517 & 0.520 & 0.512 & 0.513 &0.516 \\
\end{tabular}
\end{center}
\mycap{\label{table:baseline} Base language model error rates}
\end{table}

Since the back-off models consistently performed worse than the MLE
models, we chose to use only the MLE models in our subsequent
experiments.  Therefore, we only ran comparisons between the measures
that could utilize unsmoothed data, namely, the $L_1$ norm,
the total divergence to the mean, and the confusion probability.
It should be noted, however, that on \bo1 data, the KL divergence
performed slightly better than the $L_1$ norm; in the next chapter, we
will study the performance of the KL divergence more carefully.

\subsection{Sample Closest Words}
\label{sec:sampleclose}

In this section, we examine the closest words to a randomly selected
noun, ``guy'', according to the three measures $\Lone$, $A$, and $\conf$.

Table \ref{table:closeMLE1} shows the ten closest words, in order,
when the base language model is \MLE1.  There is some overlap between
the closest words for $\Lone$ and the closest words for $A$, but very
little overlap between the closest words for these measures and the
closest words with respect to $\conf$: only the words ``man'' and
``lot'' are common to all three.  Also observe that the word ``guy'' itself is
only  fourth on the list of words with the
highest confusion probability with respect to ``guy''.

\begin{table}[t]
\begin{center}
\begin{tabular}{ll|ll|ll}
\multicolumn{2}{c}{$L$} &\multicolumn{2}{c}{$A$}
&\multicolumn{2}{c}{$\conf$} \\ \hline
GUY & 0.000000   & GUY & 0.000000   & role & 0.032925 \\
kid & 1.229067   & kid & 0.304297   & people & 0.024149 \\
lot & 1.354890   & thing & 0.329062 & fire & 0.013092 \\
thing & 1.394644 & lot & 0.330871   & GUY & 0.012744 \\
man & 1.459825   & man & 0.350695   & man & 0.011985 \\
doctor & 1.460766& mother & 0.368966& year & 0.009801 \\
girl & 1.479976  & doctor & 0.369644& lot & 0.009477 \\
rest & 1.485358  & friend & 0.372563& today & 0.009095 \\
son & 1.497497   & boy & 0.373881   & way & 0.008778 \\
bit & 1.497502   & son & 0.375474   & part & 0.008772 \\
\multicolumn{2}{l|}{(role: rank 173)}	&
\multicolumn{2}{l|}{(role:  rank 43)} &
\multicolumn{2}{l}{(kid:  rank 80)} \\
\end{tabular}
\end{center}
\caption[Closest words to ``guy'': \MLE1]{\label{table:closeMLE1} 10 closest words to the word ``guy''
for $A$, $L$, and $\conf$, using \MLE1 as the base language model.  The
rank of the words ``role'' and ``kid'' are also shown if they are not
among the top ten.}
\end{table}

Let us examine the case of the nouns ``kid'' and ``role'' more
closely.  According to the similarity functions $\Lone$ and $A$, ``kid'' is
the second closest word to ``guy'', and ``role'' is considered
relatively distant.  In the $\conf$ case, however, ``role'' has
the highest confusion probability with respect to ``guy,'' whereas
``kid'' has only the 80th highest confusion probability.  What
accounts for the difference between $A$ and $\Lone$ on the one hand and
$\conf$ on the other?

Table \ref{table:verbs}, which gives the ten verbs most likely to
occur with ``guy'', ``kid'', and ``role'', indicates that both $\Lone$
and $A$ rate words as similar if they tend to cooccur with the same
words in $\contexts$.  Observe that four of the ten most likely verbs to
occur with ``kid'' are also very likely to occur with ``guy'', whereas
only the verb ``play'' commonly occurs with both ``role'' and ``guy''.

\begin{table}[t]
\begin{center}
\begin{tabular}{l|l}
Object & Most Likely Verbs \\ \hline
guy &  see get play let  give catch tell do pick need \\ \hline
kid & {\bf get} {\bf see} take help want {\bf tell} teach send {\bf give}
love \\
role & {\bf play} take lead support assume star expand accept sing limit \\
\end{tabular}
\end{center}
\caption[Verbs with high cooccurrence rates]{\label{table:verbs} For each object $\obj$, the ten verbs $\context$ with
highest
$P(\context|\obj)$.  Boldface verbs occur with both the given noun and with
``guy.'' The base language model is \MLE1.}
\end{table}

\begin{table}
\begin{center}
\begin{tabular}{l}
(1) electrocute
(2) shortchange
(3) bedevil
(4) admire
(5) bore
(6) fool \\
$\quad$ (7) bless
$\cdots$
(26) play
$\cdots$
 (49) get
$\cdots$
\end{tabular}
\end{center}
\caption[Exceptional verbs]{\label{table:sigverbs}  Verbs with highest $P(\context|
\mbox{``guy''})/P(\context)$ ratios.  The numbers in parentheses are ranks.}
\end{table}
If we sort the verbs by decreasing $P(\context|
\mbox{``guy''})/P(\context)$, a different order emerges (table
\ref{table:sigverbs}):
``play'', the most likely verb to cooccur with ``role'', is
ranked higher than ``get'', the most likely verb to cooccur with ``kid'',
thus indicating why ``role'' has a higher confusion probability with
respect to ``guy'' than ``kid'' does.

\begin{table}
\begin{center}
\begin{tabular}{ll|ll|ll}
\multicolumn{2}{c}{$L$} &\multicolumn{2}{c}{$A$}
&\multicolumn{2}{c}{$\conf$} \\ \hline
GUY & 0.000000    & GUY & 0.000000    & role & 0.050326     \\
kid & 1.174243     & kid & 0.300681     & people & 0.024545   \\
lot & 1.395178     & thing & 0.321719   & fire & 0.021434    \\
thing & 1.407363   & lot & 0.346137     & GUY & 0.017669    \\
reason & 1.416542  & mother & 0.364610  & work & 0.015519    \\
break & 1.424242   & answer & 0.366333  & man & 0.012445     \\
ball & 1.438618    & reason & 0.367112  & lot & 0.011255     \\
answer & 1.440296  & doctor & 0.373428  & job & 0.010992     \\
tape & 1.448657    & boost & 0.377174   & thing & 0.010919   \\
rest & 1.452688    & ball & 0.381274    & reporter & 0.010551\\
\end{tabular}
\end{center}
\caption[Closest words to ``guy'': \MLEo1]{\label{table:closeMLEo1} 10 closest words to the word ``guy'' for
$A$, $L$, and
$\conf$, using \MLEo1 as the base language model.}
\end{table}

Finally, we examine the effect of deleting singletons from the base
language model.  Table
\ref{table:closeMLEo1} shows the ten closest words, in order, when the base
language model is \MLEo1. The relative order of the four closest words
remains the same; however, the next six words are quite different from
those for \MLE1.  This data suggests that the effect of singletons on
calculations of similarity is quite strong, as is borne out by the
experimental evaluations described in section
\ref{sec:wsd-eval}. We conjecture that this effect is due to the fact
that there are many very low frequency verbs $\context$ in the
data. Omitting singletons involving such words could then drastically
alter the number of $\context$'s that cooccur with both $\obj$ and
$\obj'$. Since our similarity functions depend on such words, it is
perhaps not so surprising that the effect on similarity values of
deleting singletons is rather dramatic. In contrast, a
\backoff\, language model is not as sensitive to missing singletons because of the Good-Turing discounting of
small counts and inflation of zero counts.


\subsection{Performance of Similarity-Based Methods}
\label{sec:wsd-eval}
\begin{figure}[htb]
\epsfscaledbox{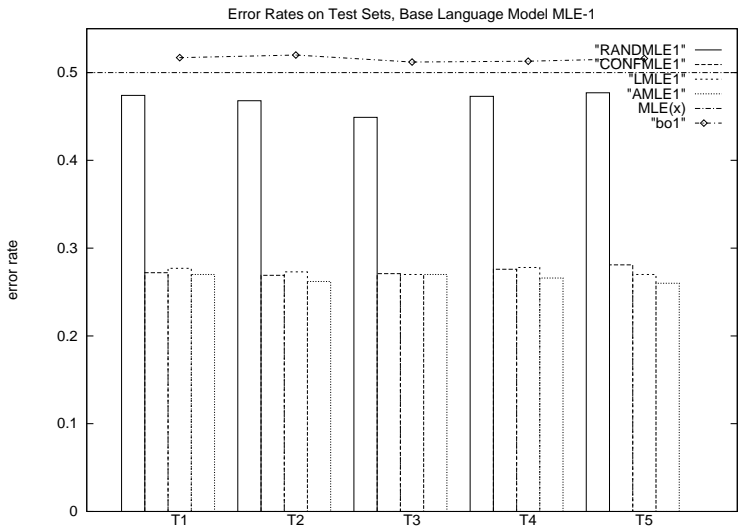}{4in}
\caption[Error rates for \MLE1]{\label{fig:MLE1}  Error rates for each test set, where the
base language model was \MLE1.  The methods, going from left to right,
are $\rand\;$, $\conf$, $\Lone$, and $A$, and the performances shown
are for settings of $\beta$ that were optimal for the corresponding
training set.  $\beta$ values for $\Lone$ ranged from $4.0$ to
$4.5$.  $\beta$ values for $A$ ranged from $10$ to $13$.}
\end{figure}

Figure \ref{fig:MLE1} shows the error rate results on the five test sets, using
\MLE1 as the base language model.  The parameter $\beta$ was always
set to the optimal value for the corresponding parameter training set.
\rand, which is shown for comparison purposes, simply chooses the
weights $W(\obj, \obj')$ randomly.  ${\cal S}(\obj)$ was set equal to
$\objects$ in all cases.

The similarity-based methods consistently outperform the MLE method
(which, recall, always had an error rate of .5) and Katz's back-off
method (which always had an error rate of about .51) by a huge margin;
therefore, we conclude that similarity information is very useful for
unseen word pairs where unigram frequency is not informative.
The similarity-based methods also do much better than $\rand$, which
indicates that it is not enough to simply combine information from
other words arbitrarily: it is quite important to take word similarity
into account.  In all cases, $A$ edged out the other methods.  The
average improvement in using $A$ instead of $\conf$ is .0082; this
difference is significant to the .1 level ($p < .085$) according to
the paired t-test.  

\begin{figure}[htb]
\epsfscaledbox{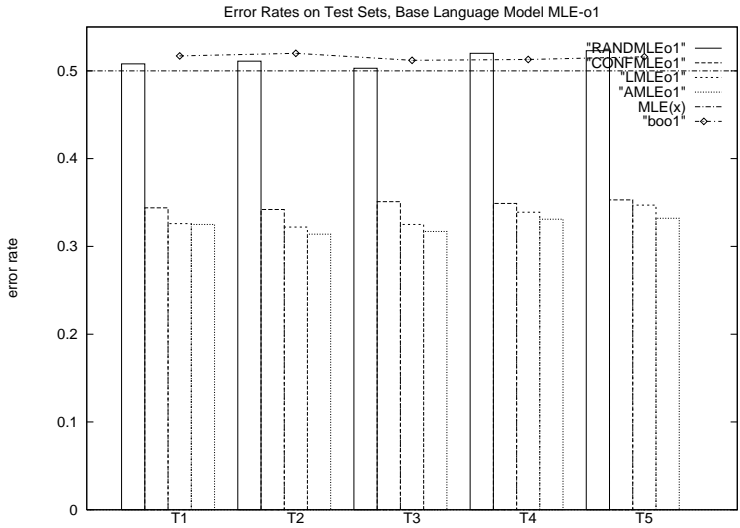}{4in}
\caption[Error rates for \MLEo1]{\label{fig:MLEo1}  Error rates for each test set, where the
base language model was \MLEo1.  The methods, going from left to right,
are $\rand\;$, $\conf$, $\Lone$, and $A$, and the performances shown
are for settings of $\beta$ that were optimal for the corresponding
training set.  $\beta$ values for $\Lone$ ranged from $6$ to
$11$.  $\beta$ values for $A$ ranged from $21$ to $22$.}
\end{figure}

The results for the \MLEo1 case are depicted in figure
\ref{fig:MLEo1}.  Again, we see the similarity-based methods
achieving far lower error rates than the MLE, back-off, and $\rand$
methods, and again, $A$ always performed the best.  However, omitting
singletons amplified the disparity between $A$ and $\conf$:
the average difference in their error rates increases to $.024$, which is
significant to the .01 level (paired t-test).

An important observation is that all methods, including \rand, were
much more effective if singletons were included in the base language
model; thus, in the case of unseen word pairs, it is clear that
singletons should not be ignored by similarity-based models.

\begin{figure}[htb]
\epsfscaledbox{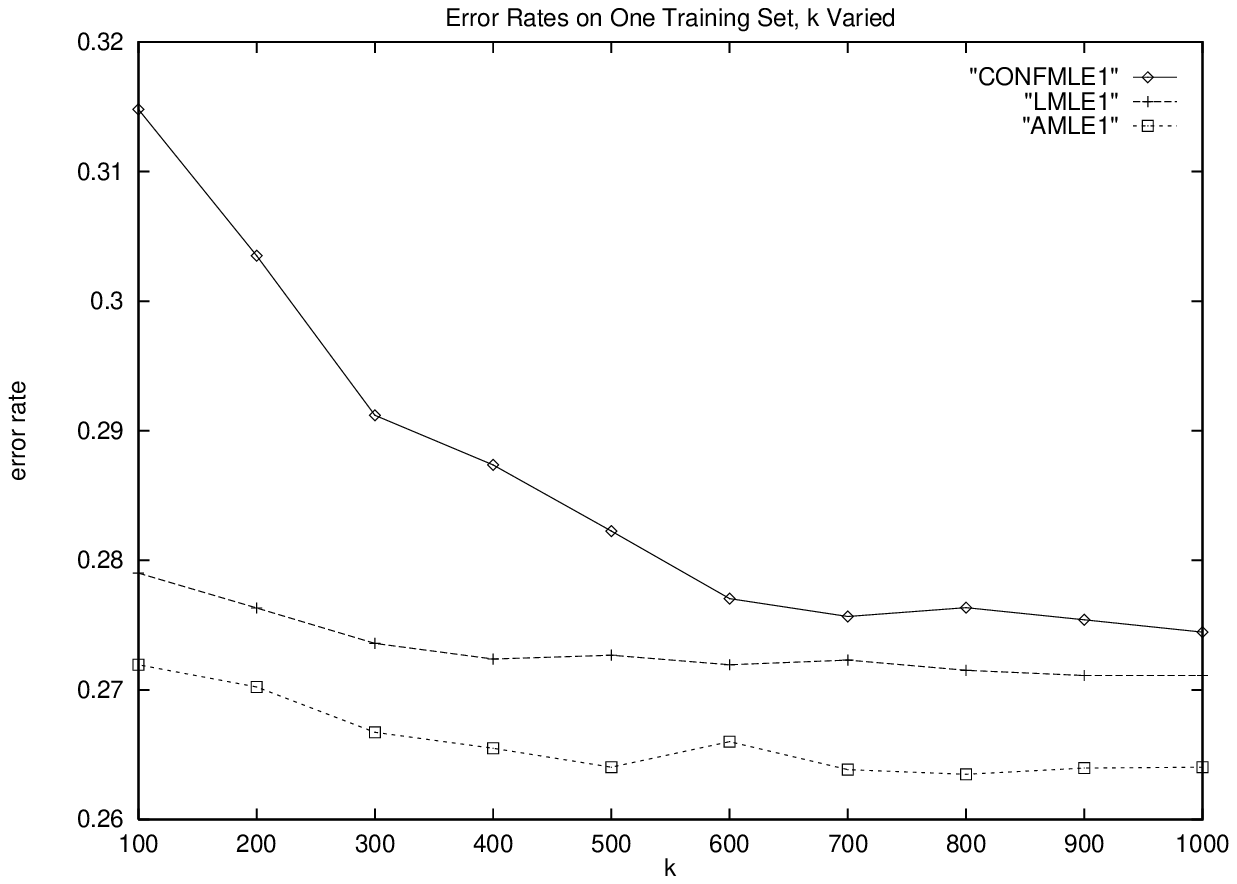}{4in}
\caption[Effect of $k$.]{\label{fig:k}Error rates for one training set as $k$ varies, where the
base language model was \MLE1.  $\beta$ was set to the optimal value
(4.5 for $L$, 13 for $A$).}
\end{figure}

Recall that in these experiments we set ${\cal S}(\obj) = \objects$.  From
the point of view of computational efficiency, it may not be desirable
to sum over all the words in $\objects$.  We experimented with
using only the $k$ closest words to $\obj$, where $k$ varied
from 100 to 1000 $(= |\objects|)$.  We see from figure \ref{fig:k} that
stopping at $k = 600$ is sufficient to capture most of the performance
improvement.  It also appears that $\Lone$ and $A$ use the closest words
more efficiently, as we could sum over $10$ times fewer words
($k=100$) at a performance penalty of less than  $1 \%$; stopping at
$k=100$ for $\conf$ would result in increasing the error rate by $4 \%$.

\section{Conclusion}

Automatically-derived similarity-based language models provide an
appealing approach for dealing with data sparseness.  We suggest a
framework which relies on maximum likelihood estimates when reliable
statistics are available, and uses similarity-based estimates only in
situations where data is lacking.

We have described and compared the performance of four such models
against two standard estimation methods, the MLE method and Katz's
back-off scheme, on a pseudo-word disambiguation task.  We observed
that the similarity-based methods perform much better then the
standard methods on unseen word pairs, with the method based on the KL
divergence to the mean being the best overall.

We also investigated Katz's claim that one can build more compact
language models without suffering significant performance degradation
by discarding singletons in the training data.  Our results indicate
that for similarity-based language modeling, singletons are quite
important; their omission leads to noticeably higher error rates.


\setcounter{chapter}{4}
\chapter{Similarity-Based Estimation for Speech Recognition}
\label{ch:pp}

The previous chapter looked at the performance of three
similarity-based methods on a simple disambiguation task.  This
chapter tackles the more realistic problems of perplexity reduction
and speech-recognition error reduction.  The distance function used
here is the KL divergence, and the base language model is the
\backoff\, estimate.  The similarity-based model considered in this
chapter is based on the model developed in chapter
\ref{ch:wsd}, but has several added features meant to improve both
performance and efficiency.

\section{Introduction}
\label{pp-introduction}

Chapter \ref{ch:wsd} introduced similarity-based methods, developed a
general framework for them, and compared several such methods on a
pseudo-word disambiguation task.  The pseudo-word task was very
convenient from an experimental point of view; it allowed us to limit
the number of senses a (pseudo-) word could have, as well as control
the probabilities of the different senses (recall that we chose to
create pseudo-words out of verbs with similar frequencies).  Thus, we
were able to perform a very clean experiment to demonstrate that
indeed, similarity-based methods do have the potential to outperform
standard approaches to sparse data problems.

However, it must be admitted that pseudo-word disambiguation seems a
bit distant from problems encountered in real-word applications. 
Therefore, in this chapter we evaluate a similarity-based method on
two tasks:  perplexity reduction and speech-recognition error rate.

{\em Perplexity} is often used as a performance metric for language modeling
systems; it is generally assumed that lowering the perplexity is
correlated with better performance in practice \cite{Jelinek:92a}.
Let $P_{{\sc LM}}$ be a probability model and $S$
some\footnote{Jelinek et al. note that the perplexity is a more
accurate measure of the difficulty of recognition if the sample is
large.} sample of text.  Then the perplexity ${\rm PP}$ measures how well
$P_{{\sc LM}}$ models $S$:
\begin{displaymath}
{\rm PP} = P_{{\sc LM}}(S)^{-1/|S|}.
\end{displaymath}
The intuition behind this expression is that a good language model
should assign high probability (and therefore low perplexity) to $S$,
since $S$ was generated by the (unknown) source distribution for the
language.  Another way to look at it is to regard the perplexity as
measuring the average branching of the text from the point of the
language model.  For example, suppose we have two
language models, $P_1$ and $P_2$.  If it turns out that according to
$P_1$, the only words that have a high probability of occurring after
the word ``San'' are ``Juan'' and ``Jose'', whereas according to
$P_2$, ``Juan'', ``Jose'', ``cat'', and ``dog'' all have a high
probability of occurring after ``San'', then we would say that $P_1$
is the better language model.  It is more certain about which words
can follow ``San''; one could say that it is less perplexed.

In this chapter, we model the probabilities of pairs of adjacent
words rather than object-verb pairs; that is, $\objects= \contexts$,
and the pair $(\obj,\context)$ refers to the event that the two-word
sequence, or {\em bigram}, ``$\obj\, \context$'' occurred in the
training sample.  We thus tackle the problem of {\em bigram language
modeling}, which is a special case of {\em $n$-gram language
modeling}; $n$-grams are the dominant language-modeling technology in
speech recognition today.  In a bigram language model, the probability
of a string of words is factored into a product of conditional word pair
probabilities:
\begin{displaymath}
P_{{\sc LM}}(w_1 w_2 \ldots w_n) = \prod_{i} P_{{\sc LM}}(w_{i}|w_{i-1}).
\end{displaymath}
Then, the perplexity of a  bigram  model $P_{{\sc
LM}}$ with respect to the string $w_1 w_2 \ldots w_n$ 
is 
\begin{displaymath}
\left( \prod_{i} P_{{\sc LM}}(w_i|w_{i-1})
\right)^{-1/n} = \exp { \left( -\frac{1}{n} \sum_{i} \log P_{{\sc
LM}}(w_i|w_{i-1}) \right)},
\end{displaymath}
where  base 10  logarithm and exponential functions are used
throughout this chapter, as in chapter \ref{ch:wsd}.

Given our concern with the practicality of similarity-based
estimation, we will also consider several heuristics for improving the
efficiency and performance of similarity-based models.  In particular,
we will be interested in the effect of limiting the number of similar
words that are consulted in making an estimate for a particular
bigram.  Another heuristic we apply is to interpolate the similarity
information with the {\em unigram} (single word) probability used by
Katz's \backoff\, method.  We find that combining these two estimates
does improve performance, although it is best not to rely too much on
the unigram probability (this is a gratifying result, as it tells us
that the similarity information is more important than the unigram
information).

The rest of this chapter proceeds as follows. Section
\ref{sec:heuristics} explains the modifications we make to the
similarity-based model introduced in the previous chapter.  Section
\ref{pp-eval} presents our evaluation results: the new similarity model
achieved a 20\% reduction in perplexity with respect to Katz's
\backoff\, model on unseen bigrams in {\em Wall Street Journa}l
data. These constituted just 10.6\% of the test sample, leading to an
overall reduction in test-set perplexity of 2.4\%. We also
experimented with an application of our language modeling technique to speech
recognition, and found that it yielded a statistically significant reduction in
recognition error.  Section \ref{sec:further} points out some
directions for further research.

\section{The Similarity Model}
\label{sec:heuristics}

Recall the general form for similarity-based models developed in
chapter \ref{ch:wsd}:
\begin{displaymath}
\hat{P}(\context| \obj) = \left\{\!\!\!\!
\begin{array}{l@{\hspace{0.6ex}}l}
P_d(\context | \obj) & \mbox{if $\counts(\obj,\context) > 0$} \\
\alpha(\obj)P_r(\context | \obj) & \mbox{otherwise  ($(x,y)$ is unseen)}
\end{array}\right.\;. \qquad \qquad  \qquad  \qquad (\ref{genmodel})  
\end{displaymath}
We defined $P_r$ to be $\psim$, where 
\begin{displaymath}
\psim(\context|\obj) = \sum_{\obj' \in {\cal S}(\obj)}
\frac{W(\obj,\obj')}{\sum_{\obj' \in {\cal
S}(\obj)}W(\obj,\obj')}{P(\context|\obj')}.  \qquad \qquad  \qquad (\ref{sim-formula})
\end{displaymath}

Now, in the last chapter, we simply set ${\cal S}(\obj)$, the set of
objects most similar to $\obj$, to be equal to the
set $\objects$.  From a computational standpoint,
though, this is somewhat unsatisfactory if $\objects$ is large.
Furthermore, it might well be the case that only a few of the closest
objects contribute to the sum in (\ref{sim-formula}).  Therefore, we
experiment in this chapter with limiting the size of ${\cal S}(\obj)$.
We now introduce parameters $k$ and $t$, and define
${\cal S}(w_1)$ to be the set of
at most $k$ words $w_1'$ (excluding $w_1$ itself) that
satisfy $D(w_1 \parallel w'_{1})<t$.
We need to tune $k$ and $t$ experimentally.

We will use the KL divergence as distance function in the experiments
described below, since we did not provide performance results for it
in the previous chapter.  
Recall that the weight function
$W(\obj,\obj')$ for the KL divergence was defined to be
\[
W(\obj,\obj') = 10^{ - \beta D(\obj \parallel \obj')}\eqpunc{.}
\]
Again, the parameter
$\beta$ controls the relative contribution of
words at different distances from $\obj$: as $\beta$
increases, the nearest words to $\beta$ get relatively more weight.
As $\beta$ decreases, remote words have a larger effect on the sum (\ref{sim-formula}). Like $k$ and
$t$, $\beta$ is tuned experimentally.

While in the preceding chapter we set $P_r$ to be $\psim$, we
shall see that it is better to smooth $\psim$ by
interpolating it with the unigram probability $P(\context)$ (recall that
Katz used $P(\context)$ as $P_r(\context|\obj)$). Using linear interpolation we
get
\begin{equation}
\label{ourPr}
P_r(\context|\obj) = \gamma P(\context) + (1 - \gamma)\psim(\context|\obj)\;,
\end{equation}
where $\gamma$ is an experimentally-determined interpolation parameter.
This smoothing appears to compensate for
inaccuracies in $\psim(\context|\obj)$, mainly for infrequent
conditioning words.
However, as the evaluation below shows, good values for $\gamma$ are
small, that is, the similarity-based model plays a stronger role than
the independence assumption.

To summarize, we construct a similarity-based model for
$P(\context|\obj)$ and then interpolate it with $P(\context)$.  The
interpolated model (\ref{ourPr}) is used as the probability
redistribution model $P_r$ in (\ref{genmodel}) to obtain better
estimates for unseen bigrams.  Four parameters, to be tuned
experimentally, are relevant for this process: $k$ and $t$, which
determine the set of similar words to be considered, $\beta$, which
determines the relative effect of these words, and $\gamma$, which
determines the overall importance of the similarity-based model.


\section{Evaluation}
\label{pp-eval}

We evaluated our method by comparing its perplexity and effect on
speech-recognition accuracy with the baseline bigram back-off model
developed by MIT Lincoln Laboratories for the {\em Wall Street
Journal} (WSJ) text and dictation corpora provided by ARPA's HLT
program
\cite{Paul:91a}.%
\footnote{The ARPA WSJ development corpora come in two
versions, one with verbalized punctuation and the other without. We
used the latter in all our experiments.}  The baseline back-off model
closely follows the Katz design discussed in section
\ref{sec:smoothing}, except that for the sake of compactness all singleton bigrams
are treated as unseen (recall that this omission of singletons was
quite detrimental to the simple similarity-based models considered in
the previous chapter). The counts used in this model and in ours were
obtained from 40.5 million words of WSJ text from the years 1987-89.

For the perplexity evaluation, we tuned the similarity model parameters
by minimizing perplexity via a simple grid search on an
additional sample of 57.5 thousand
words of WSJ text drawn from the ARPA HLT development test set.
The best parameter values found were
$k=60$, $t=2.5$, $\beta=4$ and
$\gamma=0.15$. For these values, the improvement in perplexity for
unseen bigrams in a held-out 18 thousand word sample, in which 10.6\%
of the bigrams are unseen, 
is just over 20\%. This improvement on unseen bigrams
corresponds to an overall test set perplexity improvement of 2.4\%
(from 237.4 to 231.7).
\begin{table*}
\begin{center}
\begin{tabular}{rrrrrr}
$k$ & $t$ & $\beta$ & $\gamma$ & training reduction (\%) & test reduction (\%)\\
\hline
60 & 2.5 & 4 & 0.15 & 18.4 & 20.51 \\
50 & 2.5 & 4 & 0.15 & 18.38 & 20.45 \\
40 & 2.5 & 4 & 0.2 & 18.34 & 20.03 \\
30 & 2.5 & 4 & 0.25 & 18.33 & 19.76 \\
70 & 2.5 & 4 & 0.1 & 18.3 & 20.53 \\
80 & 2.5 & 4.5 & 0.1 & 18.25 & 20.55 \\
100 & 2.5 & 4.5 & 0.1 & 18.23 & 20.54 \\
90 & 2.5 & 4.5 & 0.1 & 18.23 & 20.59 \\
20 & 1.5 & 4 & 0.3 & 18.04 & 18.7 \\
10 & 1.5 & 3.5 & 0.3 & 16.64 & 16.94
\end{tabular}
\end{center}
\caption{Perplexity reduction on unseen bigrams for different model parameters}
\label{perp-results}
\end{table*}
Table \ref{perp-results} shows reductions in training and test
perplexity, sorted by training reduction, for different choices of
$k$. The values of $\beta$, $\gamma$
and $t$ are the best ones found for each $k$.%

From equation (\ref{sim-formula}), it is clear that the computational cost of
applying the similarity model to an unseen bigram is
$O(k)$. Therefore, lower values of $k$ (and
$t$ as well) are computationally preferable.
From the table, we can see that reducing $k$ to 30 incurs
a penalty of less than 1\% in the perplexity improvement, so
relatively low values of $k$ appear to be sufficient to achieve most
of the benefit of the similarity model. As the table also shows, the
best value of $\gamma$ increases as $k$ decreases, that is, for lower
$k$ a greater weight is given to the conditioned word's frequency.
This suggests that the predictive power of neighbors beyond the
closest 30 or so can be modeled fairly well by the overall frequency
of the conditioned word.

The bigram similarity model was also tested as a language model in
speech recognition. The test data for this experiment were pruned word
lattices for 403 WSJ closed-vocabulary test sentences.  Arc scores in those
lattices are sums of an acoustic score (negative log likelihood) and a
language-model score, in this case the negative log probability
provided by the baseline bigram model.

From the given lattices, we constructed new lattices in which the arc
scores were modified to use the similarity model instead of the
baseline model.  We compared the best sentence hypothesis in each
original lattice and in the modified one, and counted the word
disagreements in which one of the hypotheses is correct.  There were a
total of 96 such disagreements. The similarity model was correct in 64
cases, and the back-off model in 32. This advantage for the similarity
model is statistically significant at the 0.01 level. The overall
reduction in error rate is small (from 21.4\% to 20.9\%) because the
number of disagreements is small compared with the overall number of
errors in the recognition setup used in these experiments.

Table \ref{rec-eg} shows some examples of speech recognition
disagreements between the two models. The hypotheses are labeled `B'
for back-off and `S' for similarity, and the bold-face words are
errors. The similarity model seems to be  better at modeling
regularities such as semantic parallelism in lists and avoiding a past
tense form after ``to.'' On the other hand, the similarity model
makes several mistakes in which a function word is inserted in a place
where punctuation would be found in written text.

\begin{table*}
\begin{center}
\begin{tabular}{l|l}
B & commitments \ldots from leaders {\bf felt the} three point six
billion dollars \\
\hline S & commitments \ldots from leaders fell to three point six
billion dollars \\
\hline \hline B & followed by France the US {\bf
agreed in}  Italy  \\
\hline S & followed by France the US Greece
\ldots Italy \\
\hline\hline B & he whispers to {\bf made a} \\
\hline
S & he whispers to an aide \\
\hline\hline B & the necessity for change
{\bf exist} \\
 \hline S & the necessity for change exists \\
\hline\hline\hline B & without \ldots additional reserves Centrust  would
have reported \\
\hline S & without \ldots additional reserves {\bf 
of} Centrust would have reported \\
\hline \hline  B & in the darkness
past the church \\
\hline S & in the darkness {\bf passed} the church
\end{tabular}
\end{center}
\caption{Speech recognition disagreements between models}
\label{rec-eg}
\end{table*}
\section{Further Research}
\label{sec:further}

The model presented in this chapter provides a modification of the
scheme for similarity-based estimation described in the preceding
chapter; several heuristics for improving speed and performance were
incorporated.  We have demonstrated that the augmented model can be of
use in practical speech recognition systems.  We now discuss some
possible further directions to explore.

It may be possible to simplify the current model parameters 
somewhat, especially with respect to the parameters $t$ and $k$
used to select the nearest neighbors of a word.  On the other hand, it
may be the case that using the same $t$ and $k$ for all words is too
simplistic, although training a model in which $t$ and $k$ differ from
word to word would involve massive sparse data problems.

A more substantial variation would be to base the model on the similarity
between conditioned words ($\context$) rather than on the similarity between
conditioning words ($\obj$).  For example, Essen and Steinbiss's variation 1
considers the confusion probability (\ref{conf}) of  contexts
rather than  objects \cite{Essen+Steinbiss:92a}.  However, they
noted that model 1-A was equivalent to model 2-B (which we discussed
in section \ref{sec:confusion}; it uses the
confusion probability of conditioning words), and that their other model
using variation 1 did not perform as well.

Other evidence may be combined with the similarity-based estimate.
For instance, it may be advantageous to weigh the similarity-based estimate by some
measure of the reliability of the similarity function and of the
neighbor distributions.  A second possibility is to take into account
negative evidence, as \namecite{\vhref} did (see the discussion in
section \ref{sec:word-clustering}).  For example, if $\obj$ is
frequent, but $\context$ never followed it, there may be enough
statistical evidence to put an upper bound on the estimate of
$P(\context|\obj)$. This may require an adjustment of the
similarity-based estimate, possibly along the lines of the work of
\namecite{Rosenfeld+Huang:92a}.  

Finally, the similarity-based model may be applied to configurations
other than bigrams.  For trigrams, it is necessary to measure
similarity between different conditioning bigrams. This can be done
directly, by measuring the distance between distributions of the form
$P(w_3|w_1,w_2)$, corresponding to different bigrams
$(w_1,w_2)$. Alternatively, and more practically, it may be possible
to define a similarity measure between trigrams as a function of the
similarities between corresponding words in them. 

\section{Conclusions}
\label{conclusions}

Similarity-based models suggest an appealing approach to dealing with
data sparseness. Based on corpus statistics, they provide analogies
between words that often agree with our linguistic and domain
intuitions.  In the previous chapter we looked at the performance of
various instantiations of a simple similarity-based model.  In this
chapter we presented a variant that provides noticeable improvement
over Katz's \backoff\, estimation method on realistic evaluation tasks.

The improvement we achieved for a bigram model is statistically
significant, although it is modest in its overall effect because of
the small proportion of unseen events.  While we have used bigrams as
an easily accessible platform to develop and test the model, more
substantial improvements might be obtainable for more informative
configurations.  An obvious case is that of trigrams, for which the
sparse data problem is much more severe.  For example, Doug Paul
(personal communication) reports that for WSJ trigrams over a 20000
word vocabulary, only 58.6\% of the test set trigrams occurred in 40
million of words of training data.

\chapter{Conclusion}
\label{ch:concl}

\begin{quote}
This paper is an absolute leviathan!  Reverberating with history and
personal recollection and occasionally exploding with well-aimed
critical bursts, it sweeps you up like a great tidal wave and carries
you along for over one hundred pages at an accelerating tempo, leaving
you at the end with a sense that its driving energy has still not
spent itself.... \cite[pg. viii]{Rosenkrantz:83a}
\end{quote}

We have presented two ways to make use of distributional similarity
for applications in natural language processing.  The first was a
distributional clustering method, which proved not only to create
clusters that seem to correspond to intuitive sense distinctions but
also to lead to a cluster-based language model with good predictive
power.  Our clustering method yields soft, hierarchical clusters.  Our
use of soft clustering in a language processing context appears to be
novel, but is rather natural, since many words are ambiguous.

We also presented a nearest-neighbor approach, where we combined
estimates from similar words rather than from cluster centroids.  This
approach has the advantage of computational efficiency, since we do
not need to engage in the iterative estimation necessary in our
clustering work.  We showed that methods based on the KL divergence
provided substantial improvement over Katz's \backoff\, method for
unseen word pairs, and noticeable improvement over Essen and
Steinbiss's confusion probability on a pseudo-word disambiguation
task.  To further demonstrate that similarity information can be
helpful for applications, we also showed that an extension of our
similarity-based model can produce both perplexity reduction and
speech recognition error-rate reduction.

We can only conclude that the incorporation of similarity information
has the potential to provide better results in the area of language
modeling.  But our techniques may extend farther than that.  Indeed,
our clustering work certainly seems applicable to other problems such
as automatic thesaurus construction or lexicon acquisition.  The fact
that the clusterings we produce are probabilistic may again be an
advantage, since for instance words may appear in more than one
thesaurus category.  It would also be interesting to experiment with
applying our techniques to the problems of document clustering and
indexing, as mentioned at the end of chapter \ref{ch:clust}.

This brings up a deeper question, though.  What is the proper way to
evaluate the inherent quality of clusterings (as opposed to measuring
the performance gain clusterings can provide)?  We need a good way to
talk about how different one clustering is from another in order to
analyze competing clustering methods.  As we move to larger and larger
data sets, it becomes more and more impractical to perform an
evaluation such as that described by \namecite{\vhref} where
automatically-derived classes were compared to classes created by
humans.  Perhaps one fruitful direction, at least for hierarchical
clusterings, would be to look at edit distances between trees (see,
e.g., \namecite{Kannan+Warnow+Yooseph:95a}).

Another key question to address is whether we can formulate {\em
adaptive} versions of our algorithms.  Since our methods do not rely
on heavily annotated samples, it is easy to acquire new training data.
What we would like is a way to incorporate new information without
having to restart the clustering process (in the distributional
clustering case) or recalculate the similarity matrix (in the
nearest-neighbor case).  For clustering, the fact that we use soft
classes may once again provide the answer, since we can reestimate
membership probabilities as new data comes in; if we built hard
clusters, we would have to adjust for  prematurely grouping two
objects together or splitting two objects apart.

\end{document}